\documentclass{bmvc2k}

\title{Category-level Text-to-Image Retrieval Improved: Bridging the Domain Gap with Diffusion Models and Vision Encoders}

\addauthor{Faizan Farooq Khan}{faizan.khan@kaust.edu.sa}{1}
\addauthor{Vladan Stojnić}{stojnvla@fel.cvut.cz}{2}
\addauthor{Zakaria Laskar}{laskazak@fel.cvut.cz}{2}
\addauthor{Mohamed Elhoseiny}{mohamed.elhoseiny@kaust.edu.sa}{1}
\addauthor{Giorgos Tolias}{toliageo@fel.cvut.cz}{2}

\addinstitution{
 King Abdullah University of Science and Technology \\
 Thuwal, Saudi Arabia
}
\addinstitution{
Visual Recognition Group \\
Faculty of Electrical Engineering\\
Czech Technical University in Prague
}

\runninghead{Khan et.al.}{Category-level Text-to-Image Retrieval Improved}

\definecolor{cvprblue}{rgb}{0.21,0.49,0.74}
\usepackage{nicefrac}
\usepackage{svg}
\usepackage{amsmath}
\usepackage{pgfplots}
\usepackage{amssymb}

\usepackage{multirow}
\usepackage{booktabs}
\usepackage{adjustbox}
\usepackage{wrapfig}
\usepackage{pgfplotstable}
\usepackage{array}
\usepackage{nicefrac}
\usepackage{subcaption}
\usepackage{wrapfig}
\usetikzlibrary{calc}
\usepackage{colortbl}
\usepackage{cleveref}

\usepackage[utf8]{inputenc}
\usepackage{graphicx}
\usepackage{lipsum}

\definecolor{blond}{rgb}{0.98, 0.94, 0.75}

\def\eg{\emph{e.g}\bmvaOneDot}
\def\Eg{\emph{E.g}\bmvaOneDot}
\def\etal{\emph{et al}\bmvaOneDot}

\begin{document}

\newcommand{\mypartight}[1]{\noindent {\bf #1}}
\newcommand{\myparagraph}[1]{\vspace{3pt}\noindent\textbf{#1}\xspace}

\newcommand{\optional}[1]{{\color{gray}{#1}}}
\newcommand{\alert}[1]{{\color{red}{#1}}}
\newcommand{\gt}[1]{{\color{purple}{GT: #1}}}
\newcommand{\gtt}[1]{{\color{purple}{#1}}}
\newcommand{\gtr}[2]{{\color{purple}\st{#1} {#2}}}
\newcommand{\gray}[1]{{\color{gray}{#1}}}
\newcommand{\vsc}[1]{{\color{blue}{VS: #1}}}
\newcommand{\zl}[1]{{\color{green}{ZL: #1}}}

\newcommand{\vst}[1]{{\color{blue}{#1}}}
\newcommand{\fai}[1]{{\color{cyan}{#1}}}


\newcommand{\fig}[2][1]{\includegraphics[width=#1\linewidth]{fig/#2}}
\newcommand{\figh}[2][1]{\includegraphics[height=#1\linewidth]{fig/#2}}
\newcommand{\figa}[2][1]{\includegraphics[width=#1]{fig/#2}}
\newcommand{\figah}[2][1]{\includegraphics[height=#1]{fig/#2}}

\newcommand{\figsup}[2][1]{\includegraphics[width=#1\linewidth]{fig_supp/#2}}
\newcommand{\figsuph}[2][1]{\includegraphics[height=#1\linewidth]{fig_supp/#2}}
\newcommand{\figsupa}[2][1]{\includegraphics[width=#1]{fig_supp/#2}}
\newcommand{\figsupah}[2][1]{\includegraphics[height=#1]{fig_supp/#2}}
\newcommand{\figsupc}[2][1]{\includegraphics[width=#1\linewidth,cfbox=OliveGreen 1.5pt 1.5pt]{fig_supp/#2}}
\newcommand{\figsupw}[2][1]{\includegraphics[width=#1\linewidth,cfbox=BrickRed 1.5pt 1.5pt]{fig_supp/#2}}
\newcommand{\figsupq}[2][1]{\includegraphics[width=#1\linewidth,cfbox=OrangeFrame 1.5pt 1.5pt]{fig_supp/#2}}
\newcommand{\bfigsupc}[2][1]{\includegraphics[width=#1\linewidth,height=#1\linewidth,cfbox=OliveGreen 1.5pt 1.5pt]{fig_supp/#2}}
\newcommand{\bfigsupw}[2][1]{\includegraphics[width=#1\linewidth,height=#1\linewidth,cfbox=BrickRed 1.5pt 1.5pt]{fig_supp/#2}}
\newcommand{\bfigsupq}[2][1]{\includegraphics[width=#1\linewidth,height=#1\linewidth,cfbox=OrangeFrame 1.5pt 1.5pt]{fig_supp/#2}}

\newcommand{\figreb}[2][1]{\includegraphics[width=#1\linewidth]{fig_reb/#2}}
\newcommand{\figrebh}[2][1]{\includegraphics[height=#1\linewidth]{fig_reb/#2}}
\newcommand{\figreba}[2][1]{\includegraphics[width=#1]{fig_reb/#2}}
\newcommand{\figrebah}[2][1]{\includegraphics[height=#1]{fig_reb/#2}}
\newcommand{\figrebc}[2][1]{\includegraphics[width=#1\linewidth,cfbox=OliveGreen 1.0pt 1.0pt]{fig_reb/#2}}
\newcommand{\figrebw}[2][1]{\includegraphics[width=#1\linewidth,cfbox=BrickRed 1.0pt 1.0pt]{fig_reb/#2}}
\newcommand{\figrebq}[2][1]{\includegraphics[width=#1\linewidth,cfbox=OrangeFrame 1.0pt 1.0pt]{fig_reb/#2}}
\newcommand{\bfigrebc}[2][1]{\includegraphics[width=#1\linewidth,height=#1\linewidth,cfbox=OliveGreen 1.0pt 1.0pt]{fig_reb/#2}}
\newcommand{\bfigrebw}[2][1]{\includegraphics[width=#1\linewidth,height=#1\linewidth,cfbox=BrickRed 1.0pt 1.0pt]{fig_reb/#2}}
\newcommand{\bfigrebq}[2][1]{\includegraphics[width=#1\linewidth,height=#1\linewidth,cfbox=OrangeFrame 1.0pt 1.0pt]{fig_reb/#2}}
\newcommand{\mr}[2]{\multirow{#1}{*}{#2}}
\newcommand{\mc}[2]{\multicolumn{#1}{c}{#2}}
\newcommand{\Th}[1]{\textsc{#1}}
\newcommand{\tb}[1]{\textbf{#1}}

\newcommand{\stddev}[1]{\scriptsize{$\pm#1$}}

\newcommand{\diffup}[1]{{\color{OliveGreen}{($\uparrow$ #1)}}}
\newcommand{\diffdown}[1]{{\color{BrickRed}{($\downarrow$ #1)}}}

\newcommand{\deltaup}[1]{{\color{OliveGreen}{$\uparrow$ #1}}}
\newcommand{\deltadown}[1]{{\color{BrickRed}{$\downarrow$ #1}}}

\newcommand{\comment} [1]{{\color{orange} \Comment     #1}} 

\def\nmsp{\hspace{-6pt}}
\def\nssp{\hspace{-3pt}}
\def\nxssp{\hspace{-1pt}}
\def\zsp{\hspace{0pt}}
\def\xssp{\hspace{1pt}}
\def\ssp{\hspace{3pt}}
\def\msp{\hspace{6pt}}
\def\lsp{\hspace{12pt}}
\def\xlsp{\hspace{20pt}}

\newcommand{\head}[1]{{\smallskip\noindent\bf #1}}
\newcommand{\equ}[1]{(\ref{equ:#1})\xspace}


\newcommand{\nn}[1]{\ensuremath{\text{NN}_{#1}}\xspace}
\newcommand{\eq}[1]{(\ref{eq:#1})\xspace}


\def\l1{\ensuremath{\ell_1}\xspace}
\def\l2{\ensuremath{\ell_2}\xspace}


\newcommand{\tran}{^\top}
\newcommand{\mtran}{^{-\top}}
\newcommand{\zcol}{\mathbf{0}}
\newcommand{\zrow}{\zcol\tran}

\newcommand{\ind}{\mathds{1}}
\newcommand{\expect}{\mathbb{E}}
\newcommand{\nat}{\mathbb{N}}
\newcommand{\zahl}{\mathbb{Z}}
\newcommand{\real}{\mathbb{R}}
\newcommand{\proj}{\mathbb{P}}
\newcommand{\prob}{\mathbf{Pr}}

\newcommand{\mif}{\textrm{if }}
\newcommand{\other}{\textrm{otherwise}}
\newcommand{\minimize}{\textrm{minimize }}
\newcommand{\maximize}{\textrm{maximize }}

\newcommand{\id}{\operatorname{id}}
\newcommand{\const}{\operatorname{const}}
\newcommand{\sgn}{\operatorname{sgn}}
\newcommand{\erf}{\operatorname{erf}}
\newcommand{\var}{\operatorname{Var}}
\newcommand{\mean}{\operatorname{mean}}
\newcommand{\trace}{\operatorname{tr}}
\newcommand{\diag}{\operatorname{diag}}
\newcommand{\vect}{\operatorname{vec}}
\newcommand{\cov}{\operatorname{cov}}

\newcommand{\softmax}{\operatorname{softmax}}
\newcommand{\clip}{\operatorname{clip}}

\newcommand{\defn}{\mathrel{:=}}
\newcommand{\peq}{\mathrel{+\!=}}
\newcommand{\meq}{\mathrel{-\!=}}

\newcommand{\floor}[1]{\left\lfloor{#1}\right\rfloor}
\newcommand{\ceil}[1]{\left\lceil{#1}\right\rceil}
\newcommand{\inner}[1]{\left\langle{#1}\right\rangle}
\newcommand{\norm}[1]{\left\|{#1}\right\|}
\newcommand{\frob}[1]{\norm{#1}_F}
\newcommand{\card}[1]{\left|{#1}\right|\xspace}
\newcommand{\diff}{\mathrm{d}}
\newcommand{\der}[3][]{\frac{d^{#1}#2}{d#3^{#1}}}
\newcommand{\pder}[3][]{\frac{\partial^{#1}{#2}}{\partial{#3^{#1}}}}
\newcommand{\ipder}[3][]{\partial^{#1}{#2}/\partial{#3^{#1}}}
\newcommand{\dder}[3]{\frac{\partial^2{#1}}{\partial{#2}\partial{#3}}}

\newcommand{\wb}[1]{\overline{#1}}
\newcommand{\wt}[1]{\widetilde{#1}}

\newcommand{\cA}{\mathcal{A}}
\newcommand{\cB}{\mathcal{B}}
\newcommand{\cC}{\mathcal{C}}
\newcommand{\cD}{\mathcal{D}}
\newcommand{\cE}{\mathcal{E}}
\newcommand{\cF}{\mathcal{F}}
\newcommand{\cG}{\mathcal{G}}
\newcommand{\cH}{\mathcal{H}}
\newcommand{\cI}{\mathcal{I}}
\newcommand{\cJ}{\mathcal{J}}
\newcommand{\cK}{\mathcal{K}}
\newcommand{\cL}{\mathcal{L}}
\newcommand{\cM}{\mathcal{M}}
\newcommand{\cN}{\mathcal{N}}
\newcommand{\cO}{\mathcal{O}}
\newcommand{\cP}{\mathcal{P}}
\newcommand{\cQ}{\mathcal{Q}}
\newcommand{\cR}{\mathcal{R}}
\newcommand{\cS}{\mathcal{S}}
\newcommand{\cT}{\mathcal{T}}
\newcommand{\cU}{\mathcal{U}}
\newcommand{\cV}{\mathcal{V}}
\newcommand{\cW}{\mathcal{W}}
\newcommand{\cX}{\mathcal{X}}
\newcommand{\cY}{\mathcal{Y}}
\newcommand{\cZ}{\mathcal{Z}}

\newcommand{\vA}{\mathbf{A}}
\newcommand{\vB}{\mathbf{B}}
\newcommand{\vC}{\mathbf{C}}
\newcommand{\vD}{\mathbf{D}}
\newcommand{\vE}{\mathbf{E}}
\newcommand{\vF}{\mathbf{F}}
\newcommand{\vG}{\mathbf{G}}
\newcommand{\vH}{\mathbf{H}}
\newcommand{\vI}{\mathbf{I}}
\newcommand{\vJ}{\mathbf{J}}
\newcommand{\vK}{\mathbf{K}}
\newcommand{\vL}{\mathbf{L}}
\newcommand{\vM}{\mathbf{M}}
\newcommand{\vN}{\mathbf{N}}
\newcommand{\vO}{\mathbf{O}}
\newcommand{\vP}{\mathbf{P}}
\newcommand{\vQ}{\mathbf{Q}}
\newcommand{\vR}{\mathbf{R}}
\newcommand{\vS}{\mathbf{S}}
\newcommand{\vT}{\mathbf{T}}
\newcommand{\vU}{\mathbf{U}}
\newcommand{\vV}{\mathbf{V}}
\newcommand{\vW}{\mathbf{W}}
\newcommand{\vX}{\mathbf{X}}
\newcommand{\vY}{\mathbf{Y}}
\newcommand{\vZ}{\mathbf{Z}}

\newcommand{\va}{\mathbf{a}}
\newcommand{\vb}{\mathbf{b}}
\newcommand{\vc}{\mathbf{c}}
\newcommand{\vd}{\mathbf{d}}
\newcommand{\ve}{\mathbf{e}}
\newcommand{\vf}{\mathbf{f}}
\newcommand{\vg}{\mathbf{g}}
\newcommand{\vh}{\mathbf{h}}
\newcommand{\vi}{\mathbf{i}}
\newcommand{\vj}{\mathbf{j}}
\newcommand{\vk}{\mathbf{k}}
\newcommand{\vl}{\mathbf{l}}
\newcommand{\vm}{\mathbf{m}}
\newcommand{\vn}{\mathbf{n}}
\newcommand{\vo}{\mathbf{o}}
\newcommand{\vp}{\mathbf{p}}
\newcommand{\vq}{\mathbf{q}}
\newcommand{\vr}{\mathbf{r}}
\newcommand{\Vs}{\mathbf{s}}
\newcommand{\vt}{\mathbf{t}}
\newcommand{\vu}{\mathbf{u}}
\newcommand{\vv}{\mathbf{v}}
\newcommand{\vw}{\mathbf{w}}
\newcommand{\vx}{\mathbf{x}}
\newcommand{\vy}{\mathbf{y}}
\newcommand{\vz}{\mathbf{z}}

\newcommand{\vone}{\mathbf{1}}
\newcommand{\vzero}{\mathbf{0}}

\newcommand{\valpha}{{\boldsymbol{\alpha}}}
\newcommand{\vbeta}{{\boldsymbol{\beta}}}
\newcommand{\vgamma}{{\boldsymbol{\gamma}}}
\newcommand{\vdelta}{{\boldsymbol{\delta}}}
\newcommand{\vepsilon}{{\boldsymbol{\epsilon}}}
\newcommand{\vzeta}{{\boldsymbol{\zeta}}}
\newcommand{\veta}{{\boldsymbol{\eta}}}
\newcommand{\vtheta}{{\boldsymbol{\theta}}}
\newcommand{\viota}{{\boldsymbol{\iota}}}
\newcommand{\vkappa}{{\boldsymbol{\kappa}}}
\newcommand{\vlambda}{{\boldsymbol{\lambda}}}
\newcommand{\vmu}{{\boldsymbol{\mu}}}
\newcommand{\vnu}{{\boldsymbol{\nu}}}
\newcommand{\vxi}{{\boldsymbol{\xi}}}
\newcommand{\vomikron}{{\boldsymbol{\omikron}}}
\newcommand{\vpi}{{\boldsymbol{\pi}}}
\newcommand{\vrho}{{\boldsymbol{\rho}}}
\newcommand{\vsigma}{{\boldsymbol{\sigma}}}
\newcommand{\vtau}{{\boldsymbol{\tau}}}
\newcommand{\vupsilon}{{\boldsymbol{\upsilon}}}
\newcommand{\vphi}{{\boldsymbol{\phi}}}
\newcommand{\vchi}{{\boldsymbol{\chi}}}
\newcommand{\vpsi}{{\boldsymbol{\psi}}}
\newcommand{\vomega}{{\boldsymbol{\omega}}}

\newcommand{\rLambda}{\mathrm{\Lambda}}
\newcommand{\rSigma}{\mathrm{\Sigma}}

\makeatletter
\DeclareRobustCommand\onedot{\futurelet\@let@token\@onedot}
\def\@onedot{\ifx\@let@token.\else.\null\fi\xspace}
\def\eg{\emph{e.g}\onedot} \def\Eg{\emph{E.g}\onedot}
\def\ie{\emph{i.e}\onedot} \def\Ie{\emph{I.e}\onedot}
\def\vs{\emph{vs\onedot}}
\def\cf{\emph{cf}\onedot} \def\Cf{\emph{C.f}\onedot}
\def\etc{\emph{etc}\onedot} \def\vs{\emph{vs}\onedot}
\def\wrt{w.r.t\onedot} \def\dof{d.o.f\onedot}
\def\etal{\emph{et al}\onedot}
\makeatother

\newcommand\rurl[1]{%
  \href{https://#1}{\nolinkurl{#1}}%
}

\maketitle

\begin{abstract}
    This work explores text-to-image retrieval for queries that specify or describe a semantic category. While vision-and-language models (VLMs) like CLIP offer a straightforward open-vocabulary solution, they map text and images to distant regions in the representation space, limiting retrieval performance. To bridge this modality gap, we propose a two-step approach. First, we transform the text query into a visual query using a generative diffusion model. Then, we estimate image-to-image similarity with a vision model. Additionally, we introduce an aggregation network that combines multiple generated images into a single vector representation and fuses similarity scores across both query modalities. Our approach leverages advancements in vision encoders, VLMs, and text-to-image generation models. Extensive evaluations show that it consistently outperforms retrieval methods relying solely on text queries. Source code is available at: \href{https://github.com/faixan-khan/cletir}{https://github.com/faixan-khan/cletir} 
    \end{abstract}

\section{Introduction}
\label{sec:intro}
This work explores category-level image retrieval using a textual query that names or describes a semantic class, aiming to retrieve all images depicting objects of the specified class. This task is particularly crucial in open-world scenarios, where systems must handle arbitrary categories. It has practical applications in navigating large-scale digital image archives and visual datasets, such as computer vision training sets containing millions or billions of images. Moreover, such retrieval serves as a fundamental component in more complex computer vision pipelines~\cite{susx,liu23react,stojnic24zlap}.

Despite its significance, category-level text-to-image retrieval has received limited attention in prior research. Existing approaches often rely on text-based image crawling from the web, followed by training an image classifier~\cite{chatfield2015journal}, utilizing handcrafted representations~\cite{chatfield2012visor} or early deep models~\cite{chatfield2014gpuvisor}. In contrast, more general text-to-image retrieval tasks have been more extensively studied~\cite{huang2018learning, chen2021learning, faghri2017vse++, li2019visual, chun2021probabilistic, song2019polysemous}, though they typically depend on domain-specific training and lack true open-vocabulary capabilities. The emergence of CLIP~\cite{radford2021clip} revolutionized the field by enabling training-free, open-world retrieval~\cite{mode,wang2025diffusion}.

Building on advancements in vision-and-language models (VLMs)~\cite{radford2021clip,li2022blip,jia2021scaling}, we revisit category-level text-to-image retrieval. Leveraging VLMs makes this task straightforward, \ie, obtaining a text representation of the query and performing Euclidean search within the visual representations of database images. We evaluate this approach across multiple benchmarks.
Despite their strong performance, VLMs exhibit a known modality gap, where text and image representations remain well-separated in the feature space~\cite{liang2022mind,schrodi2024twoeffects,shi2023towards}. Inspired by prior work~\cite{zhang23cafo,wysoczanska2023clipdinoiser,iscen2023retrieval} demonstrating the effectiveness of intra-modal operations over cross-modal ones, we propose bridging this gap by mapping text to images and subsequently performing image-to-image comparisons. To achieve this, we transform the text query into an image query using a text-to-image Generative Diffusion-based Model (GDM)~\cite{sd,add,flux}. Instead of relying on the VLM's vision encoder, we employ a foundational Vision Model (VM) for image-to-image similarity estimation. By properly fusing the multiple queries from both modalities, our approach achieves consistent improvements over the text-only baseline across fifteen benchmarks.

\section{Related Work}
\label{sec:related_work}
In this section, we review the related work on text-to-image retrieval, the use of VLMs in visual recognition tasks, the synergy between VLMs and VMs for cross-modal recognition, and the use of image generation models as free training data generators.

\myparagraph{Text-to-Image Retrieval} is a cross-modal retrieval task aimed at finding images relevant to text descriptions such as captions. Traditional methods~\cite{huang2018learning, chen2021learning, faghri2017vse++, li2019visual, chun2021probabilistic, song2019polysemous} rely on domain-specific training and lack open-vocabulary generalization. Some approaches improve model architectures~\cite{huang2018learning, chen2021learning, li2019visual}, others propose new loss functions~\cite{chun2021probabilistic, faghri2017vse++}, or design alternative embedding representations~\cite{chun2021probabilistic, song2019polysemous}. Category-level retrieval~\cite{chatfield2012visor, chatfield2014gpuvisor, chatfield2015journal, vakhitov2016internet} is a special case where the query defines a category rather than a detailed caption. These works use Google Image Search to retrieve representative images and perform image-based retrieval. In contrast, we leverage modern foundation models for category-level retrieval.

\myparagraph{VLMs for Image Recognition Tasks}
Vision-Language Models (VLMs)\cite{radford2021clip,jia2021scaling,cherti2023reproducable,siglip}, trained on large image-text datasets\cite{schuhmann2022laion,kakaobrain2022coyo-700m,metaclip,gadre2023datacomp}, achieve strong performance on various vision tasks. CLIP~\cite{radford2021clip} and SigLIP~\cite{siglip} excel at zero-shot classification, further improved by methods like Tip-Adapter~\cite{zhang21tip}, SuS-X~\cite{susx}, and CaFO~\cite{zhang23cafo}. CoCa~\cite{yu2022coca} and Florence~\cite{yuan2021florence} extend VLMs to video action recognition. 
Additionally, the advent of VLMs~\cite{li2022blip,sun2024evaclip18b,radford2021clip} opened the possibilities for performing text-to-image retrieval in the open-vocabulary setting without the necessity for domain-specific training. We take advantage of these capabilities and propose a way to utilize VLMs for category-level retrieval.

\myparagraph{VLMs and VMs for Cross-modal Tasks}
Although VLMs perform well on cross-modal tasks, their embeddings are less effective for intra-modal tasks due to the modality gap~\cite{liang2022mind,shi2023towards,schrodi2024twoeffects}. To address this, several works incorporate Vision Models (VMs) for intra-modal components. CaFO~\cite{zhang23cafo} uses a self-supervised VM to boost few-shot classification. CLIP-DINOiser~\cite{wysoczanska2023clipdinoiser}, ProxyCLIP~\cite{lan2024proxyclip}, LaVG~\cite{kang2024defense}, and LPOSS~\cite{stojnic2025lposs} leverage DINO~\cite{caron2021dino} for patch-level relationships, improving VLM-based semantic segmentation. Additionally, previous works like~\cite{tong2024eyes,shen2024longvu,tong2024cambrian} show VMs can enhance multi-modal LLMs. Inspired by this, we use DINOv2~\cite{oquab2023dinov2} to extract embeddings for image-to-image retrieval.

\myparagraph{Generative Models as Training Data Generators}
The emergence of realistic image generation models~\cite{sd,saharia2022imagenet,ramesh2021dalle} has prompted interest in their use for image recognition tasks. Sariyildiz \etal~\cite{sariyildiz2023fakeit} show that models trained on synthetic ImageNet data transfer as well as those trained on real data. Azizi \etal~\cite{azizi2023synthetic} further demonstrate improved performance when combining real and synthetic data. For segmentation, FreeMask~\cite{yang2023freemask} and DatasetDiffusion~\cite{nguyen2023datasetdiffusion} generate synthetic training images. Compared to these works, we investigate that if synthetic images can be used during inference for category-level image retrieval by using synthetic data to enrich the given text queries.

\section{Preliminaries}
\label{sec:pre}

\myparagraph{Task Formulation}
We study category-level text-to-image retrieval, where the goal is to retrieve images based on \textit{category or class labels}. Unlike image-to-image retrieval~\cite{gordo2017beyond,an2023unicom,ermolov2022hyperbolic}, which retrieves images similar to a query image, this task retrieves images relevant to a query text. In contrast to instance-level retrieval~\cite{chen2023retrievalsurvey}, where relevance is based on depicting the same specific object, here it depends on belonging to the same semantic class. This cross-modal task takes a class name as input (e.g., ``dog'') and retrieves all images depicting that category. We explore three query types: a class name, a class description, and both jointly.

\myparagraph{VLM}
Vision-Language Models (VLMs)\cite{radford2021clip,jia2021scaling,li2022blip} are well-suited for cross-modal tasks. These models consist of a textual encoder $f$ that maps text $y$ to its representation vector $f(y)$, and a vision encoder $g$ that maps image $x$ to its representation vector $g(x)$ on a shared representation space. These models are trained on large image-caption datasets like WIT-400M\cite{radford2021clip} and LAION~\cite{schuhmann2022laion} via contrastive learning between $f(y)$ and $g(x)$. We primarily use CLIP~\cite{radford2021clip}, but also report results on EVA02-CLIP~\cite{fang2022eva}, MetaCLIP~\cite{metaclip}, SigLIP~\cite{siglip}, and OpenCLIP~\cite{openclip}. During test time, with the use of a VLM, the similarity between words and images is estimated straightforwardly.

\myparagraph{GDM} Generative Diffusion-based Models~\cite{sd,saharia2022imagenet, add} are a class of large generative models that function on the principle of denoising diffusion to generate images. 
During inference, starting from a noisy input, the backward diffusion process is run to obtain a denoised image. 
The text-to-image GDMs use textual input conditioning to guide the generation process. 
Instead of starting from just a noisy 
input,
the textual representation~\cite{radford2021clip,vaswani2017attention} of text $y$ forms an additional input.
This work primarily uses Stable Diffusion (SD)\cite{sd}, Single-step Distilled Diffusion\cite{add}, and FLUX~\cite{flux}. 

\section{Method}
\label{sec:method}

\begin{figure*}[t]
\begin{center}
\includegraphics[width=0.9\textwidth]{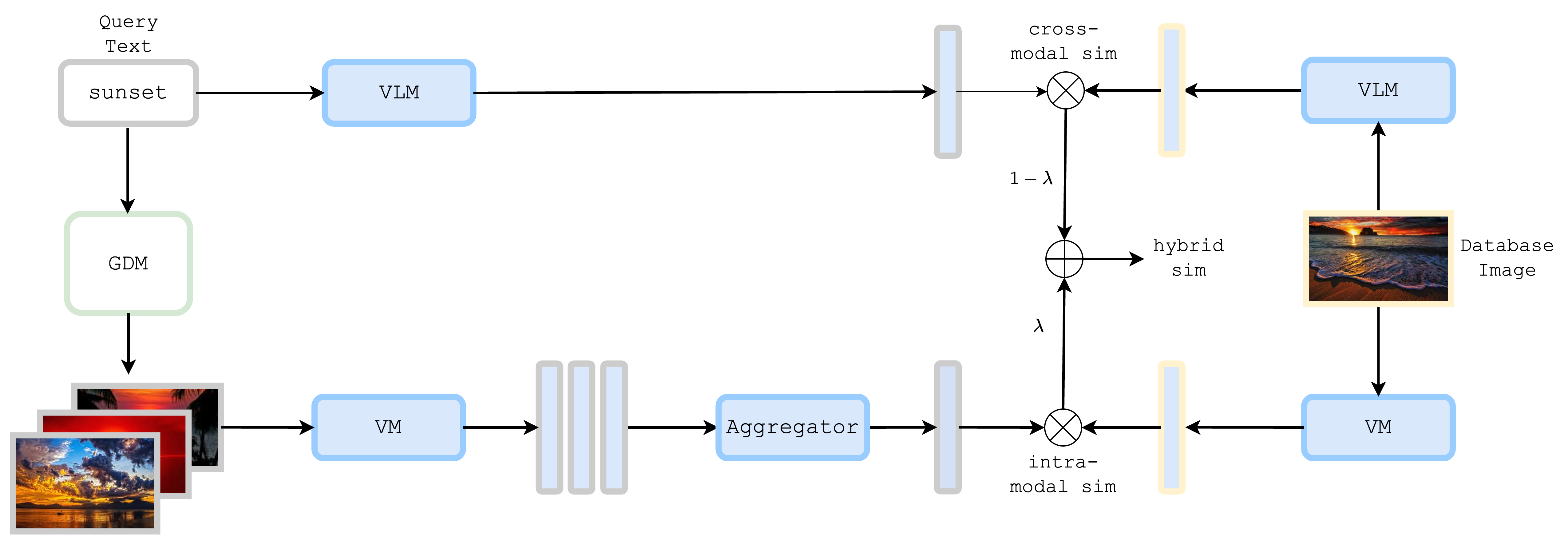}
\vspace{5pt}
\caption{
Overview of the proposed method.
The text query is used as input to generate multiple image queries using a Generative Diffusion-based Model. Text query, database images, and the Vision-Language Model are used to estimate the cross-modal text-query-to-database-image similarity. Image query representations extracted with a Vision Model get aggregated via an aggregator module, then used to estimate intra-modal image-query-to-database-image similarity with the database images.  The hybrid similarity is a weighted average of the two separate similarities based on a learned $\lambda$.
\label{fig:method}
\vspace{-20pt}
}
\end{center}
\end{figure*}

Given text query $y$, we describe the proposed approach enabling the similarity estimation between $y$ and each database (db) image $z$.
An overview is shown in Figure~\ref{fig:method}.
The vanilla approach is to perform cross-modal retrieval via computing the text query to db image similarity via a VLM.

\subsection{Generating Image Queries}
Cross-modal retrieval suffers from the modality gap~\cite{liang2022mind, iscen2023retrieval} due to insufficient alignment between textual and visual representations in the pre-training stage. 
Lack of visual context in the form of query images hinders the application of standard intra-modal retrieval that is shown to be superior to cross-modal retrieval~\cite{iscen2023retrieval}. 
We bridge this gap by generating image queries using a pre-trained text-to-image GDM. 
We use $y$ to prompt the diffusion model using a template \texttt{``A photo of a} $[y]$ \texttt{''}. We generate a set of $k$ visual queries $x=\{x_1,\ldots, x_k\}$
per text query by varying the seed value to the diffusion model.
Therefore, we are now given one text query and several image queries to perform the retrieval; the query is bi-modal, while the visual modality contains multiple queries.
In our experiments, we explore the option of using multiple generative models~\cite{sd, add, flux}, both for training and testing, to capture complementary aspects of a class and to cancel each other's mistakes.

\subsection{Similarity Estimation} 
\label{subsec:sim}
\textbf{Cross-Modal Similarity} 
The cross-modal similarity between image $z$ and text query $y$ is estimated with the use of a VLM via a simple dot product
 $s^c = g(z)^\top f(y).$

\noindent\textbf{Intra-modal Similarity} Given the db image $z$ and generated images $x_i$, the similarity between query $y$ and db image $z$ can be estimated indirectly through intra-modal similarity between $z$ and $x_i$.

Instead of using the visual encoder $g$ of the VLM for this task, we assume access to the encoder $h$ of a Vision Model (VM) that maps images to a $d$ dimensional descriptor space. 
Estimating intra-modal similarity is the task VMs are originally optimized for, in contrast to VLMs, whose training objective only includes cross-modal terms and not intra-modal.

We obtain the db image representation $h(z)$ and the representation for the generated query images given in set $x = \{h(x_1), \ldots, h(x_k))\}$.
We propose to first aggregate the $k$ representations and then compute the similarity to the db image.
The aggregation is performed by the function $a: \real^{d \times k} \rightarrow \real^d$.
Then, $a(x)$ is used to compute the intra-modal similarity between query and db image given by
$s^i = h(z)^\top a(x) = h(z)^\top a(\{h(x_1), \ldots, h(x_k)\}).$
%
For $k=1$ there is no need for aggregation; we simply set $a(x)=h(x_1)$.

\noindent\textbf{Hybrid Similarity}
The hybrid similarity is a weighted combination of intra-modal and cross-modal similarities $s = (1 - \lambda) s^{c} + \lambda s^i $
with $\lambda \in [0,1]$. For extreme values 0 and 1, the similarity is equivalent to the \emph{cross-modal} only or \emph{intra-modal} only, respectively. We refer to those as \emph{text-only} and \emph{image-only} approaches, respectively, as well as \emph{hybrid} when $\lambda \in (0,1)$.

\noindent\textbf{Aggregator architecture}
A baseline approach for $a$ is to perform average pooling, denoted by $a_m$, which we evaluate in the experiments.
Instead, we propose a learnable parametric model $a_{\theta}$ as the aggregator, whose parameters are learned directly from data. 

The aggregator design relies on a sequence of simple self-attention layers that have distinctive differences from the standard practice.
They use symmetric attention (query and key projections are the same), and value projections are identity functions. Additionally, the CLS token at the input of the first layer is not learnable and is set equal to the average pooling of all other input tokens. Those other input tokens are fed as input to all attention layers without any modification, with the CLS being the only one affected by the attention processing.  An overview of the aggregator architecture is demonstrated in the supplementary material.

Concretely, the aggregator is a sequence of attention layers with the $l$-th layer $A_l: \real^{d\times (k+1)} \rightarrow \real^{d\times (k+1)}$ given by  $A_l(u) = \softmax\left(\phi_{\theta_l} (u)^\top \phi_{\theta_l} (u)\right) u^\top,$
%
where $\phi_{\theta_l}: \real^d \rightarrow \real^d$ is a linear transform applied independently per column, subscript $\theta_l$ indicates that this function is parametric with learnable parameters.
Now, let $u_1 = [u_{1,1}, \ldots, u_{1,k}, u_{1,k+1}] \in \real^{d\times (k+1)}$ be the input of the first layer, whose tokens (columns) are equal to $u_{1, i}=h(x_i)$ for $i=1,\ldots,k$ and $u_{1,k+1}=a_m(x)$.
The last token can be seen as corresponding to the CLS token. 
The output of the first layer is $\hat{u}_1 = A_1(u_1)$, with tokens (columns) denoted by $\hat{u}_{1,i}, i=1,\ldots, k+1$. 

The input of the $l$-th layer is formed by concatenating the $k$ inputs of the first layer and CLS output token of layer $l-1$, \ie $u_l=[u_{1,1}, \ldots,  u_{1,k}, \hat{u}_{l-1,(k+1)}]$. The same process is repeated across all $L$ layers.
The final output vector of the aggregator is the CLS token of the last attention layer, \ie $\hat{u}_{L,k+1}$.
The learnable parameters $\theta$ of the aggregator are the parameters of all linear transforms, one per layer.

The proposed sequence of attention layers performs a more intuitive operation than standard attention or transformers, which include feed-forward layers and skip connections.
By setting value projections to identity, the output representation space remains unchanged; \ie, the architecture performs only weighted mean operation with context-dependent weights.
Thus, the final representation stays compatible with the db image's representation space. Viewing attention layers as in-context mappings, we feed the same input tokens to all layers, iteratively transforming the CLS token in the context of the input vectors being aggregated.

\subsection{Training: Aggregator and Modality Balance}

The role of the aggregator function $a_{\theta}$ is to robustly aggregate the query representations such that relevant database images (positives) are ranked higher than irrelevant images (negatives). To learn the aggregator but also $\lambda$ (VLM and VM are frozen), we generate a large synthetic training set. We prompt two GDMs, SD~\cite{sd} and FLUX~\cite{flux}, using category names from the OpenImages text corpus~\cite {openimages}. To simulate a zero-shot setup where we test on unseen classes, we remove classes from this corpus that match those of the benchmark datasets\footnote{We use CLIP to find the nearest neighbor of every class from the fifteen test benchmarks and remove all of them.}. This process provides us with a training set of about 390k images with ten images per class per GDM, whose label is considered the class used as a prompt.

To construct a training batch, we sample $M$ classes, where each class name forms the text query, and then randomly sample $N+1$ generated images from this class to use as image queries ($N$) and as a positive (1). 
A negative image per class is chosen to be the hardest negative among the $M-1$ positives of other classes that are already sampled. The hardness is estimated using the hybrid similarity, taking into account the current status of the model. 
Contrastive loss is computed taking into account the hybrid similarity between query and positive image ($s_\text{pos}$) and query and negative image ($s_{\text{neg}}$) given by 

\begin{equation}
\ell = -\log\left(\frac{\exp(s_\text{pos})/\tau}{\exp(s_\text{pos})/\tau + \exp(s_\text{neg})/\tau}\right).
\end{equation}

\begin{wrapfigure}[13]{b}{0.39\textwidth}
\pgfplotsset{compat=1.17}
\begin{tikzpicture}[scale=0.7]
\begin{axis}[
    ybar,
    bar width=8pt,
    width=7cm, height=5cm,
    bar width=8.2pt,  
    xlabel={similarity},
    ylabel={},
    xtick align=center,
    grid=both,
    ymin=0,
    xmin=-0.15, xmax=1.05,
    xtick={0,  0.5,1.0},
    xticklabel style={/pgf/number format/fixed},
    legend pos=north east,
    legend entries={Syn2Real, Syn2Syn}
]

\addplot[
    ybar,
    fill=orange!70,
    opacity=0.5,
    draw=black
] coordinates {
    (-0.1, 314) (-0.05, 12268) (0.0, 75987) (0.04, 121008)
    (0.09, 119639) (0.15, 107387) (0.20, 96377) (0.26, 89818) 
    (0.31, 82104) (0.37, 76968) (0.43, 72412) (0.48, 69116) 
    (0.54, 64424) (0.59, 59666) (0.65, 55692) (0.70, 51202) 
    (0.76, 41750) (0.81, 29838) (0.87, 14335) (0.93, 1790)
};

\addplot[
    ybar,
    fill=blue!70,
    opacity=0.5,
    draw=black
] coordinates {
    (-0.11, 139) ( -0.06, 6374) (-0.02, 64915) (0.03, 122536) 
    (0.09, 117135) (0.14, 113712) (0.20, 115347) (0.26, 118724) 
    (0.31, 120540) (0.37, 122294) (0.43, 121948) (0.48, 121445) 
    (0.54, 122848) (0.60, 122782) (0.65, 121966) (0.71, 119135) 
    (0.76, 113680) (0.82, 99727) (0.88, 73077) (0.94, 21376)
};

\end{axis}
\end{tikzpicture}

\vspace{2pt}
  \caption{Image-to-image similarity distributions for synthetic-to-real or synthetic-to-synthetic images.}
  \label{fig:simdistr}
\end{wrapfigure}
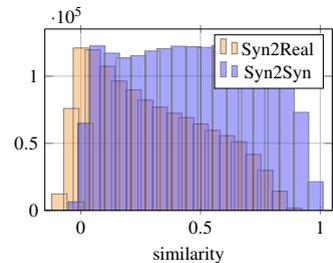

\looseness=-1
We set the parameter $\lambda$ to be learnable and observe that back-propagation needs to be performed only through the intra-modal similarity term. This is due to the fact that the encoder models are frozen. 
We come up with the following empirical trick, which effectively increases the performance, and is motivated by the following two observations.
We train with only synthetic images, but during testing, the similarity between real and synthetic images is computed. 
There is a discrepancy between the synthetic-to-real and the synthetic-to-synthetic image similarities, as shown in Figure~\ref{fig:simdistr}.
Therefore, we set the cross-modal similarity for positives to be fixed to 1 (the maximum similarity) as if we are dealing with perfect text-to-image similarity for the positives. 
The cross-modal similarity for negatives is properly estimated. Setting it to a fixed value would result in a trivial solution of $\lambda=0$, making the aggregator irrelevant.

    \vspace{-5mm}
\section{Experiments}
\label{sec:exp}
\begin{table*}[t]
\centering
\resizebox{\textwidth}{!}{
{
\footnotesize
\begin{tabular}{lrrrrrrrrrrrrrrlr}
& \rotatebox{90}{ImageNet} & \rotatebox{90}{DTD} & \rotatebox{90}{Stanford Cars}     & \rotatebox{90}{SUN397}   & \rotatebox{90}{Food}     & \rotatebox{90}{FGVC Aircraft} & \rotatebox{90}{Oxford Pets}     & \rotatebox{90}{Caltech101} & \rotatebox{90}{Flowers 102}  & \rotatebox{90}{UCF101}      & \rotatebox{90}{Kinetics-700}    & \rotatebox{90}{RESISC45}      & \rotatebox{90}{CIFAR-10}      & \rotatebox{90}{CIFAR-100}     & \rotatebox{90}{Places-365} & \rotatebox{90}{Average} \\
\midrule
image-only (D) & 63.9 & 38.0 & 30.2 & 51.6 & 75.3 & 11.1 & 84.8 & 87.5 & 63.0 & 59.4 & 28.5 & 43.9 & 65.6 & 75.1 & 27.4 & 53.7\\
\midrule
query text - Class \\
\midrule
text-only (C) & 64.9 & 41.9 & 64.4 & 54.4 & 88.3 & 28.4 & 88.0 & 90.8 & 76.4 & 66.3 & 36.2 & 64.3 & 88.4 & 61.7 & 25.8 & 62.7\\
image-only (C) & 32.1 & 25.2 & 35.8 & 37.4 & 62.3 & 14.3 & 47.0 & 73.8 & 42.6 & 46.7 & 14.9 & 42.2 & 78.8 & 51.6 & 17.0 & 41.4\\
text (C)+image (C) & 51.3 & 36.5 & 53.1 & 50.7 & 82.2 & 21.8 & 73.7 & 85.3 & 67.6 & 59.3 & 26.7 & 58.1 & 89.7 & 65.6 & 23.7 & 56.4\\
text(C)+image(D)& 73.2 & 43.9 & 37.0 & 56.9 & 80.9 & 15.6 & 87.9 & 91.3 & 72.8 & 65.0 & 33.4 & 57.2 & 87.0 & 82.1 & 30.6 & 61.0\\
Tip-adapter  (C,D)~\cite{zhang21tip} & 73.0 & 50.0 & 51.1 & 61.0 & 86.8 & 22.9 & 90.4 & 92.9 & 78.7 & 70.5 & 38.1 & 64.2 & \textbf{93.2} & \textbf{83.5} & \textbf{32.2} & 65.9\\
\rowcolor{blond}
Ours  (C,D) & \textbf{73.8} & \textbf{50.1} & \textbf{67.1} & \textbf{62.4} & \textbf{90.7} & \textbf{29.0} & \textbf{91.0} & \textbf{93.0} & \textbf{79.6} & \textbf{72.5} & \textbf{40.8} & \textbf{66.6} & 90.4 & 80.3 & 31.9 & \textbf{67.9} \\
\midrule
text-only (S) & 72.4 & 49.9 & \textbf{89.2} & 59.8 & \textbf{93.1} & \textbf{45.6} & 92.3 & 94.6 & 83.5 & 74.3 & 42.0 & 63.5 & 93.6 & 72.3 & 28.7 & 70.3\\
image-only (S) & 38.7 & 31.6 & 56.4 & 41.9 & 65.5 & 17.4 & 61.7 & 79.4 & 56.3 & 51.9 & 22.4 & 51.2 & 80.1 & 60.5 & 20.7 & 49.0\\
text (S)+image (S)& 60.1 & 42.0 & 76.1 & 55.0 & 85.9 & 32.2 & 82.1 & 89.8 & 78.0 & 65.2 & 33.2 & 60.6 & 90.5 & 72.9 & 27.5 & 63.4\\
text(S)+image(D) & 70.8 & 46.7 & 49.6 & 57.9 & 82.6 & 23.6 & 89.0 & 92.5 & 77.9 & 67.6 & 34.9 & 57.8 & 91.1 & 84.3 & 31.3 & 63.8\\
Tip-adapter  (S,D)~\cite{zhang21tip} & 75.2 & 51.9 & 68.6 & 61.5 & 88.3 & 33.8 & 91.4 & \textbf{95.6} & 82.3 & 73.1 & 39.8 & 62.7 & \textbf{94.4} & \textbf{86.3} & 33.0 & 69.1\\
\rowcolor{blond}
Ours (S,D) & \textbf{77.4} & \textbf{54.6} & 88.2 & \textbf{64.4} & 92.9 & 44.1 & \textbf{92.9} & 95.5 & \textbf{84.0} & \textbf{77.1} & \textbf{44.0} & \textbf{65.1} & 94.0 & 85.4 & \textbf{33.3} & \textbf{72.9}\\
\midrule
query text - Description-only \\
\midrule
text-only (C) & 34.4 & 23.5 & 8.5 & 38.1 & 67.9 & 12.6 & 21.3  &  76.4 & 35.3 & 48.0 & 18.5 & 42.4 & 73.0 & 42.3 & 16.2 & 37.2 \\
\cellcolor{blond}Ours (C,D) & \cellcolor{blond}\textbf{42.4} & \cellcolor{blond}\textbf{29.5} & \cellcolor{blond}\textbf{10.1} & \cellcolor{blond}\textbf{45.5} & \cellcolor{blond}\textbf{71.4} & \cellcolor{blond}\textbf{13.6} & \cellcolor{blond}\textbf{32.5} & \cellcolor{blond}\textbf{82.2} & \cellcolor{blond}\textbf{41.5}  & \cellcolor{blond}\textbf{57.2} & \cellcolor{blond}\textbf{24.8} & \cellcolor{blond}\textbf{47.8} & \cellcolor{blond}\textbf{79.0} & \cellcolor{blond}\textbf{55.5} & \cellcolor{blond}\textbf{21.3} & \cellcolor{blond}\textbf{43.6} \\
text-only (S) & 46.3 & 32.7 & \textbf{12.3} & 42.5 & \textbf{79.8} & 14.3 & 37.1 & 80.0 & 43.5  & 53.4 & 26.8 & 47.5 & 75.3 & 57.4 & 20.3 & 44.6 \\
\cellcolor{blond}Ours (S,D)& \cellcolor{blond}\textbf{49.9} & \cellcolor{blond}\textbf{35.3} & \cellcolor{blond}12.1 & \cellcolor{blond}\textbf{47.3} & \cellcolor{blond}77.7 & \cellcolor{blond}\textbf{14.9} & \cellcolor{blond}\textbf{44.2} & \cellcolor{blond}\textbf{85.1} & \cellcolor{blond}\textbf{48.3}  & \cellcolor{blond}\textbf{61.4} & \cellcolor{blond}\textbf{30.1} & \cellcolor{blond}\textbf{50.5} & \cellcolor{blond}\textbf{78.9} & \cellcolor{blond}\textbf{63.8} & \cellcolor{blond}\textbf{23.9} & \cellcolor{blond}\textbf{48.2} \\

\midrule
query text - Description + Class \\
\midrule
text-only (C) & 65.5 & \textbf{47.0} & 66.2 & 53.6 & 90.5 & \textbf{30.9} & 85.2  & 93.4 & 80.4 & 66.6 & 31.4 & 54.6 & 94.4 & 70.0 & 24.5 & 63.6 \\
\cellcolor{blond}Ours (C,D) & \cellcolor{blond}\textbf{71.8} & \cellcolor{blond}46.8 & \cellcolor{blond}\textbf{66.8} & \cellcolor{blond}\textbf{58.1} & \cellcolor{blond}\textbf{90.8} & \cellcolor{blond}29.6 & \cellcolor{blond}\textbf{91.2} & \cellcolor{blond}\textbf{95.4} & \cellcolor{blond}\textbf{82.2}  & \cellcolor{blond}\textbf{73.0} & \cellcolor{blond}\textbf{37.7} & \cellcolor{blond}\textbf{59.3} & \cellcolor{blond}\textbf{95.5} & \cellcolor{blond}\textbf{81.3} & \cellcolor{blond}\textbf{29.1} & \cellcolor{blond}\textbf{67.2} \\
text-only (S) & 73.1 & \textbf{51.4} & \textbf{88.5} & 57.4 & \textbf{93.6} & \textbf{48.3} & 93.5 & 95.6 & 87.9 & 73.5 & 40.3 & 59.7 & 95.4 & 77.5 & 28.7 & 70.9 \\
\cellcolor{blond}Ours (S,D)& \cellcolor{blond}\textbf{76.0} & \cellcolor{blond}50.8 & \cellcolor{blond}85.5 & \cellcolor{blond}\textbf{59.8} & \cellcolor{blond}92.2 & \cellcolor{blond}45.3 & \cellcolor{blond}\textbf{94.0} & \cellcolor{blond}\textbf{96.4} & \cellcolor{blond}\textbf{89.1} & \cellcolor{blond}\textbf{77.4} & \cellcolor{blond}\textbf{42.7} & \cellcolor{blond}\textbf{63.1} & \cellcolor{blond}\textbf{96.3} & \cellcolor{blond}\textbf{84.7} & \cellcolor{blond}\textbf{31.2} & \cellcolor{blond}\textbf{72.3} \\
\bottomrule

\end{tabular}
}}

\vspace{3mm}
\caption{Retrieval performance on 15 benchmark datasets using different query types - class names, class description that does not include class name, and both jointly as ``[class name]: [current-description]. For each case, encoders for cross-modal (text-to-image) and intra-modal (image-to-image) similarity are also presented.
\textbf{C: CLIP, D: DINOv2, S: SigLIP}.
\label{tab:main_figure}
}
\end{table*}

\subsection{Experimental Setup}
\label{subsec:exp_setup}
We perform experiments across 15 datasets: ImageNet~\cite{russakovsky2015imagenet}, Stanford Cars~\cite{6755945}, Describable Textures Dataset (DTD)~\cite{Cimpoi_2014_CVPR}, Scene UNderstanding (SUN397)~\cite{5539970}, Food101~\cite{food}, FGVC Aircraft~\cite{maji13finegrained}, Oxford
Pets~\cite{Parkhi2012CatsAD}, Caltech101~\cite{FeiFei2004LearningGV}, Flowers 102~\cite{4756141}, UCF101~\cite{soomro2012ucf101},
Kinetics-700~\cite{carreira2022short}, Remote Sensing Image Scene Classification (RESISC45)~\cite{7891544}, CIFAR-10~\cite{Krizhevsky2009LearningML}, CIFAR-100~\cite{Krizhevsky2009LearningML},
and Places365~\cite{zhou2017places}. 
For the two video datasets, we follow the same procedure as done in~\cite{radford2021clip} by extracting the
middle frame of the video. 
We report the scores using the official test sets as an image database for datasets that provide them, and for datasets that do not report an official split, we use the splits described in Zhou \etal~\cite{zhou2022cocoop}. 
As there is no conventional split for RESISC45, we perform the retrieval using all images as the database.

We use CLIP~\cite{radford2021clip} as a VLM with its visual and textual encoders as $f$ and $g$, respectively. We use DINOv2~\cite{oquab2023dinov2} as a VM and encoder $h$. All the models are used with the ViT-L14 backbone~\cite{dosovitskiy2021image}. 
Results for other backbones~\cite{metaclip, sun2024evaclip18b, openclip} are reported in the supplementary.

To generate image queries for testing, we leverage Stable-Diffusion-Turbo~\cite{add} (SD-Turbo), Stable-Diffusion 2.1~\cite{sd} (SD), FLUX~\cite{flux}. The last two are the ones used for the training, too. We generate $5$ images per text query $y$ by changing the seed values per GDM.
The training is performed for $k=5$, while during testing, we use the aggregator for any number of input images since the attention architecture allows it.
We use $L=2$ attention layers.
Unless otherwise stated, we test with $5$ image queries from SD.

\subsection{Results with Class Name as Query}
\label{subsec:res}
We compare with the baseline approaches using text-only query ($\lambda=0$), image-only query ($\lambda=1$) with five images, and text+image with equal modality importance ($\lambda=0.5$), which all use the average vector aggregator.
We additionally compare it to the Tip-adapter~\cite{zhang21tip} similarity, which considers a text query and multiple image queries, even though it was proposed for zero-shot classification. We tune its hyperparameters based on grid search and performance evaluation for retrieval on our training set.

Table~\ref{tab:main_figure} summarizes results from two CLIP variants: original CLIP~\cite{radford2021clip} and SigLIP~\cite{siglip}.   The vanilla text-only approach is a strong starting point.  Despite DINOv2 performing much better for the image-only baseline, the image queries solely are inferior to the text query, which better represents the ``mode'' of the class.
Neither of the hybrid baselines manages to surpass the text-only approach.
The proposed approach performs best and outperforms Tip-adapter~\cite{zhang21tip}, which fails to beat the baseline for the stronger encoder, \ie SigLIP.

To compare with the only previous approaches that perform category-level image retrieval, we evaluate on PASCAL VOC~\cite{pascal-voc-2007} and compare with the reported numbers.
The reproducibility of these methods is not as straightforward as they rely on crawling images from Google Image Search.
Our proposed method achieves a Mean Precision of 96.1 at the top 100 ranks on the test split. This is higher than 92.1~\cite{chatfield2014gpuvisor} achieved in prior work.
Note that contributions in that line of research, such as the on-the-fly training of a binary classifier, are complementary to the methods explored in this work.

\begin{wraptable}[5]{b}{0.5\textwidth} 
  \centering
  \vspace{-8pt}
  \centering
\small
\resizebox{0.5\textwidth}{!}{
\begin{tabular}{lrrrrrrrrrlr}
\hline
\multicolumn{11}{c}{SD query images only}\\
\hline
$k$ & 1 & 2 & 3 & 4 & 5 & 6 & 7 & 8 & 9 & 10 \\
mAP & 67.1 & 67.6 & 67.8 & 67.9 & 68.0 & 68.0 & 68.0 & 68.0 & 68.0 & 68.0 &  \\
\hline
\multicolumn{11}{c}{SD+SD-Turbo+Flux query images}\\
\hline
$k$ & \multicolumn{2}{c}{1+1+1} & \multicolumn{2}{c}{2+2+2} & \multicolumn{2}{c}{3+3+3} & \multicolumn{2}{c}{4+4+4} & \multicolumn{2}{c}{5+5+5} \\
mAP & \multicolumn{2}{c}{68.4} & \multicolumn{2}{c}{68.8} & \multicolumn{2}{c}{68.9} & \multicolumn{2}{c}{69.0} & \multicolumn{2}{c}{69.0} \\
\hline
\end{tabular}
}

  \vspace{5pt}
\caption{Impact of more and diverse queries.}
\label{tab:sat}
\end{wraptable} 

\noindent\textbf{More and diverse generated image queries}:
We evaluate our approach with images from all generators rather than relying on a single one. 
As shown in Table~\ref{tab:sat}, this leads to a clear performance boost, demonstrating that leveraging multiple generators can enhance the overall results.
Performance using one generator saturates after $k=5$, but after $k=10$ for 3 generators.

\subsection{Results with Class Description as Query}
\label{subsec:dbr}
\noindent\textbf{Using only class description} We introduce a novel task for image retrieval based on querying only by the category description. We consider a setup where users want to retrieve images of objects in a situation where they can not recall the exact name of the category but can describe the object's looks or properties. To address this challenge, we establish a benchmark by prompting an LLM~\cite{openai-gpt} with the class category $y$. We prompt the LLM to generate coherent sentences $y' = LLM(y)$ describing the category $y$ without explicitly mentioning the category $y$.  The description for class ``airplane'' is, ``a flying vehicle made of metal, equipped with wings is commonly used for air travel.'' We create this benchmark for all fifteen datasets reported in~\cref{subsec:exp_setup} spanning fine-grained and coarse-grained tasks.

\begin{figure}[t]
  \centering
  \includegraphics[width=\textwidth]{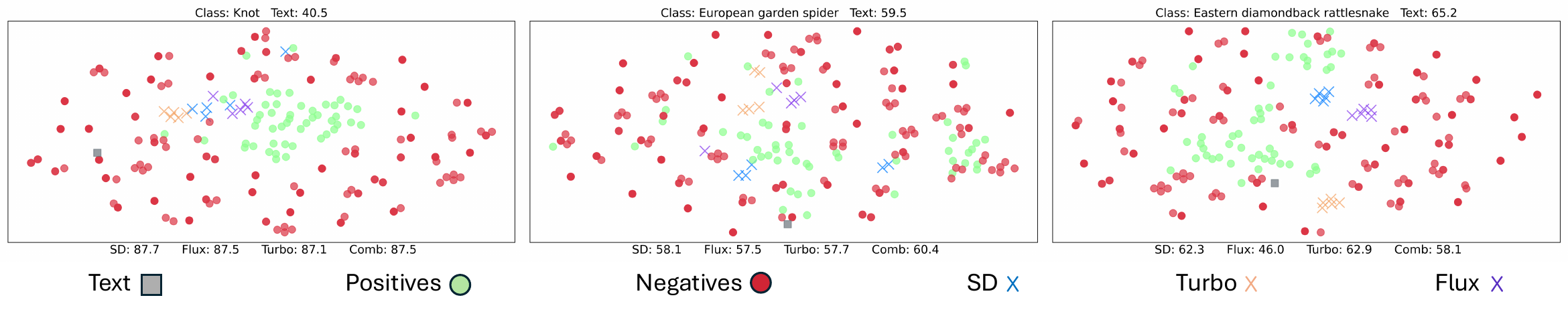}
  \vspace{-5pt}
  \caption{t-SNE visualization of features for text and image queries from 3 generators, the positive db images, and the top-ranked negatives. Performance is reported for the text-only baseline, our approach using one(SD, Turbo, Flux) of the generators (5 images), and all(comb) three generators combined(15 images).}
  \label{fig:analyses}
  \vspace{-5pt}
\end{figure}

To generate image queries, we adopt a procedure similar to the one used for handling category-level image retrieval. However, rather than directly using the class label $y$ (which is unavailable in this setting), we instead utilize the generated description $y'$ as the input to the GDM. 

\noindent\textbf{Using class name with description} In this task, we combine the class label $y$ and its description $y'$ as ``$y$ : $y'$''. Image queries are generated as before. The input to GDM is the combined query of class label $y$ and the generated description $y'$,\ie  ``$y$ : $y'$''.

We then follow our approach to perform retrieval for both tasks. Using description only for retrieval is quite challenging as there is no mention of the class name. For example, the description for class ``pink primrose'' is, ``the flower's petals are a deep pink hue, with a bright yellow center.''. As shown in Table~\ref{tab:main_figure}, our proposed method significantly improves performance across both tasks for CLIP~\cite{radford2021clip} and SigLIP~\cite{siglip}. For description-only based retrieval, we improve the average performance over CLIP~\cite{radford2021clip} by 6.4\% and by 3.6\% over SigLIP~\cite{siglip}. Adding the description to the class name enhances the text-only performance for both CLIP and SigLIP. However, our approach is still able to improve on both models. This highlights the effectiveness of the visual information generated by the GDM, which, when aggregated by our module, provides important contextual cues to enhance the retrieval process.

\subsection{Ablation Study}
\label{sub:ablation}

\Cref{tab:ablation} shows the impact of different architectural and training choices on performance. Below, we detail the effects of each design choice. All variants use 5 SD query images.

\begin{wraptable}[11]{r}{0.45\textwidth} 
  \centering
  \vspace{-3pt}
  
\centering{
\scalebox{0.8}{
\small
\begin{tabular}{lc}
\toprule
Method & Average mAP \\
\midrule
average aggregation & 61.0 \\
text baseline~\cite{radford2021clip} & 62.6 \\
\midrule
Ours Full& 68.0 \\
- w/o $\lambda$~tuning &  63.2\\
- w/o dual generator & 66.1 \\
- w/o dynamic negative mining & 67.1 \\
- w/o repeating input & 67.8 \\
\bottomrule
\end{tabular}
}}
\vspace{5pt}
\caption{Ablation study for design choices explaining the final architectural design. mAP: mean Average Precision.}
\label{tab:ablation}
\end{wraptable}

\noindent\textbf{$\lambda$ tuning } Our approach, even without tuning $\lambda$ (fixed at 0.5), outperforms CLIP~\cite{radford2021clip}, unlike simple average aggregation, though the improvements are modest. However, with a properly tuned $\lambda$, we observe significant performance gains, highlighting the importance of balancing contributions from different modalities.

\begin{table}[t]
\centering
\footnotesize
\newcolumntype{M}[1]{>{\centering\arraybackslash}m{#1}}
\setlength{\tabcolsep}{8pt}
   \begin{adjustbox}{width=0.95\textwidth}
   \begin{tabular}
   {M{3cm}M{2.1cm}M{1.5cm}M{1.5cm}M{1.5cm}M{1.5cm}}
   \toprule
Class Name & Ground Truth & \multicolumn{2}{c}{Generated} & \multicolumn{2}{c}{Top Negative}\\\midrule

Hammerhead Shark 
&\includegraphics[height=0.08\textwidth, width=0.08\textwidth]{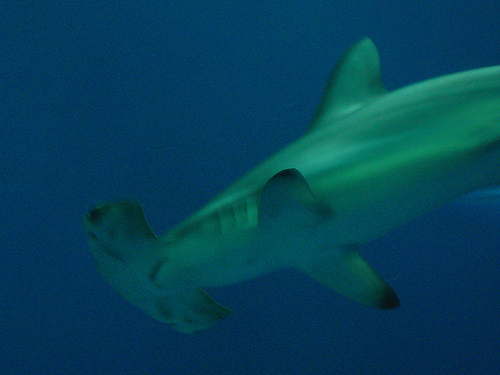}
&\includegraphics[height=0.08\textwidth, width=0.08\textwidth]{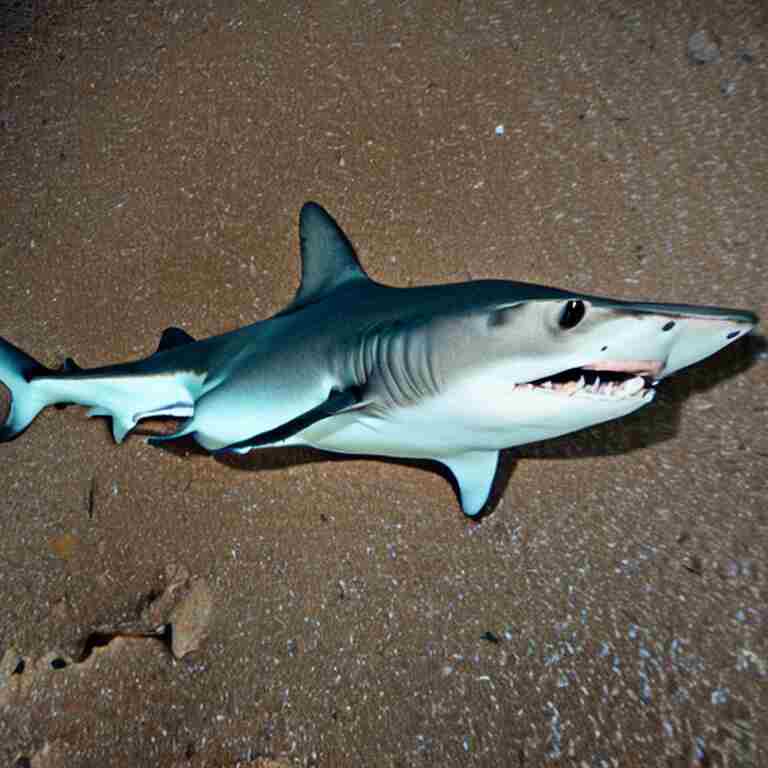}
&\includegraphics[height=0.08\textwidth, width=0.08\textwidth]{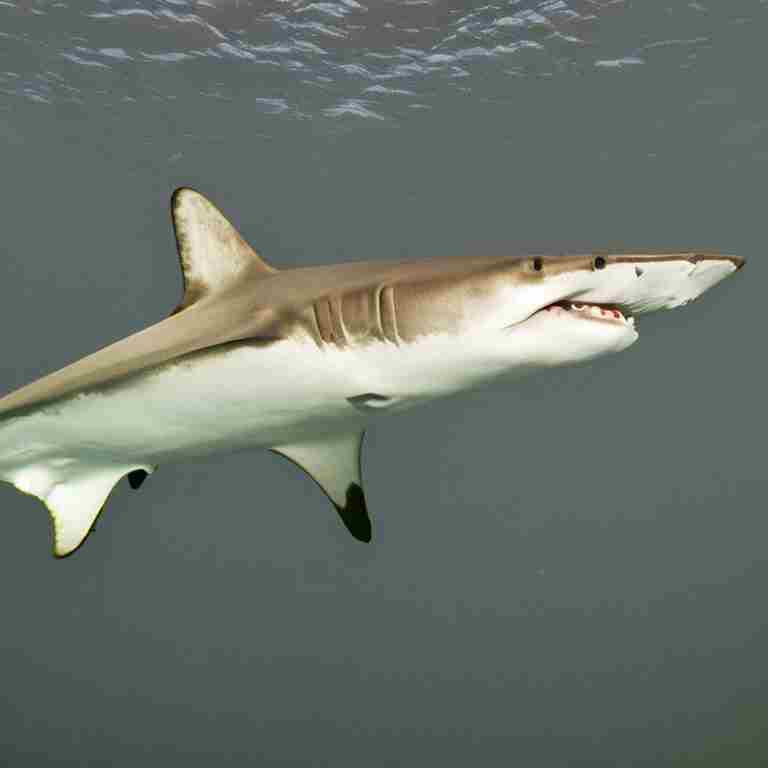}
&\includegraphics[height=0.08\textwidth, width=0.08\textwidth]{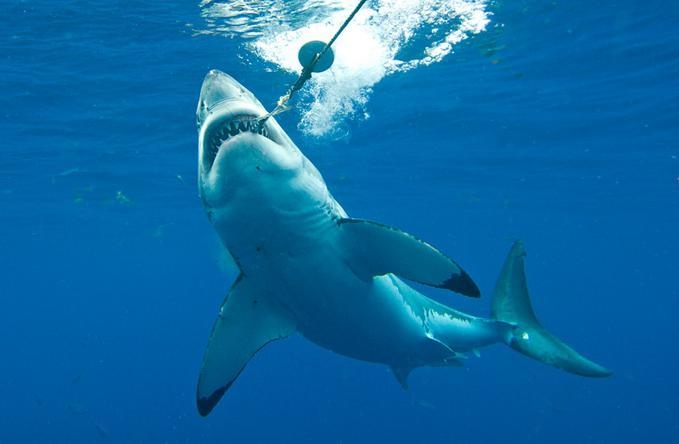}
&\includegraphics[height=0.08\textwidth, width=0.08\textwidth]{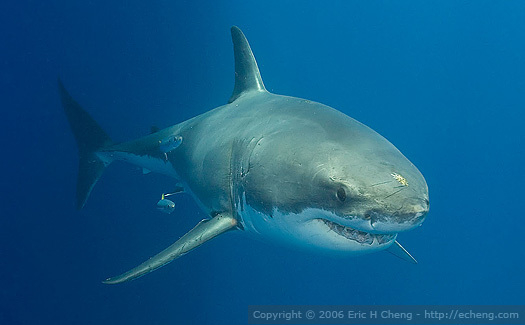}\\

Frilled-necked Lizard 
&\includegraphics[height=0.08\textwidth, width=0.08\textwidth]{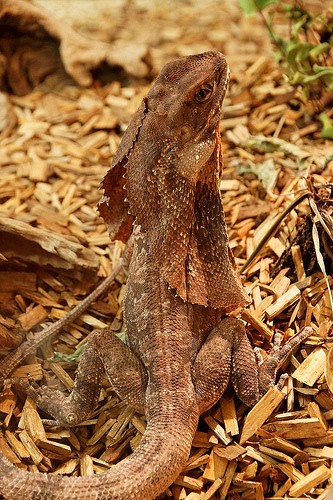}
&\includegraphics[height=0.08\textwidth, width=0.08\textwidth]{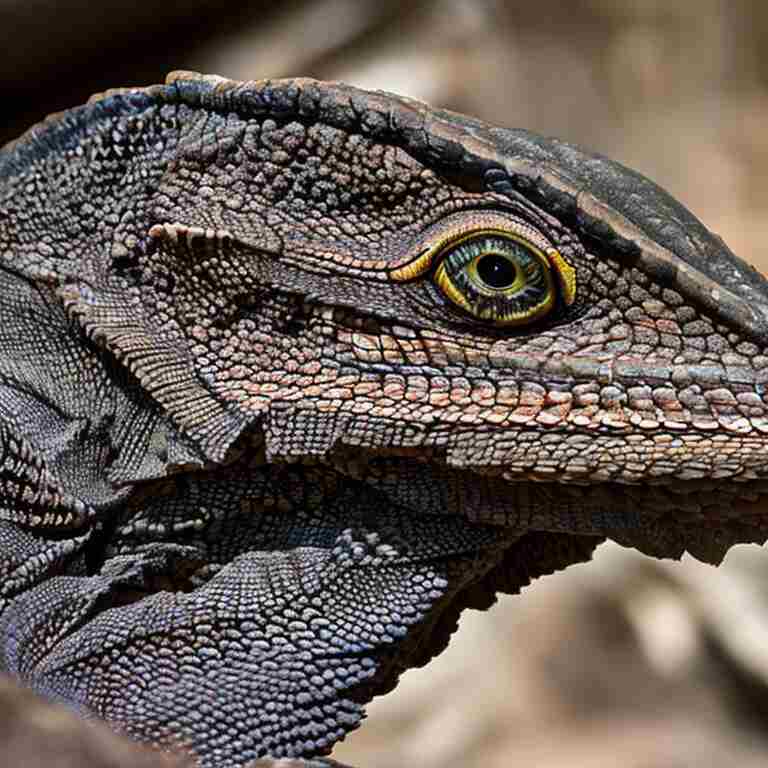}
&\includegraphics[height=0.08\textwidth, width=0.08\textwidth]{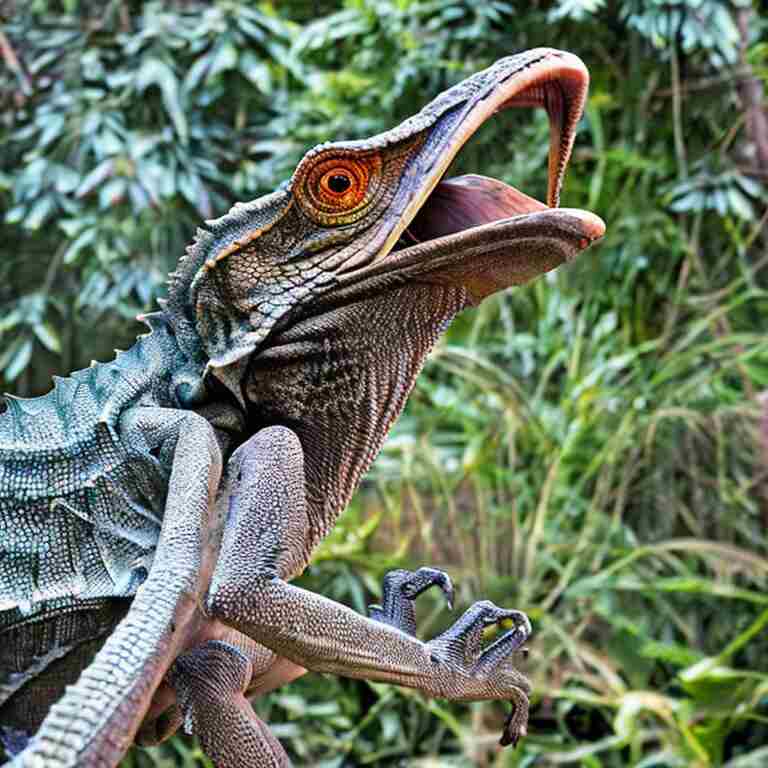}
&\includegraphics[height=0.08\textwidth, width=0.08\textwidth]{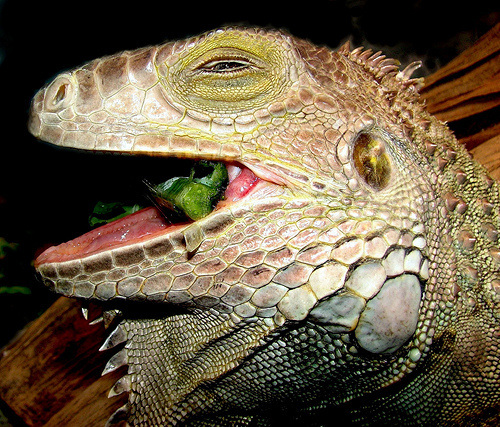}
&\includegraphics[height=0.08\textwidth, width=0.08\textwidth]{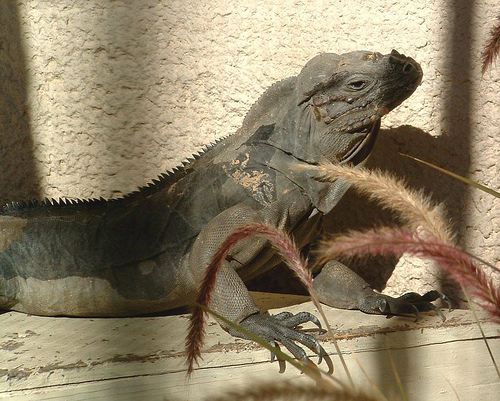}\\

Leatherback Sea Turtle 
&\includegraphics[height=0.08\textwidth, width=0.08\textwidth]{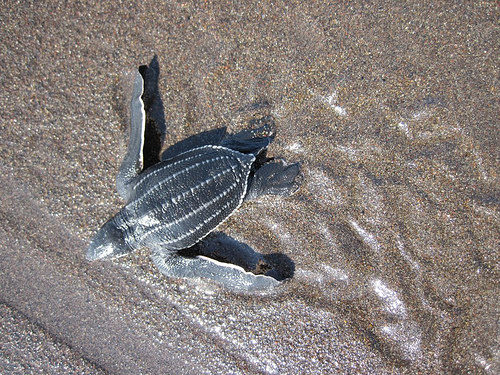}
&\includegraphics[height=0.08\textwidth, width=0.08\textwidth]{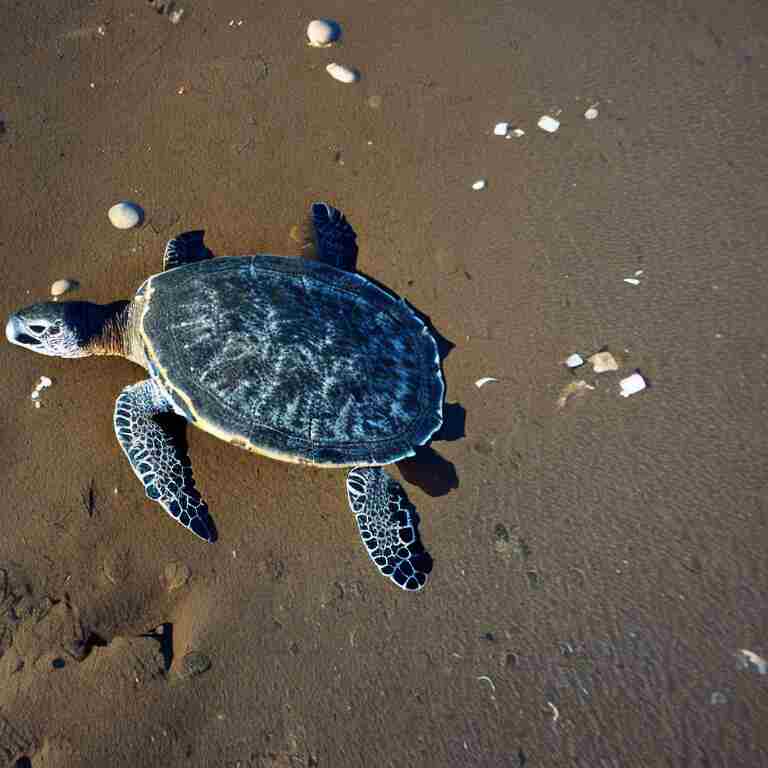}
&\includegraphics[height=0.08\textwidth, width=0.08\textwidth]{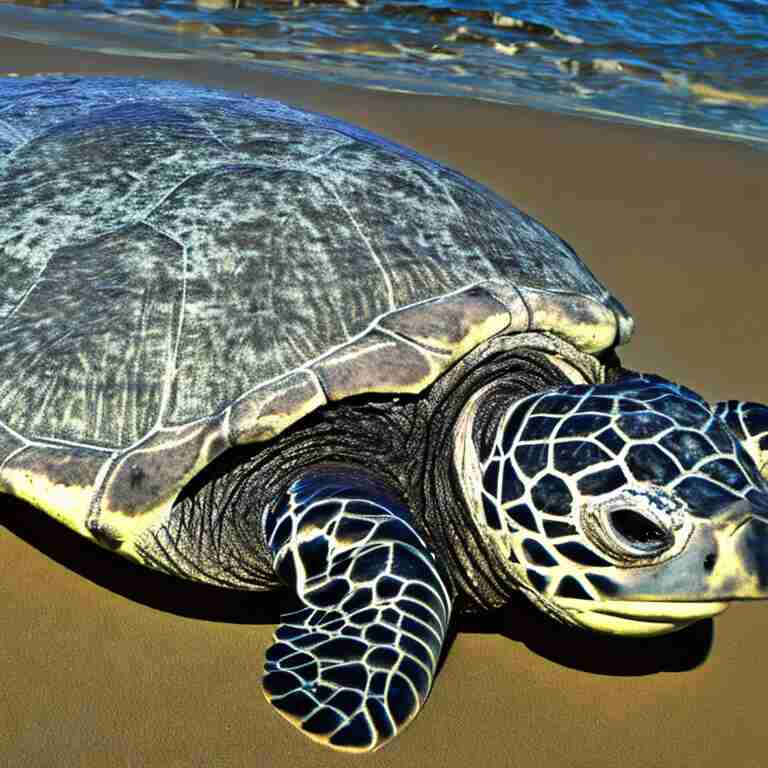}
&\includegraphics[height=0.08\textwidth, width=0.08\textwidth]{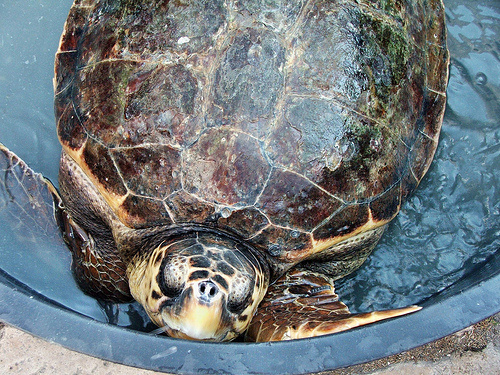}
&\includegraphics[height=0.08\textwidth, width=0.08\textwidth]{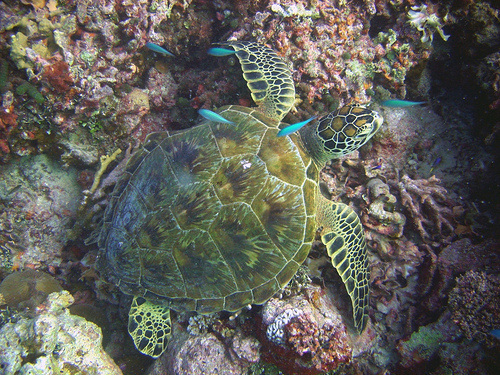}\\

\midrule

Pig 
&\includegraphics[height=0.08\textwidth, width=0.08\textwidth]{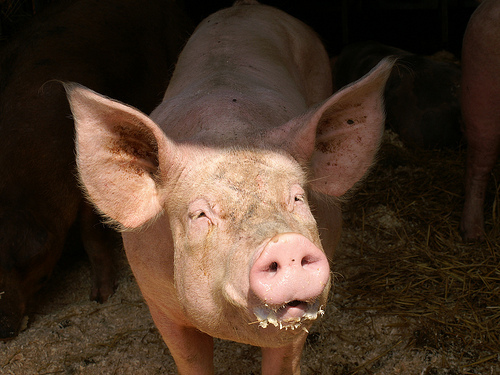}
&\includegraphics[height=0.08\textwidth, width=0.08\textwidth]{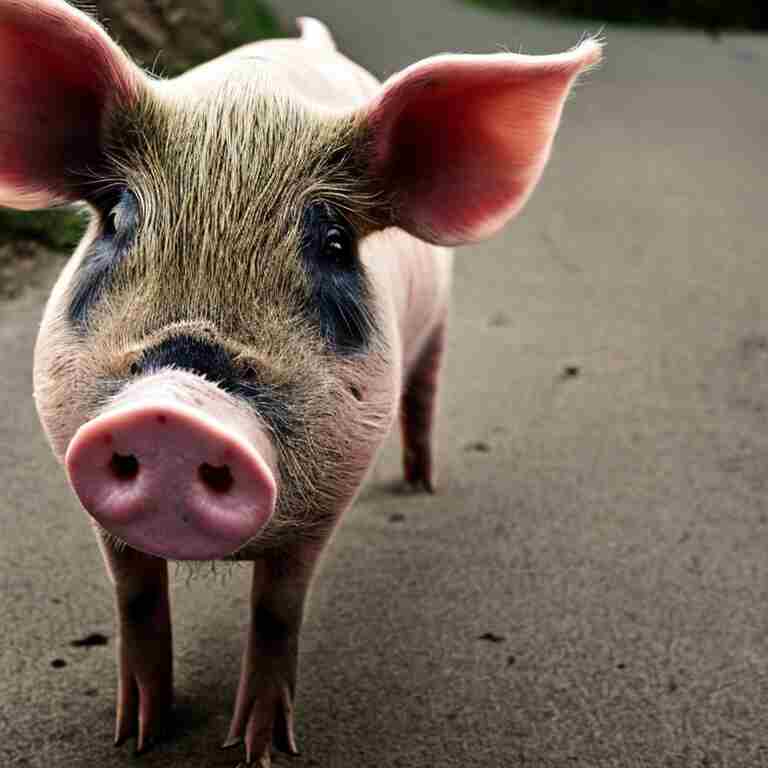}
&\includegraphics[height=0.08\textwidth, width=0.08\textwidth]{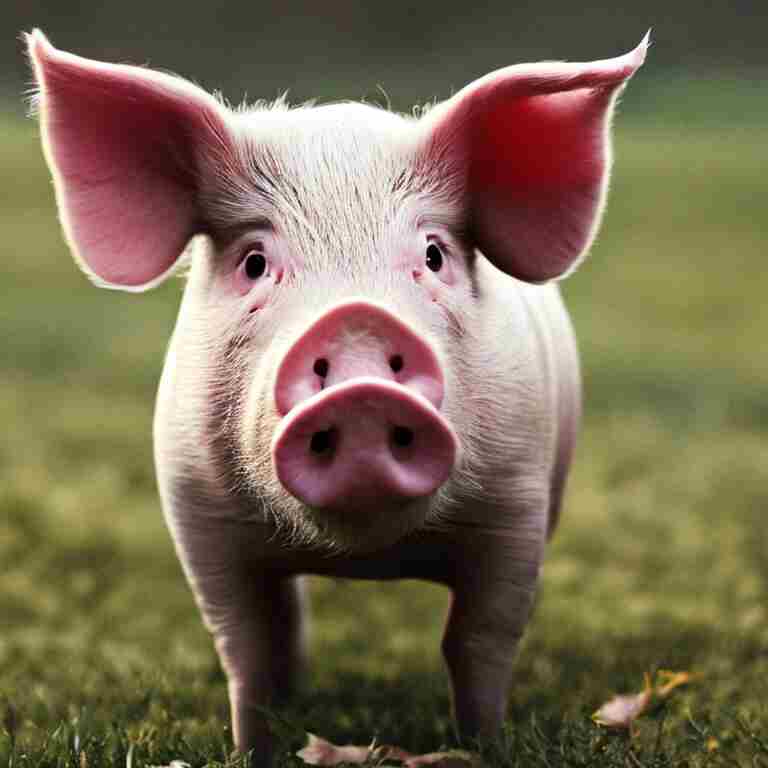}
&\includegraphics[height=0.08\textwidth, width=0.08\textwidth]{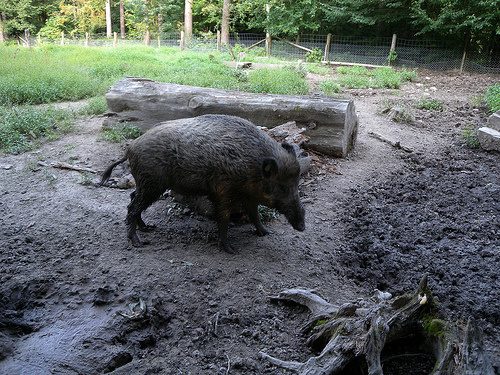}
&\includegraphics[height=0.08\textwidth, width=0.08\textwidth]{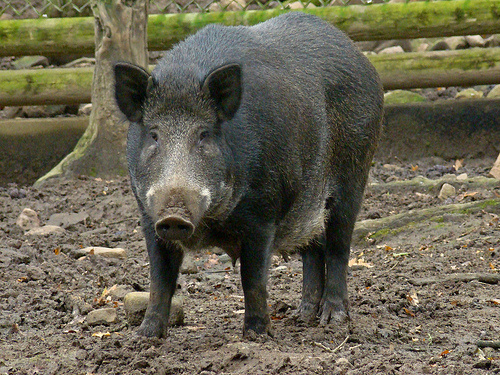}\\

Maltese 
&\includegraphics[height=0.08\textwidth, width=0.08\textwidth]{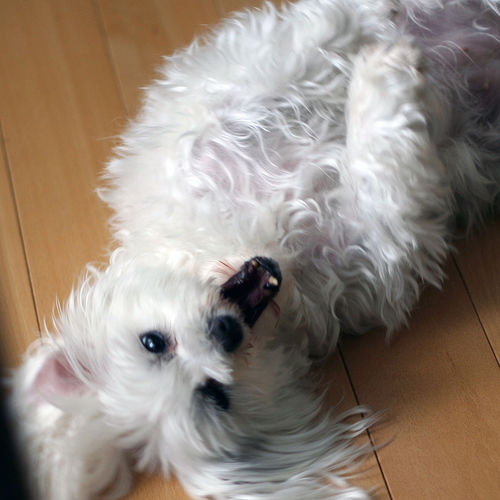}
&\includegraphics[height=0.08\textwidth, width=0.08\textwidth]{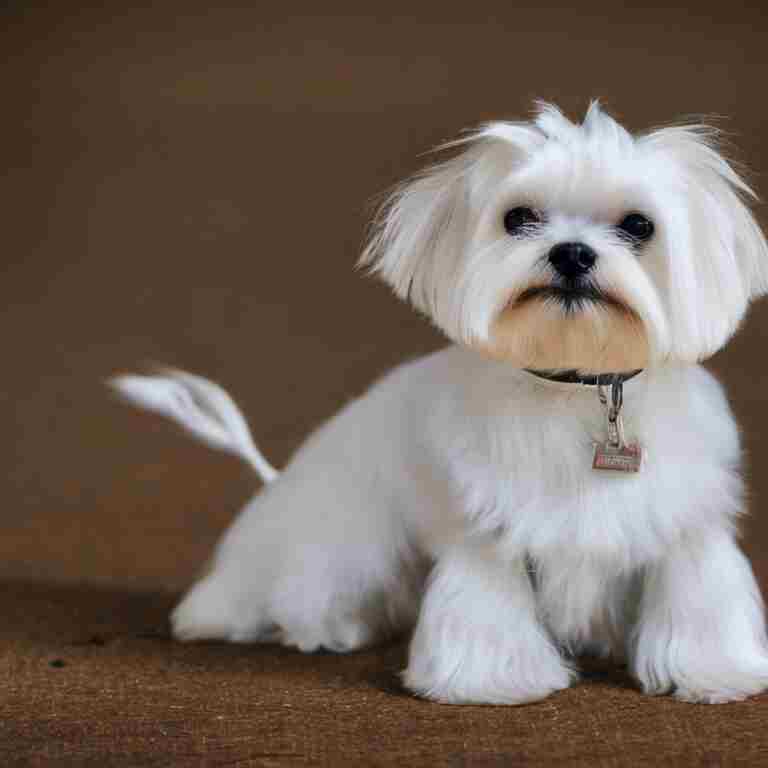}
&\includegraphics[height=0.08\textwidth, width=0.08\textwidth]{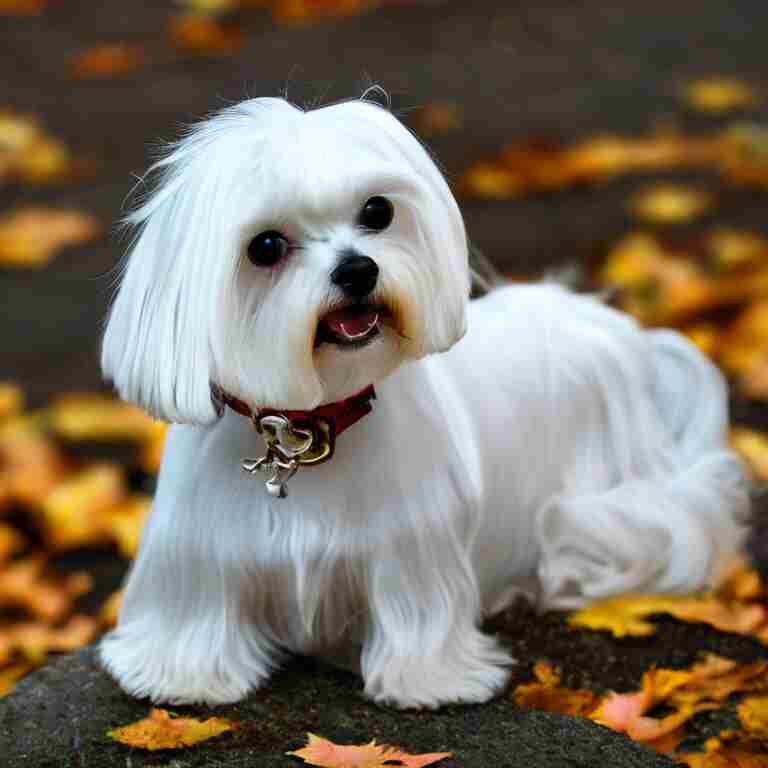}
&\includegraphics[height=0.08\textwidth, width=0.08\textwidth]{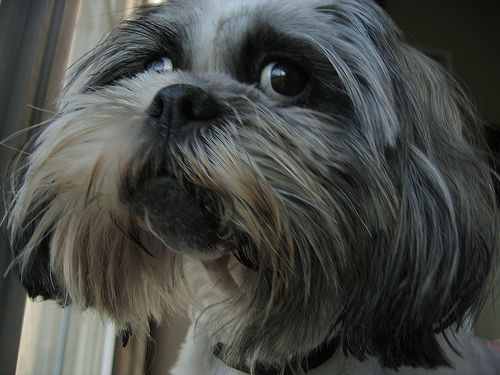}
&\includegraphics[height=0.08\textwidth, width=0.08\textwidth]{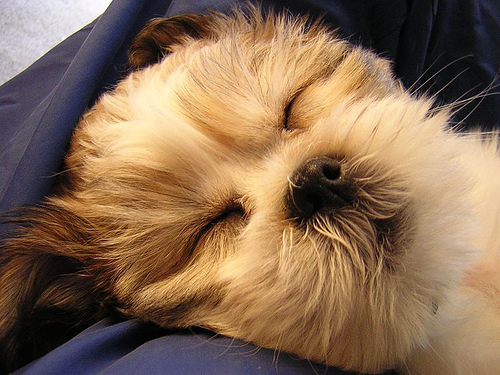}\\

Prairie Grouse 
&\includegraphics[height=0.08\textwidth, width=0.08\textwidth]{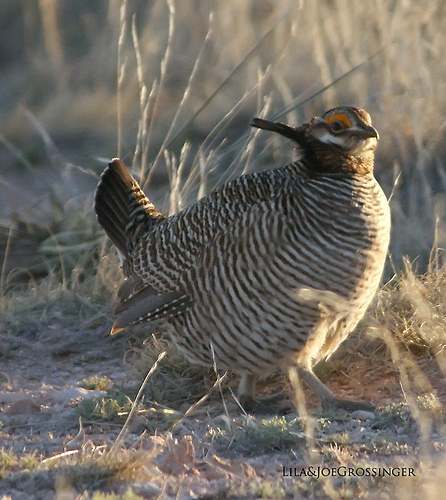}
&\includegraphics[height=0.08\textwidth, width=0.08\textwidth]{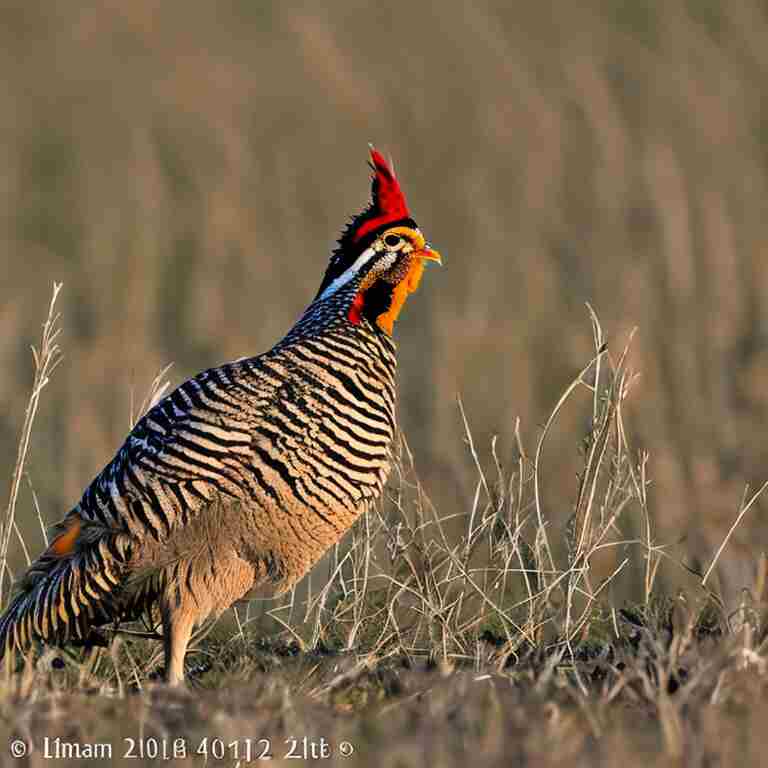}
&\includegraphics[height=0.08\textwidth, width=0.08\textwidth]{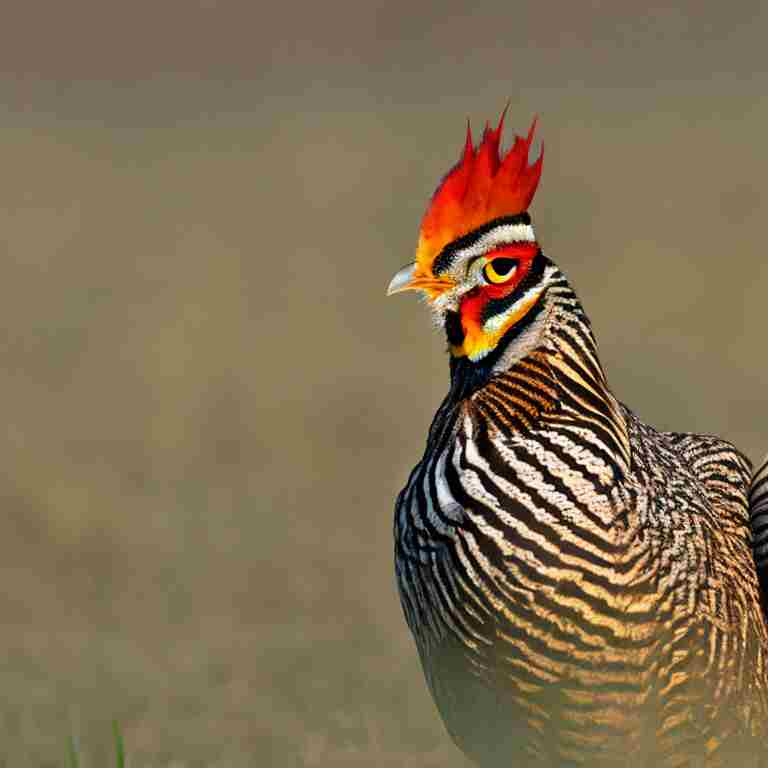}
&\includegraphics[height=0.08\textwidth, width=0.08\textwidth]{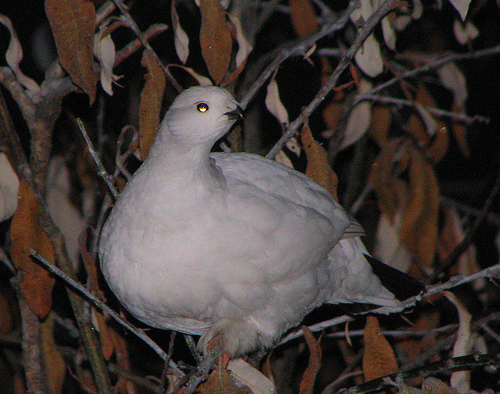}
&\includegraphics[height=0.08\textwidth, width=0.08\textwidth]{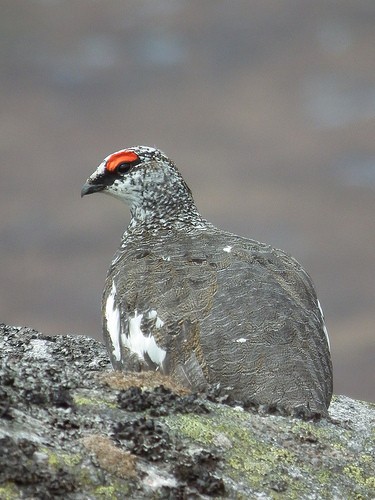} \\

\bottomrule

\end{tabular}
\end{adjustbox}

\vspace{5pt}
\caption{Overview of the classes where our approach has the largest (gain at bottom rows, loss at top rows) difference in performance over the simple class-name text-only approach. 
Using our approach, we show the class name, a real image from the class, the generated images, and the top-ranked negative images.
} 
\label{tab:worst-best}
\end{table} 

\noindent\textbf{Dual Generator} Our model is trained using a combination of SD~\cite{sd} and FLUX~\cite{flux} to enhance diversity during training. To examine its impact, we compare it to a model trained exclusively with synthetic images from SD~\cite{sd}. Using two generators is clearly better. 

\noindent\textbf{Dynamic Negative Mining} We dynamically mine the hardest negative within the batch in the standard approach and compare with mining once at the beginning of the training using the hybrid baseline method. Dynamic sampling provides a more adaptive selection of challenging negatives, leading to improved model robustness. 

\noindent\textbf{Repeating Attention Inputs}  In this experiment, the input to the second attention layer is the output of the first layer for all tokens instead of re-feeding the $k$ original input tokens.  This results in a drop of 0.2\% across fifteen benchmarks.

\subsection{Analysis}

\textbf{Where do synthetic images help/harm?}
For better understanding, we visualize the distribution of the relevant features using t-SNE in Figure~\ref{fig:analyses}. 
The most common case of improvements is due to the different generators capturing different aspects of a class, which often works even if one of them is making mistakes. However, when image queries appear mostly near negatives, our approach can hurt the baseline.

\noindent\textbf{Examples}
Table~\ref{tab:worst-best} highlights ImageNet classes with the largest performance gaps between our method and the text-only baseline. Performance drops often stem from GDM missing key visual cues—e.g., "hammer-shaped head" for hammerhead sharks—leading to confusion with similar-looking negatives. In contrast, our approach improves retrieval for visually similar classes like "pig" vs. "guinea pig" or "Prairie Grouse" vs. "Ruffed Grouse" by leveraging visual cues that help disambiguate where text alone falls short.

\subsection{Results for Non-Class-Related Queries}

\begin{wraptable}[12]{b}{0.5\textwidth} 
  \centering
  \vspace{-5pt}
\vspace{-5mm}
\centering
\setlength\tabcolsep{12pt}
\scalebox{0.5}{
\begin{tabular}{lcccc}
\toprule
  & & \multicolumn{3}{c}{\textbf{Flickr-30k~\cite{flickr30}}} \\
Variant & Method & R@1 & R@5 & R@10   \\
\cmidrule(r){1-1} 
\cmidrule(r){2-2} 
\cmidrule(lr){3-5} 
\multirow{3}{*}{Clip-based} &
CLIP & 64.9 & 87.3 & 92.0 \\
&DIVA~\cite{wang2025diffusion} & 64.4 & 86.9 & 92.0 \\
& Ours(C,D) & \textbf{69.9} & \textbf{90.0} & \textbf{94.4} \\
\hline
\multirow{4}{*}{MetaCLIP-based} &
MetaCLIP & 73.4 & 92.3 & 95.8 \\
&MODE-2~\cite{mode} & 73.4 & 92.5 & 95.8 \\
&MODE-4~\cite{mode} & 73.5 & 92.1 & 95.9 \\
&Ours(M,D) & \textbf{74.8} & \textbf{93.1} & \textbf{96.3}  \\
\hline
\multirow{3}{*}{LLM-based} &
TIgER(LaVIT)~\cite{tiger} & 68.8 & 82.9 & 86.4 \\
&TIgER(SEED-LLaMA)~\cite{tiger} & 71.7 & 91.8 & 95.4 \\
&Frozen~\cite{pang2024frozen} & 50.2 &  82.3 & 90.1 \\
\bottomrule
\end{tabular}
}

  \vspace{5pt}
\caption{Results of text-to-image retrieval(IR) on Flickr30K. Image captions form text query.C: CLIP, D: DINOv2, M: MetaCLIP. Both MODE-2~\cite{mode} and MODE-4~\cite{mode} are initialized from MetaCLIP~\cite{metaclip}.}

\label{tab:flickr}
\end{wraptable}

While our work focuses on category-level image retrieval, we also show that our approach can enhance CLIP-like models for general retrieval where image captions form the text query. We compare our method against several works that improve or extend CLIP: DIVA~\cite{wang2025diffusion} leverages generative feedback from text-to-image diffusion models to refine CLIP representations using only images; Mixture of Data Experts (MoDE)~\cite{mode} optimizes a system of CLIP data experts via clustering, with each expert trained on a specific data cluster to reduce sensitivity to false negatives. Both TIgER~\cite{tiger} and FrozenLLM~\cite{pang2024frozen} explore the discriminative abilities of Multi-modal Large Language Models(MLLMs). TIgER~\cite{tiger} introduces a generative retrieval method that operates in a training-free manner, and FrozenLLM~\cite{pang2024frozen} demonstrates that frozen transformer blocks from pre-trained language models can serve as effective visual encoders. We report the results in~\Cref{tab:flickr}. It can be seen that our model shows better improvements on both CLIP and MetaCLIP compared to previous works. We also perform better than baselines that utilize MLLMs for retrieval.

\section{Conclusions}
In this work, we revisit category-level text-to-image retrieval, expanding on the capabilities of VLMs. While VLMs serve as a robust starting point, we significantly advance beyond this by leveraging a diverse suite of foundational generative and representation models. By incorporating synthetic image generation via text prompts and specialized encoders for image-to-image similarity, we achieve substantial performance gains across a wide range of datasets. Our improvements enable better browsing of large image archives and research training sets. 
\section{Acknowledgement}
This work was supported by the Junior Star GACR GM 21-28830M, the Czech Technical University in Prague grant No. SGS23/173/OHK3/3T/13, and by KAUST, under Award No. BAS/1/1685-01-01.
\bibliography{egbib}
\appendix
\clearpage
\setcounter{page}{1}

\section*{Supplementary Content for Category-level Text-to-Image Retrieval Improved: Bridging the Domain Gap with Diffusion Models and Vision Encoders:}

This supplementary document includes the following 
\begin{itemize}
    \item \textbf{Section \ref{sec:arch}}:  We show the overview of the aggregator architecture.
    \item \textbf{Section \ref{sec:rvss}}:  We compare the performance of our synthetic visual queries with ``perfect'' visual queries.
    \item \textbf{Section \ref{sec:dbr}}:  We report further results for Description Based Retrieval and show examples where it outperforms class name-based retrieval.
    \item \textbf{Section \ref{sec:clip}:} We report results on three more CLIP-based backbones.
    \item  \textbf{Section \ref{sec:rob}:} We show the robustness of our approach by analyzing the performance on ImageNet-C~\cite{hendrycks2019robustness}.
    \item  \textbf{Section \ref{sec:qual_top10ret}:} We present the top retrieved images across various categories, comparing two settings: one where the class name is provided and another where only the class description is used.
\end{itemize}

\section{Architecture}
\begin{figure}[h]
  \includegraphics[width=\textwidth]{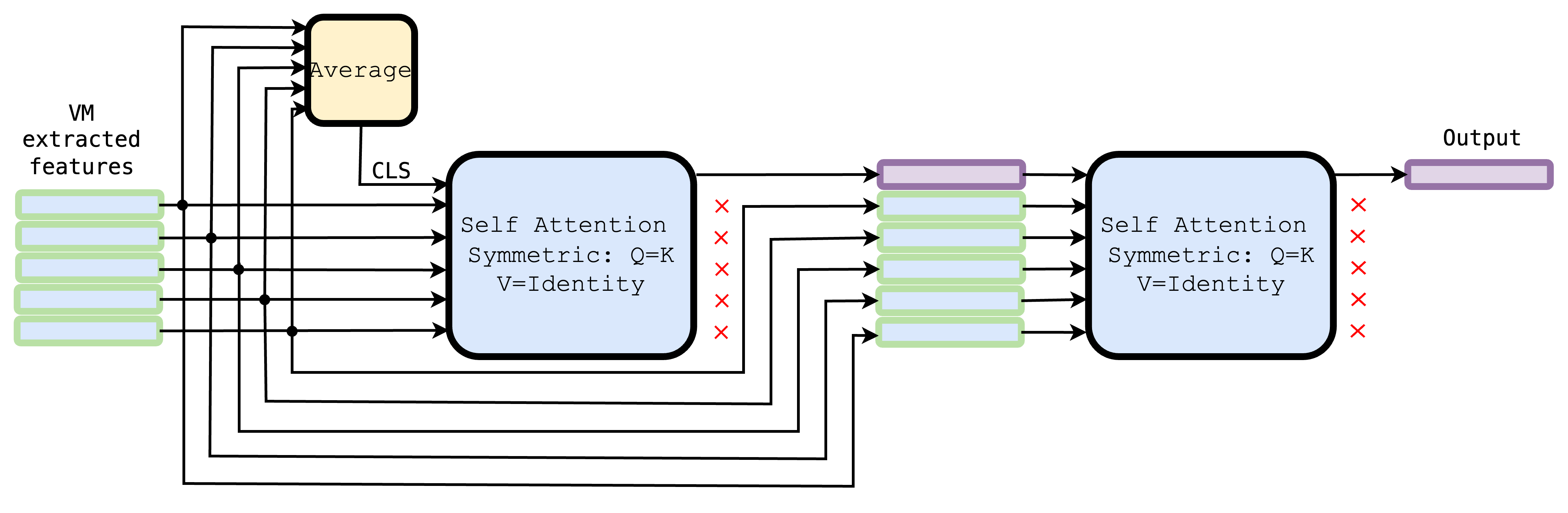}
  \caption{Overview of the aggregator network leveraging symmetric self-attention where the CLS token equals the average representation of input features. The same input features are processed through multiple self-attention layers to generate the final aggregated output.}
  \label{fig:arch}
  \vspace{-5mm}
\end{figure}

Figure~\ref{fig:arch} illustrates our proposed architecture for aggregating vision model (VM) extracted features using a symmetric self-attention mechanism. Given the visual features, we first prepend a CLS token(average of the inputs) and pass the sequence through a self-attention block where the query and key matrices are shared (Q=K) and the value is set to identity (V=Identity). This symmetric setup simplifies the attention computation while maintaining the ability to contextualize the CLS token. 
\label{sec:arch}

\section{Real vs. Synthetic Queries}
\label{sec:rvss}

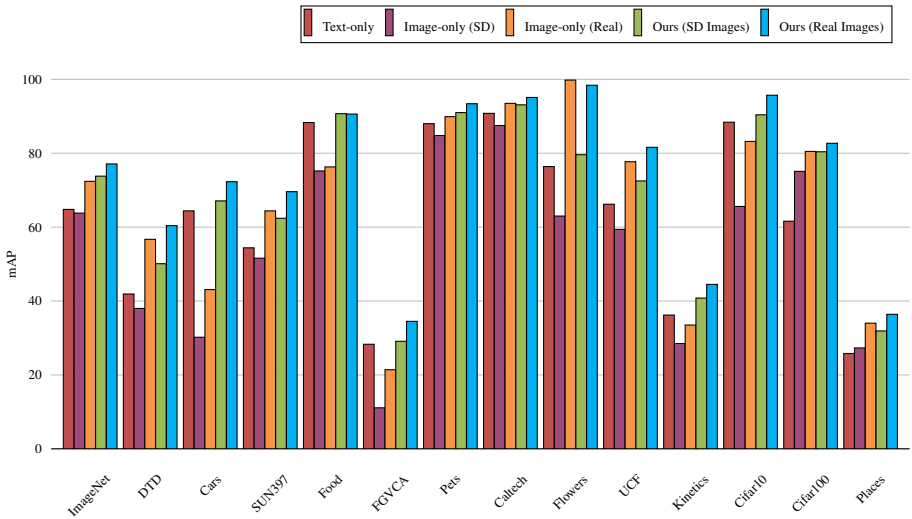
\begin{figure*}
    \centering
    \pgfplotstableread{
dataset        combo_sd  combo_real   image_sd    image_real   text
ImageNet	73.8	77.1	63.8	    72.4	    64.8
DTD	        50.1	60.4	38.0	    56.7	    41.9
Cars	    67.1	72.3	30.2	    43.1	    64.4
SUN397	    62.4	69.6	51.6	    64.4	    54.4
Food	    90.7	90.6	75.2	    76.3	    88.3
FGVCA	    29.1	34.5	11.1	    21.4	    28.3
Pets	    91.0	93.4	84.8	    89.9	    88.0
Caltech	    93.1    95.1	87.5	    93.5	    90.8
Flowers	    79.6	98.4	63.0	    99.8	    76.4
UCF	        72.5	81.6	59.4	    77.7	    66.2
Kinetics	40.8    44.5	28.5	    33.5	    36.2
Cifar10	    90.4	95.7	65.6	    83.2	    88.4
Cifar100	80.4	82.7	75.1	    80.5	    61.6
Places	    31.9	36.4	27.3	    34.0	    25.8
}{\realsddata}
\definecolor{rred}{HTML}{C0504D}
\definecolor{ggreen}{HTML}{9BBB59}
\definecolor{ppurple}{HTML}{9F4C7C}
\definecolor{bblue}{HTML}{4F81BD}   
\definecolor{oorange}{HTML}{F79646} 
\definecolor{tteal}{HTML}{00B0F0}   

\pgfplotsset{every x tick label/.append style={font=\tiny, rotate=45}}
\pgfplotsset{every y tick label/.append style={font=\tiny}}

\begin{tikzpicture}
    \begin{axis}[
        width  = \linewidth,
        height = 0.5\linewidth,,
        major x tick style = transparent,
        ybar=0.1\pgflinewidth,
        bar width=4pt,
        ylabel = {mAP},
        symbolic x coords={ImageNet,DTD,Cars,SUN397,Food,FGVCA,Pets,Caltech,Flowers,UCF,Kinetics,Cifar10,Cifar100,Places},
        xtick = data,
        scaled y ticks = false,
        enlarge x limits=0.05,
        ymin=0,
        ymax=100,
        ylabel style={font=\tiny, yshift=-20pt},
        scatter/position=absolute,
        grid=none,
        ymajorgrids=true,
        axis x line*=bottom,
        axis y line*=none,
        every outer y axis line/.append style={draw=none},
        every y tick/.append style={draw=none},
        legend image code/.code={
        \draw [#1] (0cm,-0.1cm) rectangle (0.1cm,0.25cm); },
        legend style={
                at={(0.98,1.2)},
                column sep=2pt,
                legend columns=5,
                font=\tiny,
                inner sep=1pt,
        }
    ]
        \addplot[style={black,fill=rred,mark=none}] 
            table[x=dataset, y expr={\thisrow{text}}]  \realsddata; \addlegendentry{Text-only}
        \addplot[style={black,fill=ppurple,mark=none}] 
            table[x=dataset, y expr={\thisrow{image_sd}}]  \realsddata; \addlegendentry{Image-only (SD)}
        \addplot[style={black,fill=oorange,mark=none}] 
            table[x=dataset, y expr={\thisrow{image_real}}]  \realsddata; \addlegendentry{Image-only (Real)}
        \addplot[style={black,fill=ggreen,mark=none}] 
            table[x=dataset, y expr={\thisrow{combo_sd}}]  \realsddata; \addlegendentry{Ours (SD Images)}
        \addplot[style={black,fill=tteal,mark=none}] 
            table[x=dataset, y expr={\thisrow{combo_real}}]  \realsddata; \addlegendentry{Ours (Real Images)}
    \end{axis}
\end{tikzpicture}
    \vspace{5pt}
    \caption{Comparison between synthetically generated and real images used as queries. Performance is measured via Mean average precision (mAP).
    \label{fig:real_sd}
    \vspace{-15pt}
    }
\end{figure*}

This experiment explores the potential performance improvement achievable by utilizing ``perfect'' visual queries instead of our synthetically generated ones from a GDM using the text queries $y$. These perfect visual queries are sampled($5$ for each query) from the training set of each benchmark dataset\footnote{We do not report the results on RESISC45 as we use the entire dataset as a database.}. We employ these visual queries alongside text queries using our approach. Figure~\ref{fig:real_sd} illustrates that employing perfect visual queries enhances retrieval performance across all fourteen benchmarks by approximately 11.3\% on average compared to the text-only baseline.  Additionally, observing the same comparison with the image-only variant reveals a significant performance gap of 11.6\% between SD-generated images and perfect image queries. This indicates the gap, seen through the lens of category-level retrieval, between real and generated images.
This analysis shows that there is still scope left for improving GDM. As new approaches appear, we can easily plug our approach with new GDM to get closer and closer to the ``perfect'' visual queries.

\section{Description Based Retrieval}
\label{sec:dbr}
As reported in Subsection 5.3 of the main paper, generated descriptions do not contain the explicit mention of the class name. We ensure this by reviewing the generated descriptions. In~\Cref{tab:dbr_wins} and~\Cref{tab:description_based}, we visualize the images generated using the class descriptions, and it can be seen that the synthetic images can provide useful information.

\subsection{Where Does Description Shine?}
\begin{figure*}[htbp]
    \centering
    \begin{subfigure}
        \centering
        \includegraphics[width=0.3\linewidth]{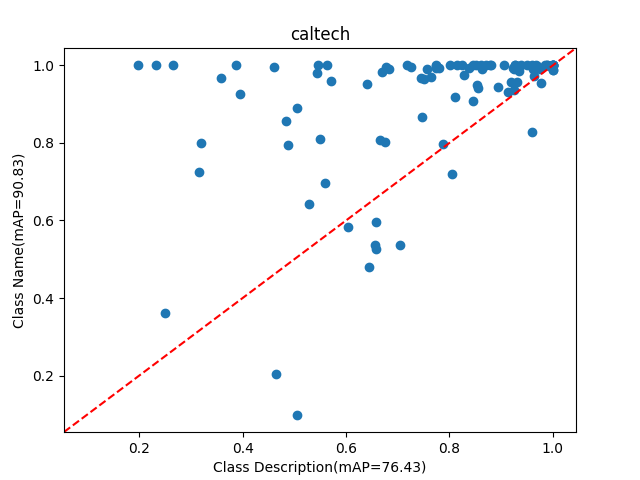}
        \label{fig:plot1}
    \end{subfigure}%
    \begin{subfigure}
        \centering
        \includegraphics[width=0.3\linewidth]{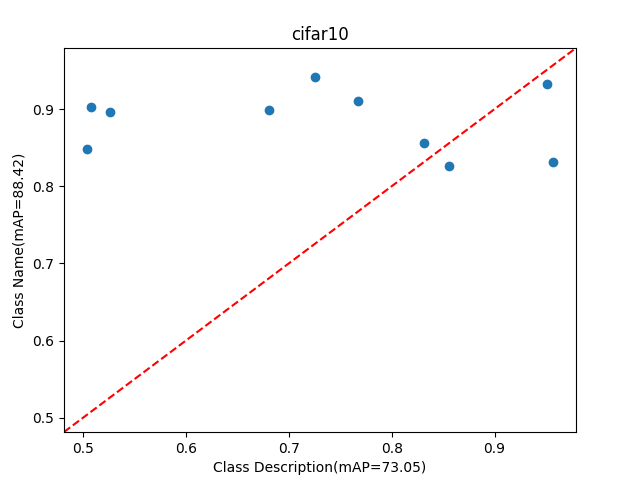}
        \label{fig:plot2}
    \end{subfigure}%
    \begin{subfigure}
        \centering
        \includegraphics[width=0.3\linewidth]{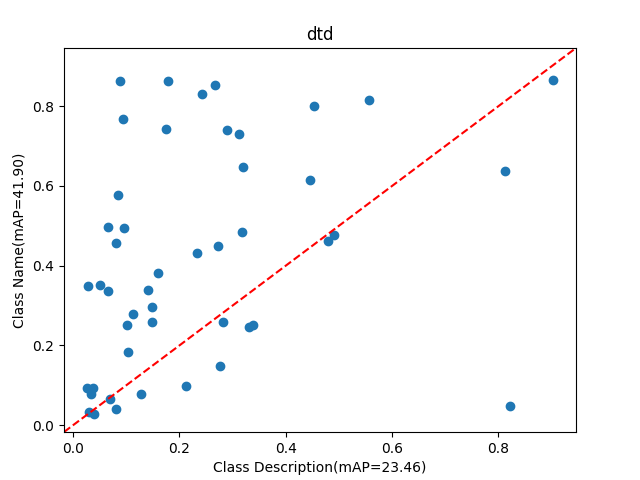}
        \label{fig:plot3}
    \end{subfigure}\\
    \begin{subfigure}
        \centering
        \includegraphics[width=0.3\linewidth]{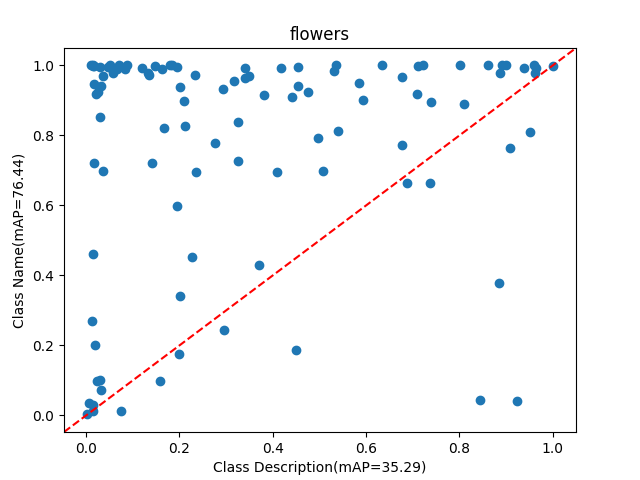}
        \label{fig:plot4}
    \end{subfigure}%
    \begin{subfigure}
        \centering
        \includegraphics[width=0.3\linewidth]{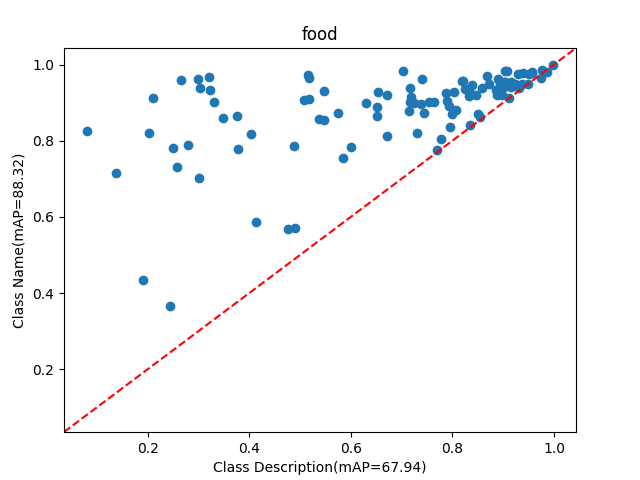}
        \label{fig:plot5}
    \end{subfigure}
    \begin{subfigure}
        \centering
        \includegraphics[width=0.3\linewidth]{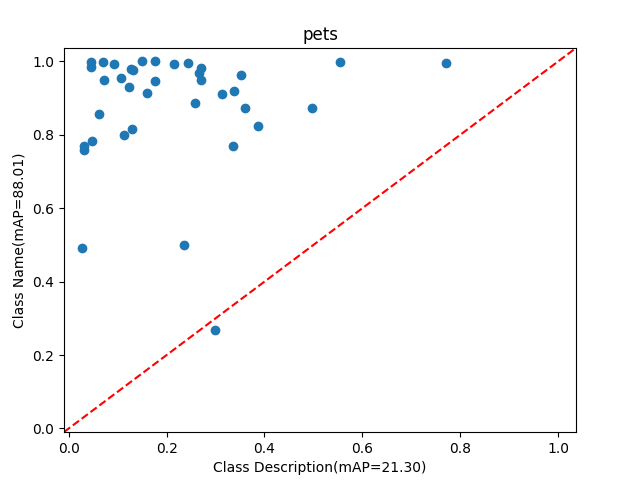}
        \label{fig:plot6}
    \end{subfigure}
    \caption{Performance Comparison of Class Name and Class Description Retrieval. Each data point represents the mAP of a class, contrasting its retrieval performance using class names (y-axis) against class descriptions (x-axis). The diagonal red line indicates equivalent performance, with points above/below revealing performance disparities between the two retrieval approaches.}
    \label{fig:all_dbr}
\end{figure*}

We analyze the performance of Class description retrieval by comparing it with Class name retrieval across six benchmarks. In certain cases, the Class description is more helpful than the Class name. We first compare the performance of the two approaches in~\Cref{fig:all_dbr} and then in~\Cref{tab:dbr_wins} we report the samples generated for classes where the Class description is more useful than the Class name. We also report more qualitative samples in~\Cref{tab:description_based}.

\begin{table*}
\newcolumntype{M}[1]{>{\centering\arraybackslash}m{#1}}
\setlength{\tabcolsep}{8pt}
\renewcommand{\arraystretch}{1.5}
   \begin{adjustbox}{width=0.95\textwidth}
   \begin{tabular}
   {M{3cm}M{4cm}M{2.1cm}M{1.5cm}M{1.5cm}M{1.5cm}M{1.5cm}}
   \toprule
Class Name & Description based text & Ground Truth & \multicolumn{2}{c}{Description Based} & \multicolumn{1}{c}{Class Name}\\\midrule

Ball Moss & the flowers of air plants are typically small and come in a variety of colors, such as white, pink, or purple.
&\includegraphics[height=0.12\textwidth, width=0.12\textwidth]{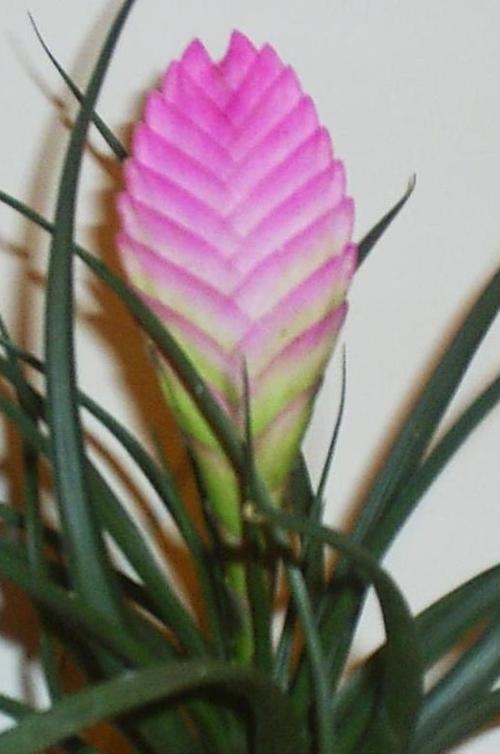}
&\includegraphics[height=0.12\textwidth, width=0.12\textwidth]{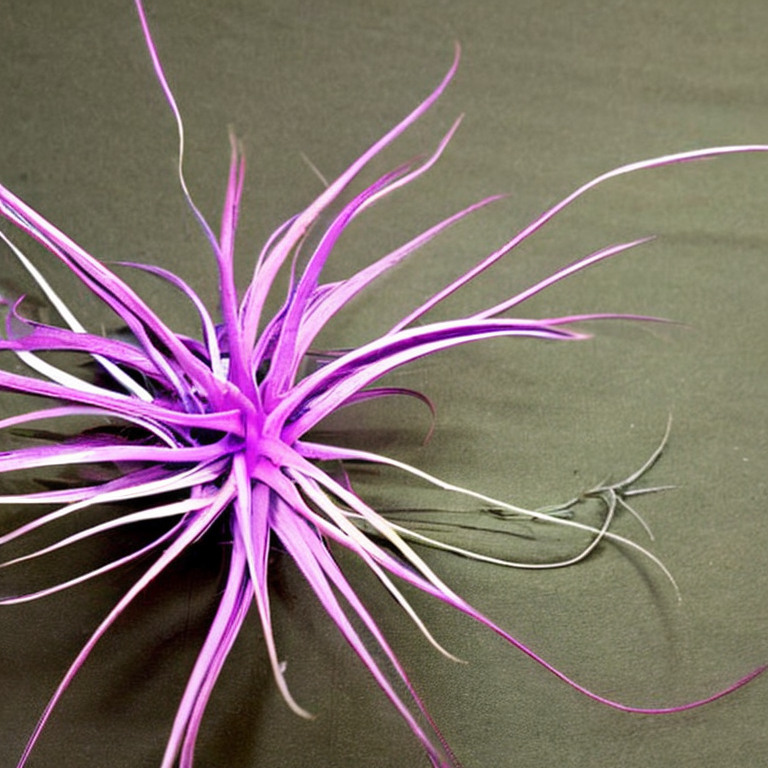}
&\includegraphics[height=0.12\textwidth, width=0.12\textwidth]{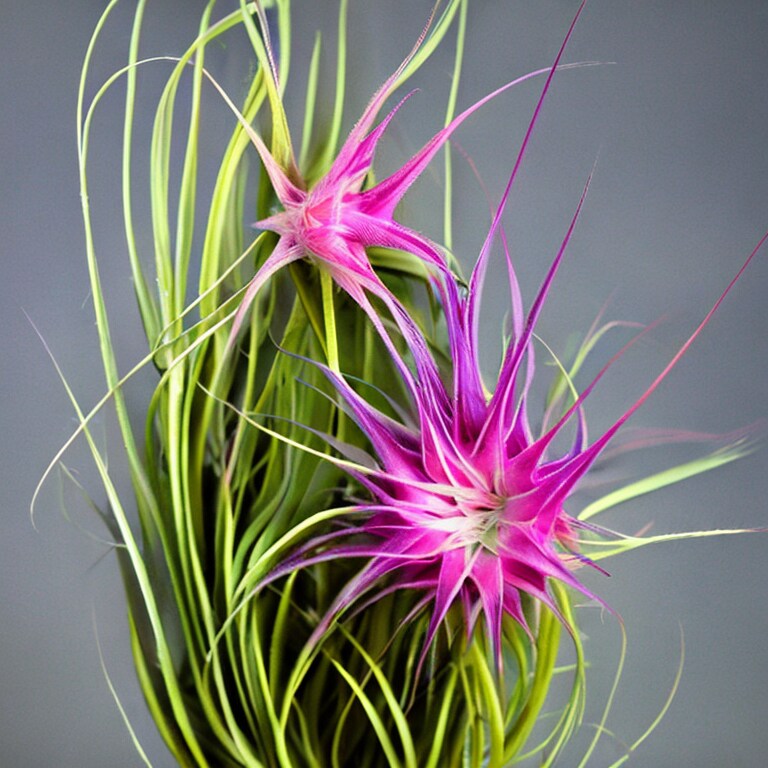}
&\includegraphics[height=0.12\textwidth, width=0.12\textwidth]{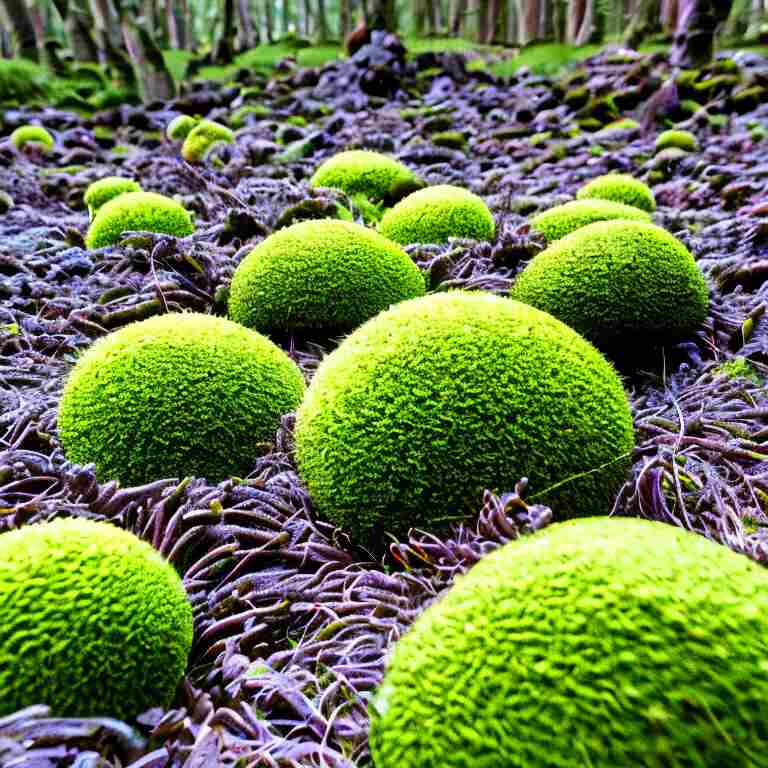} \\

Dotted & material with a series of small dots printed on it is commonly seen.
&\includegraphics[height=0.12\textwidth, width=0.12\textwidth]{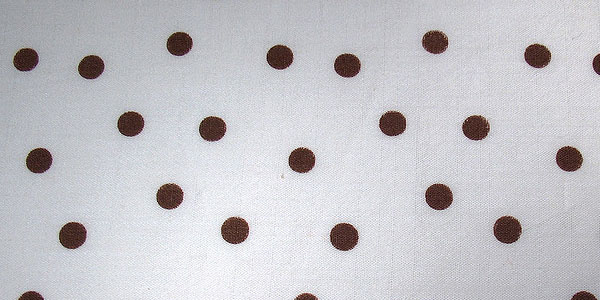}
&\includegraphics[height=0.12\textwidth, width=0.12\textwidth]{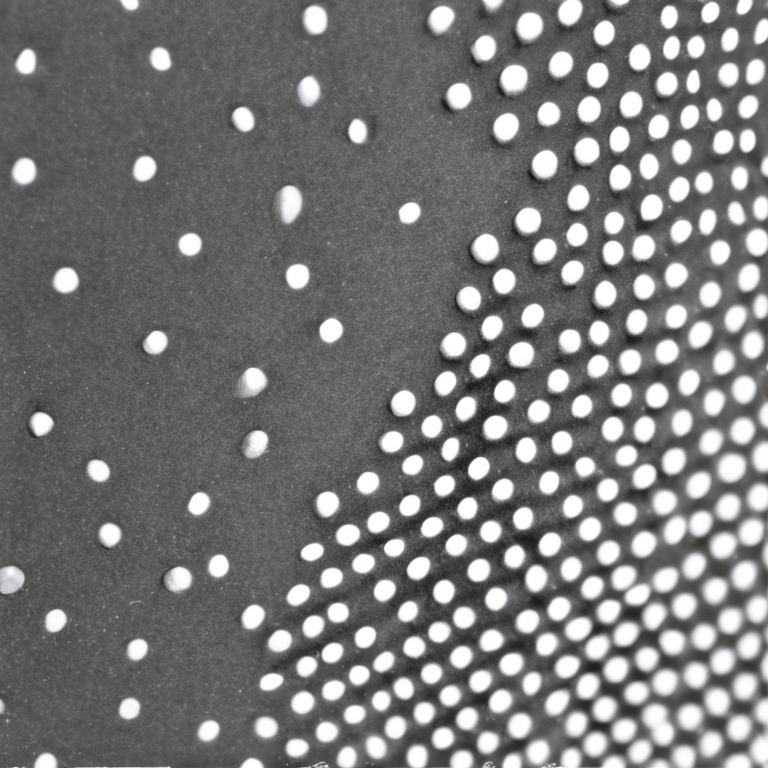}
&\includegraphics[height=0.12\textwidth, width=0.12\textwidth]{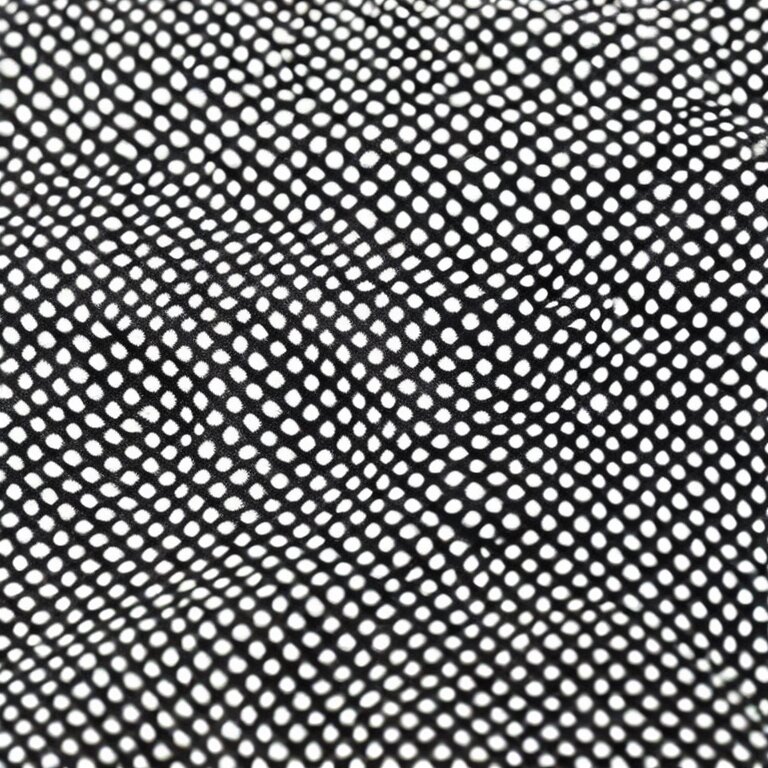}
&\includegraphics[height=0.12\textwidth, width=0.12\textwidth]{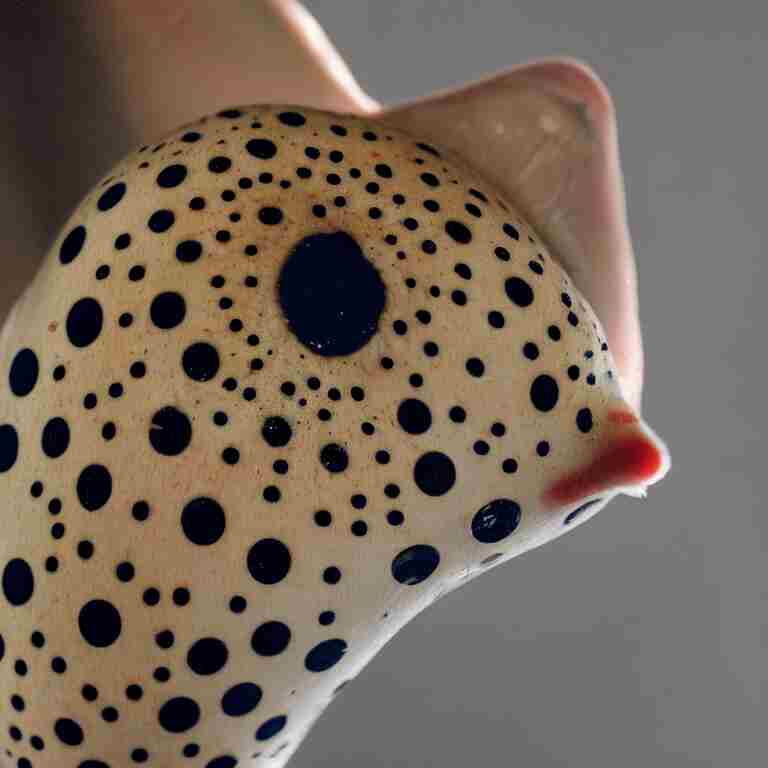} \\

Lacelike & the delicate fabric is known for its thinness and intricate patterns, often featuring small holes or gaps.
&\includegraphics[height=0.12\textwidth, width=0.12\textwidth]{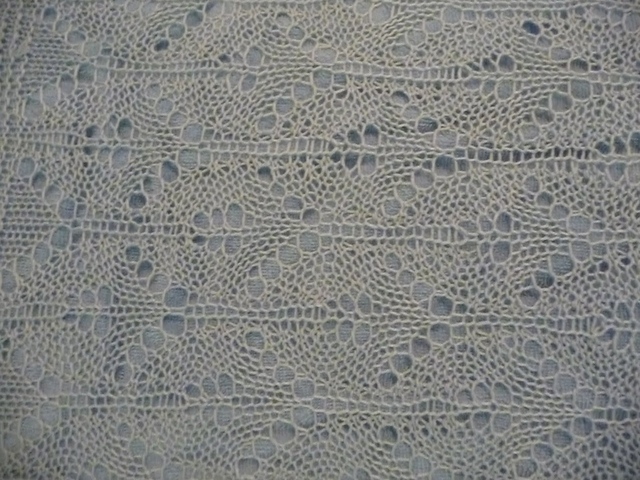}
&\includegraphics[height=0.12\textwidth, width=0.12\textwidth]{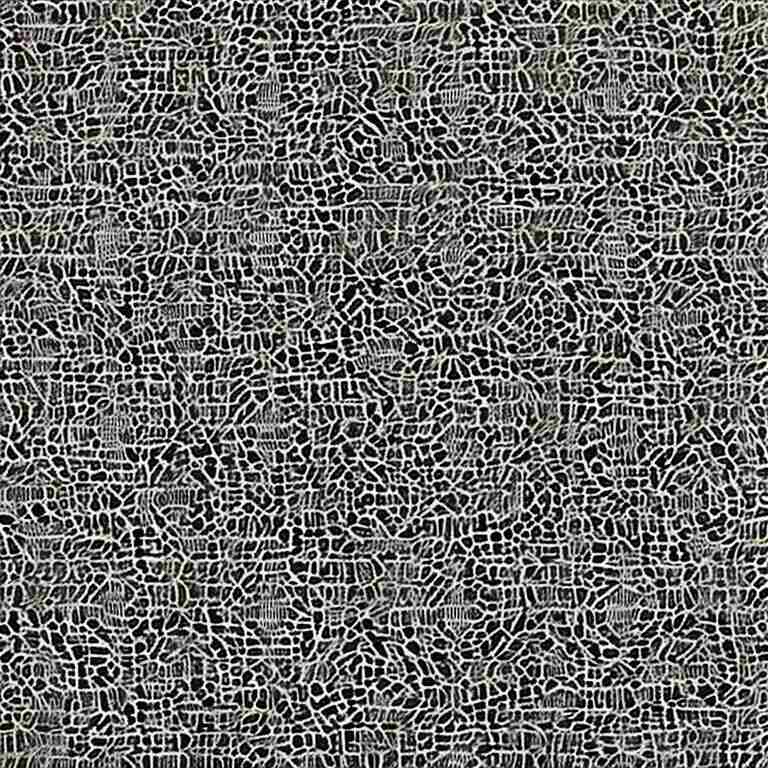}
&\includegraphics[height=0.12\textwidth, width=0.12\textwidth]{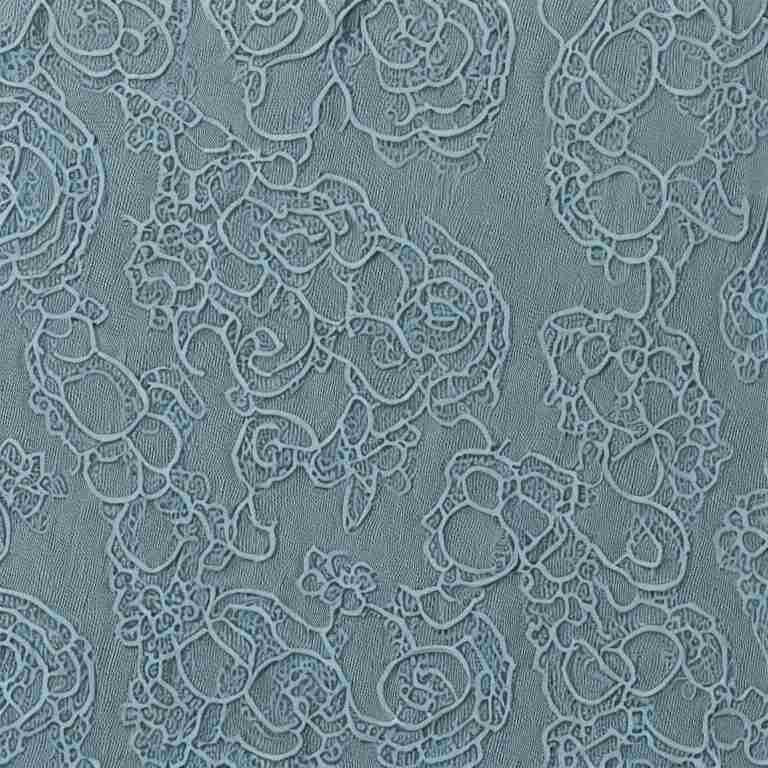}
&\includegraphics[height=0.12\textwidth, width=0.12\textwidth]{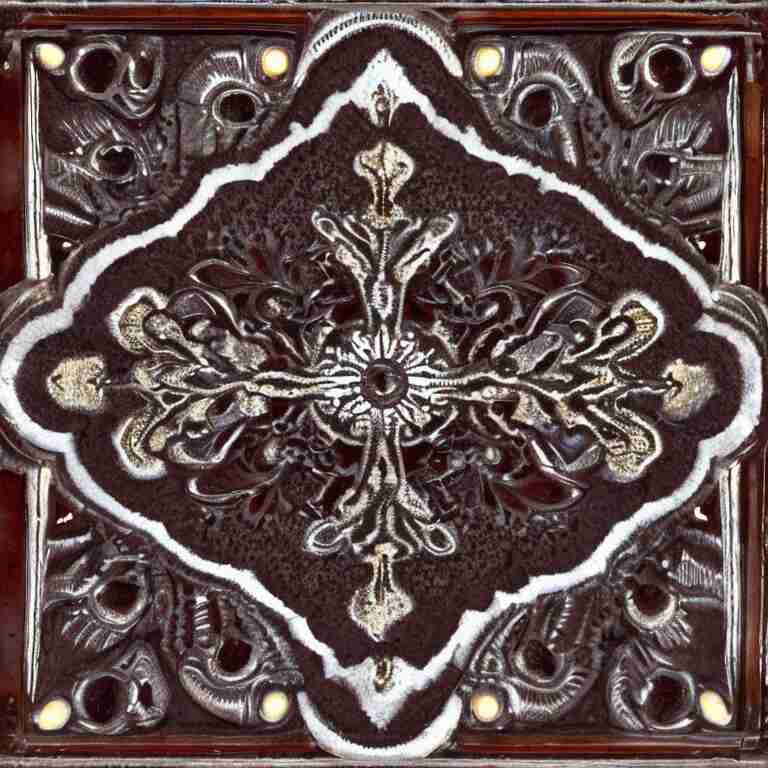} \\

Moon Orchid & the white flower with yellow and orange stripes has a unique appearance.
&\includegraphics[height=0.12\textwidth, width=0.12\textwidth]{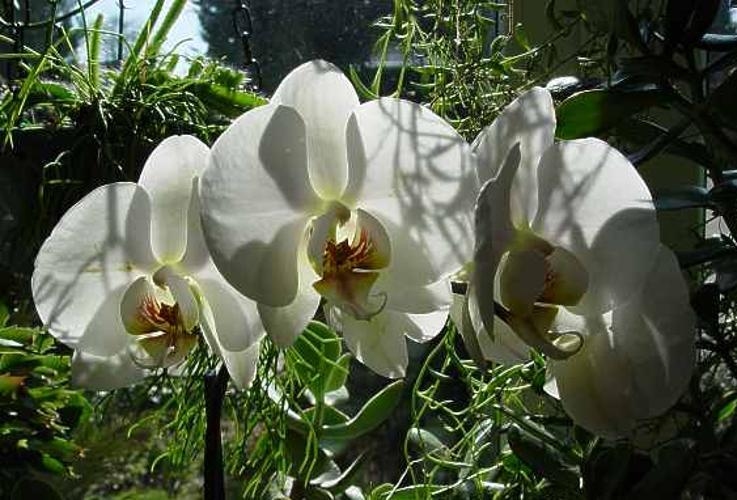}
&\includegraphics[height=0.12\textwidth, width=0.12\textwidth]{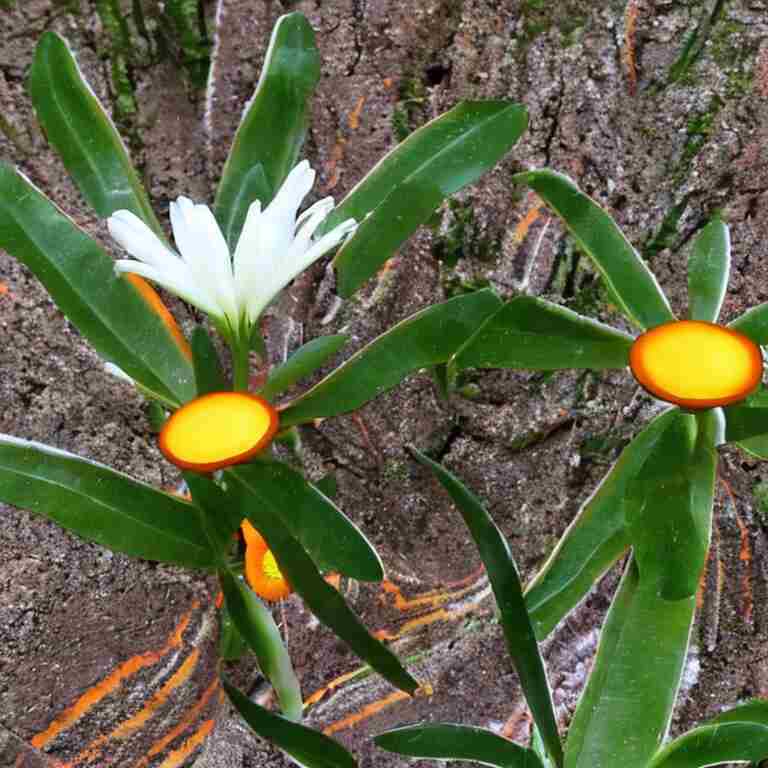}
&\includegraphics[height=0.12\textwidth, width=0.12\textwidth]{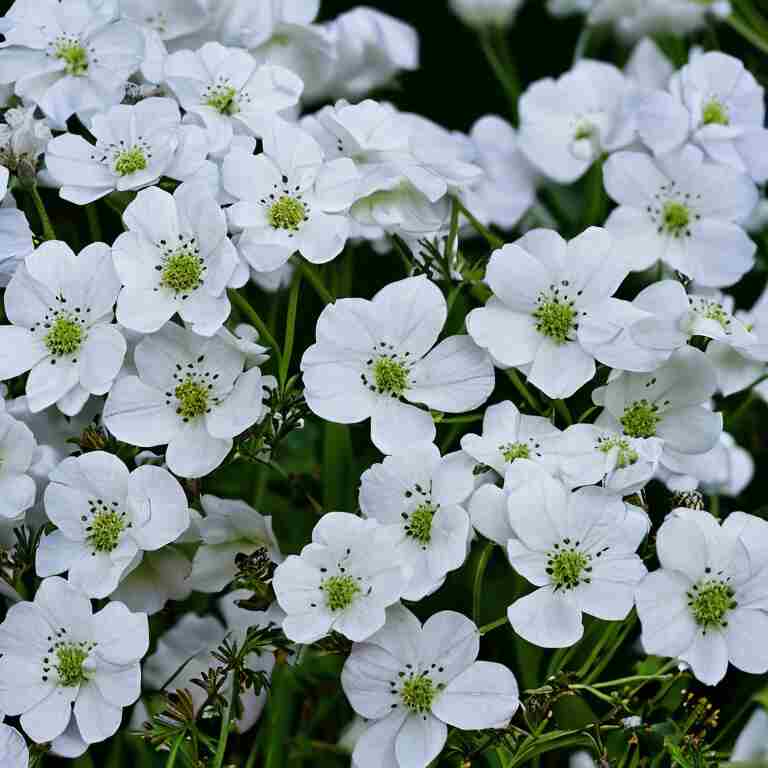}
&\includegraphics[height=0.12\textwidth, width=0.12\textwidth]{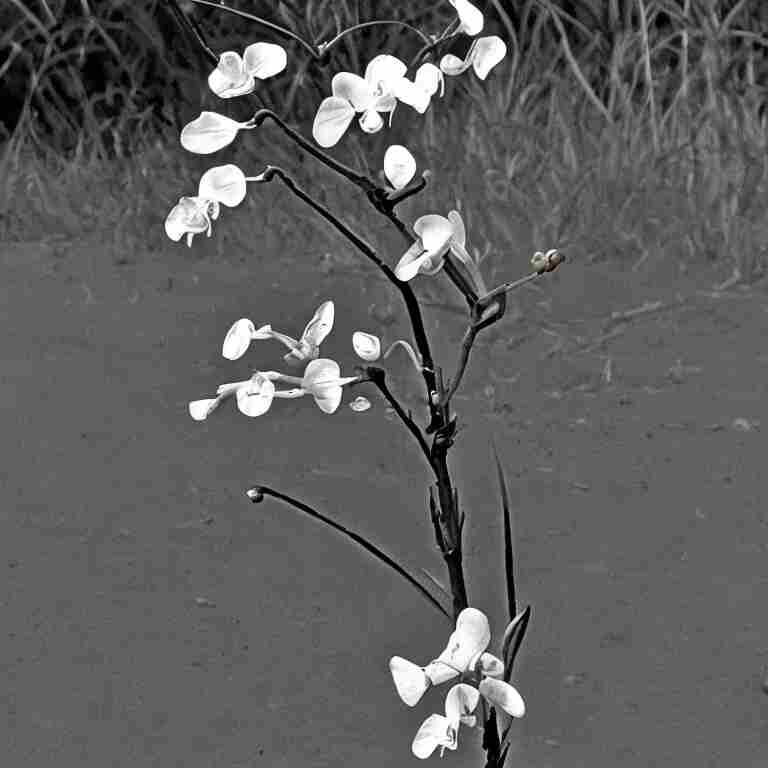} \\

Silverbush & the plant's foliage has a silvery hue and its blooms are petite and white.
&\includegraphics[height=0.12\textwidth, width=0.12\textwidth]{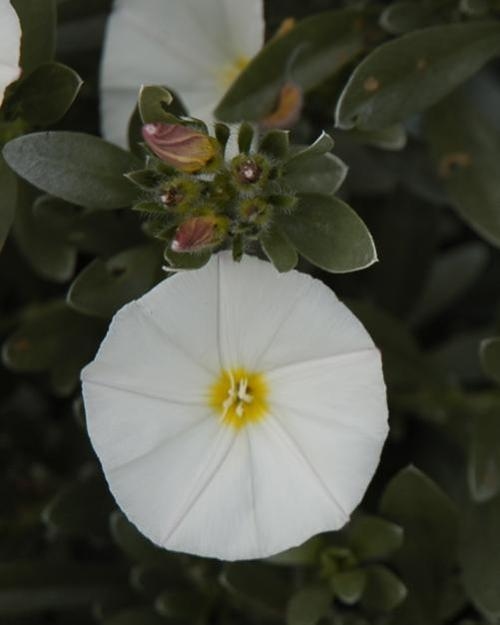}
&\includegraphics[height=0.12\textwidth, width=0.12\textwidth]{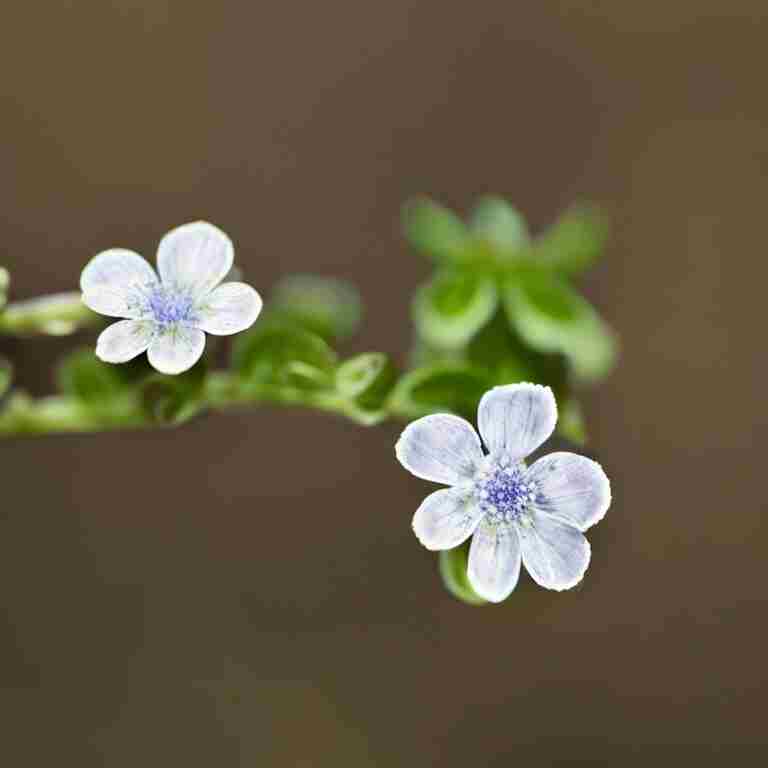}
&\includegraphics[height=0.12\textwidth, width=0.12\textwidth]{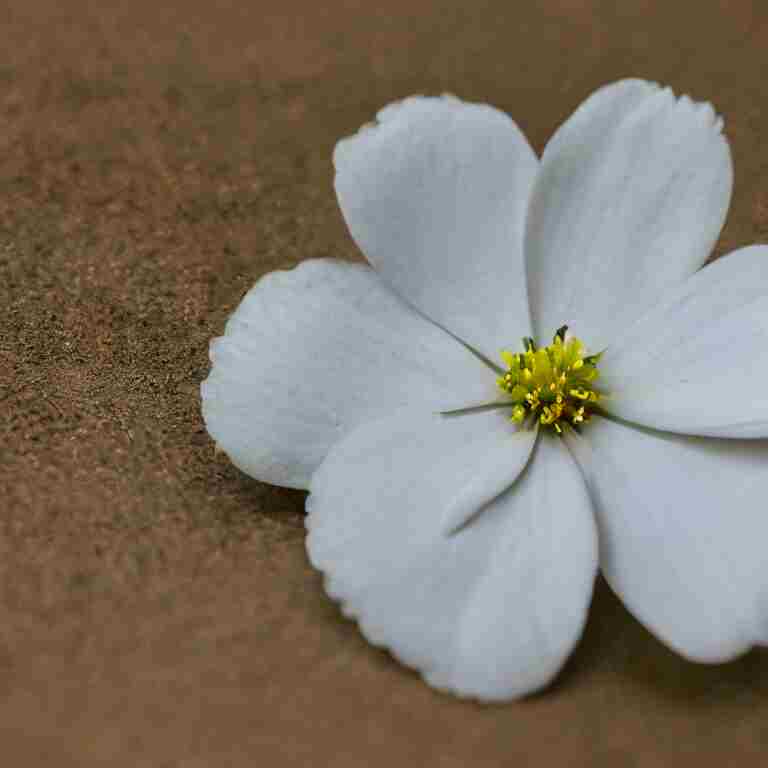}
&\includegraphics[height=0.12\textwidth, width=0.12\textwidth]{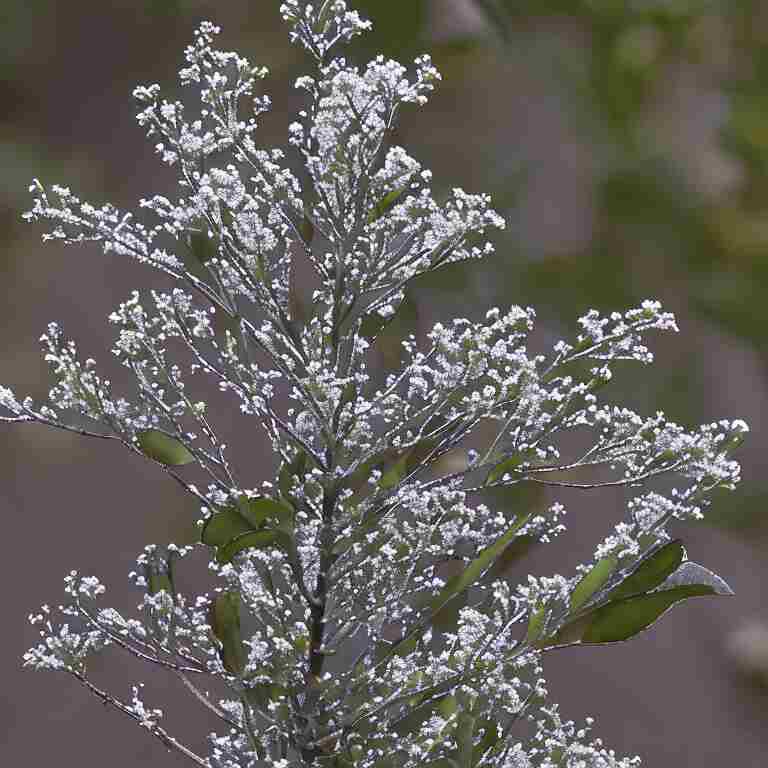} \\

Wild Cat & this feline creature has a compact and agile body, adorned with sharp ears, giving it a fierce and untamed appearance.
&\includegraphics[height=0.12\textwidth, width=0.12\textwidth]{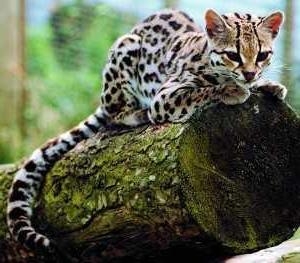}
&\includegraphics[height=0.12\textwidth, width=0.12\textwidth]{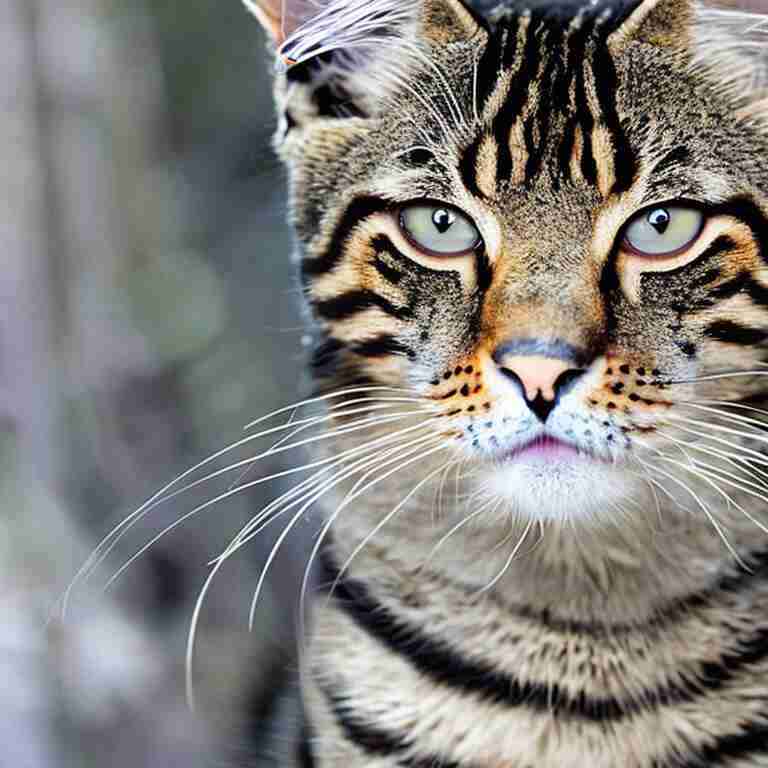}
&\includegraphics[height=0.12\textwidth, width=0.12\textwidth]{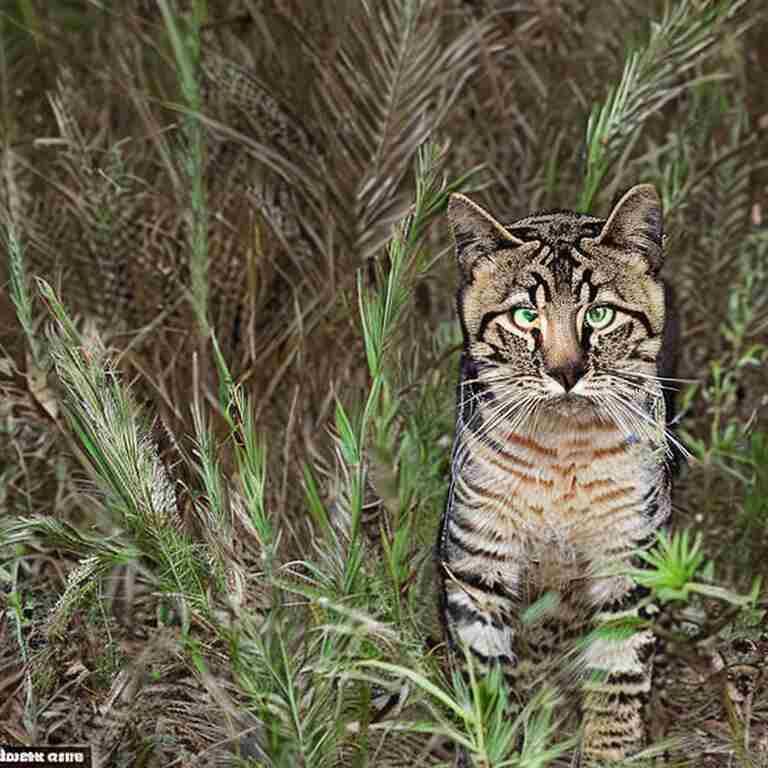}
&\includegraphics[height=0.12\textwidth, width=0.12\textwidth]{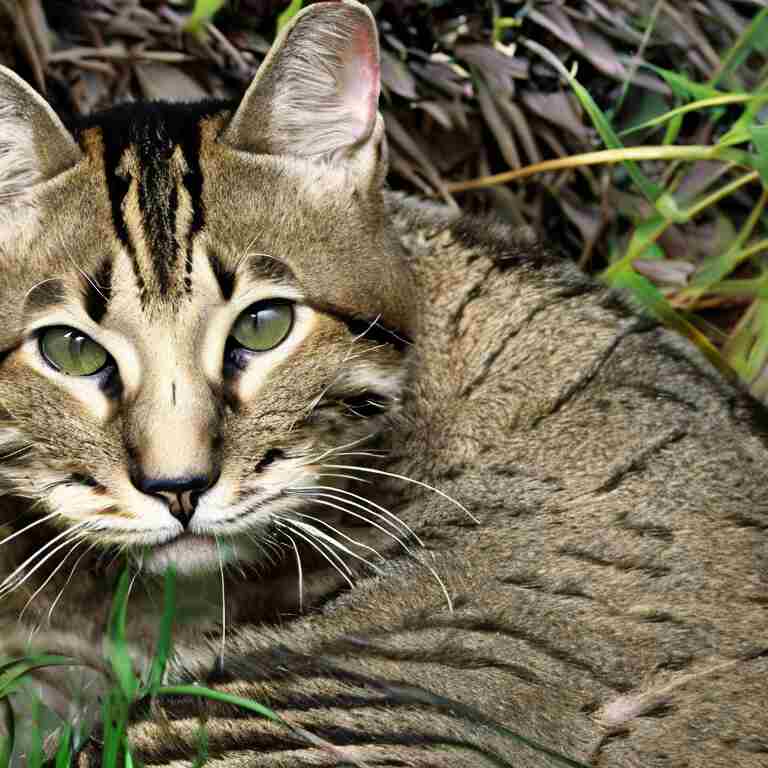} \\

\bottomrule

\end{tabular}
\end{adjustbox}

\vspace{0.5em}
\caption{Overview of samples from description-based image retrieval benchmark where Class description shows improved performance over class name. The first column displays the class name, while the second column shows a description generated by prompting the LLM to provide a sentence about the class without explicitly mentioning its name. Subsequently, we showcase the ground truth image from the training set and two images generated by Stable Diffusion~\cite{sd} utilizing the description text prompts. Finally, an image generated by Stable Diffusion using the prompt with the class name is displayed in the last column. 
\label{tab:dbr_wins}
}
\end{table*}

\section{Additional CLIP Backbones}
\label{sec:clip}

In~\Cref{tab:baseline}, we report the results from three other CLIP backbones. MetaCLIP~\cite{metaclip}, OpenCLIP~\cite{openclip}, and Eva-02 CLIP~\cite{sun2024evaclip18b}. Our method improves the average performance for all three backbones. This clearly proves that the inclusion of visual information outperforms text-only retrieval. 

\begin{table}[t]
\scalebox{0.72}{
\centering
{
\footnotesize
\begin{tabular}{lrrrrrrrrrrrrrrlr}
& \rotatebox{90}{ImageNet} & \rotatebox{90}{DTD} & \rotatebox{90}{Stanford Cars}     & \rotatebox{90}{SUN397}   & \rotatebox{90}{Food}     & \rotatebox{90}{FGVC Aircraft} & \rotatebox{90}{Oxford Pets}     & \rotatebox{90}{Caltech101} & \rotatebox{90}{Flowers 102}  & \rotatebox{90}{UCF101}      & \rotatebox{90}{Kinetics-700}    & \rotatebox{90}{RESISC45}      & \rotatebox{90}{CIFAR-10}      & \rotatebox{90}{CIFAR-100}     & \rotatebox{90}{Places-365} & \rotatebox{90}{Average} \\
\midrule
text-only(M) & 66.5 & 39.8 & 73.8 & 56.0 & 88.5 & 31.4 & 88.7 & 91.5 & 74.9 & 65.7 & 35.9 & 55.4 & 88.5 & 67.2 & 26.6 & 63.4\\
image-only(M) & 34.9 & 20.9 & 35.7 & 35.1 & 58.6 & 15.7 & 52.0 & 74.3 & 35.1 & 46.3 & 18.6 & 33.4 & 74.8 & 48.9 & 16.7 & 40.1\\
text(M) + image(M) & 54.1 & 32.4 & 54.7 & 50.2 & 81.5 & 22.4 & 75.4 & 87.4 & 62.0 & 61.8 & 30.4 & 53.0 & 88.5 & 64.3 & 23.7 & 56.1\\
text(M) + image(D) & 69.9 & 46.0 & 57.2 & 57.2 & 83.7 & 23.9 & 89.0 & 91.8 & 76.9 & 65.6 & 34.3 & 54.7 & 79.0 & 81.7 & 30.3 & 62.7\\
\rowcolor{blond}
Ours & \textbf{74.2} & \textbf{48.5} & \textbf{74.1} & \textbf{63.2} & \textbf{90.3} & \textbf{33.7} & \textbf{91.5} & \textbf{94.3} & \textbf{79.3} & \textbf{71.9} & \textbf{40.3} & \textbf{59.2} & \textbf{90.2} & \textbf{83.0} & \textbf{32.2} & \textbf{68.4}\\
\midrule

text-only(O) & 63.4 & 38.6 & \textbf{83.1} & 59.8 & 86.8 & 20.9 & 86.2 & 90.9 & 71.0 & 65.7 & 32.7 & 67.1 & 93.3 & 70.0 & 29.6 & 63.9\\
image-only(O) & 41.0 & 31.2 & 52.6 & 45.7 & 66.1 & 13.9 & 58.2 & 79.9 & 50.1 & 50.8 & 21.4 & 51.2 & 82.2 & 53.9 & 22.8 & 48.1\\
text(O)+image(O) & 56.1 & 39.8 & 70.4 & 56.2 & 80.6 & 19.3 & 75.6 & 87.6 & 65.3 & 61.2 & 29.9 & 62.3 & 92.1 & 66.8 & 28.4 & 59.4\\
text(O)+image(D) & 70.3 & 46.3 & 50.7 & 59.5 & 82.9 & 16.3 & 88.2 & 92.7 & 73.8 & 66.7 & 34.6 & 62.8 & 94.5 & 84.9 & 32.3 & 63.8\\
\rowcolor{blond}
Ours & \textbf{70.8} & \textbf{45.0} & 82.2 & \textbf{63.4} & \textbf{89.0} & \textbf{23.0} & \textbf{89.2} & \textbf{93.1} & \textbf{74.9} & \textbf{69.8} & \textbf{36.9} & \textbf{67.8} & \textbf{94.5} & \textbf{80.0} & \textbf{32.7} & \textbf{67.5} \\
\midrule

text-only(E) & 71.1 & 43.9 & \textbf{83.6} & 62.2 & 90.4 & 30.6 & 89.4 & 93.6 & 77.8 & 70.0 & 42.6 & 68.0 & \textbf{97.4} & 86.9 & 29.5 & 69.1\\
image-only(E) & 39.8 & 24.7 & 50.5 & 41.4 & 65.7 & 14.9 & 60.2 & 78.6 & 55.9 & 50.0 & 23.5 & 43.4 & 81.6 & 69.3 & 19.6 & 47.9\\
text(E)+image(E) & 55.9 & 32.5 & 66.9 & 52.7 & 83.1 & 20.9 & 78.1 & 87.4 & 71.1 & 60.5 & 32.9 & 57.7 & 88.8 & 79.8 & 25.7 & 59.6\\
text(E)+image(D) & 69.8 & 44.9 & 45.8 & 57.8 & 81.9 & 17.0 & 87.9 & 91.9 & 73.7 & 65.6 & 34.2 & 58.5 & 91.4 & 85.4 & 31.1 & 62.4\\
\rowcolor{blond}
Ours & \textbf{75.9} & \textbf{51.2} & 82.4 & \textbf{65.0} & \textbf{91.3} & \textbf{31.2} & \textbf{90.7} & \textbf{94.6} & \textbf{79.9} & \textbf{73.1} & \textbf{43.0} & \textbf{68.5} & 95.3 & \textbf{89.8} & \textbf{33.5} & \textbf{71.0}\\
\midrule

\end{tabular}
}
}
\vspace{0.5em}
\caption{Performance comparison on retrieval across 15 benchmark datasets. We report the type of text query used per variant in addition to the encoders for the cross-modal (text-to-image) and intra-modal (image-to-image) similarity. Performance is measured via Mean average precision (mAP). 
M: MetaCLIP~\cite{metaclip}, O: OpenCLIP~\cite{openclip}, E:Eva-02CLIP~\cite{sun2024evaclip18b}, D: DINOv2.}
\label{tab:baseline}
\vspace{-0pt}

\end{table}

\begin{figure}
\centering
\vspace{-5pt}
    \begin{tikzpicture}
        \begin{axis}[
            xlabel={Corruption Level},
            ylabel={\% Drop in mAP Performance},
            grid=both,
            legend style={at={(0.05,0.95)}, anchor=north west},
            width=0.45\textwidth,
            height=0.25\textheight,
            ymin=0, ymax=35
        ]
        
        \addplot[color=blue, mark=o, solid] coordinates {
            (0, 0) (1, 6.3) (2, 11) (3, 15.2) (4, 22) (5, 30.9)
        };
        \addlegendentry{Text Only}
        
        \addplot[color=red, mark=s, dashed] coordinates {
            (0, 0) (1, 3.9) (2, 7.2) (3, 10) (4, 15.1) (5, 23.5)
        };
        \addlegendentry{Ours}
        
        \end{axis}
    \end{tikzpicture}
\vspace{5pt}
\caption{Comparison of performance drop with increasing corruption severity. The text-only baseline shows a larger performance drop than our method as corruption levels rise. \label{fig:noise}
\vspace{-5pt}
}
\end{figure}
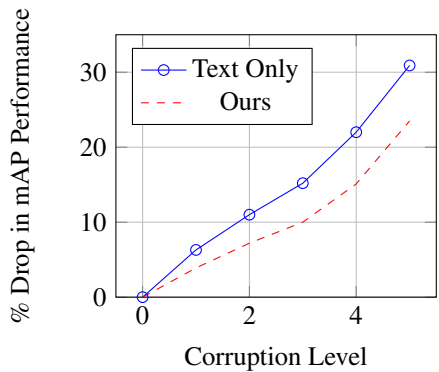

\input{images/figures/corruption}

\section{Testing on Noisy Databases}
\label{sec:rob}
In this section, we compare and analyze the robustness of our approach compared to the text-only baseline. We experiment on ImageNet-C~\cite{hendrycks2019robustness}. The ImageNet-C dataset is a benchmark for evaluating the robustness against common corruptions, such as noise, blur, weather effects, and digital distortions. It consists of various corruptions applied to the original ImageNet validation images at five severity levels. In~\Cref{fig:corruption}, we show results for all five levels, starting from level one at the top and level five at the bottom. While both approaches exhibit a performance drop as corruption severity increases, our method demonstrates significantly greater robustness. At level 1 corruption, the text-only baseline performance drops by $6.3\%$, whereas our approach shows a smaller drop of $3.9\%$. As corruption severity escalates, the gap becomes more pronounced: at level 5, the text-only baseline experiences a $30.9\%$ drop, compared to a $23.5\%$ drop for our method. This trend shows the effectiveness of our approach in mitigating the impact of increasing corruption levels. This trend is visually compared in~\Cref{fig:noise}.

\begin{table}
\newcolumntype{M}[1]{>{\centering\arraybackslash}m{#1}}
\setlength{\tabcolsep}{8pt}
\renewcommand{\arraystretch}{1.5}
   \begin{adjustbox}{width=0.95\textwidth}
   \begin{tabular}
   {M{3cm}M{4cm}M{2.1cm}M{1.5cm}M{1.5cm}M{1.5cm}M{1.5cm}}
   \toprule
Class Name & Description based text & Ground Truth & \multicolumn{2}{c}{Description Based} & \multicolumn{1}{c}{Class Name}\\\midrule

Headphone & a device for listening to audio usually includes two compact speakers connected to a band worn on the head.
&\includegraphics[height=0.12\textwidth, width=0.12\textwidth]{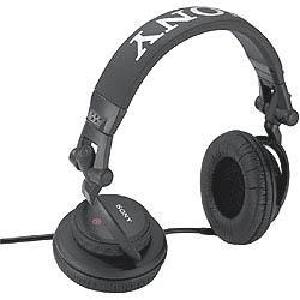}
&\includegraphics[height=0.12\textwidth, width=0.12\textwidth]{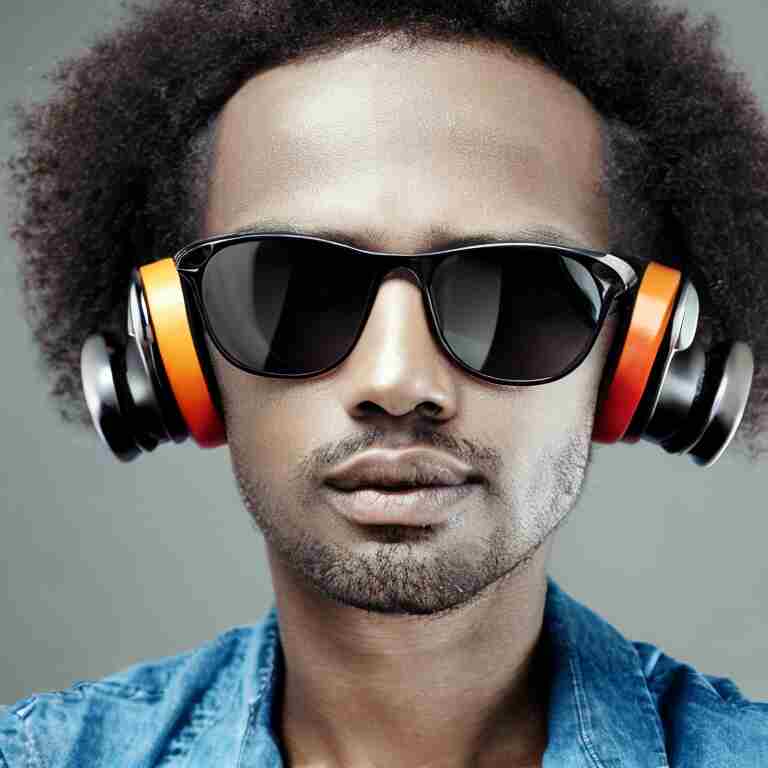}
&\includegraphics[height=0.12\textwidth, width=0.12\textwidth]{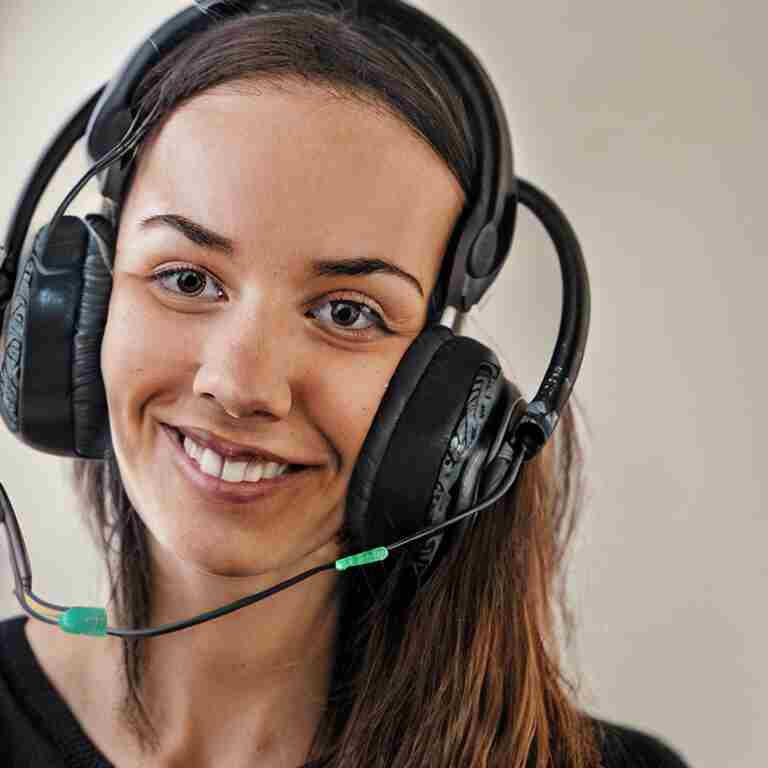}
&\includegraphics[height=0.12\textwidth, width=0.12\textwidth]{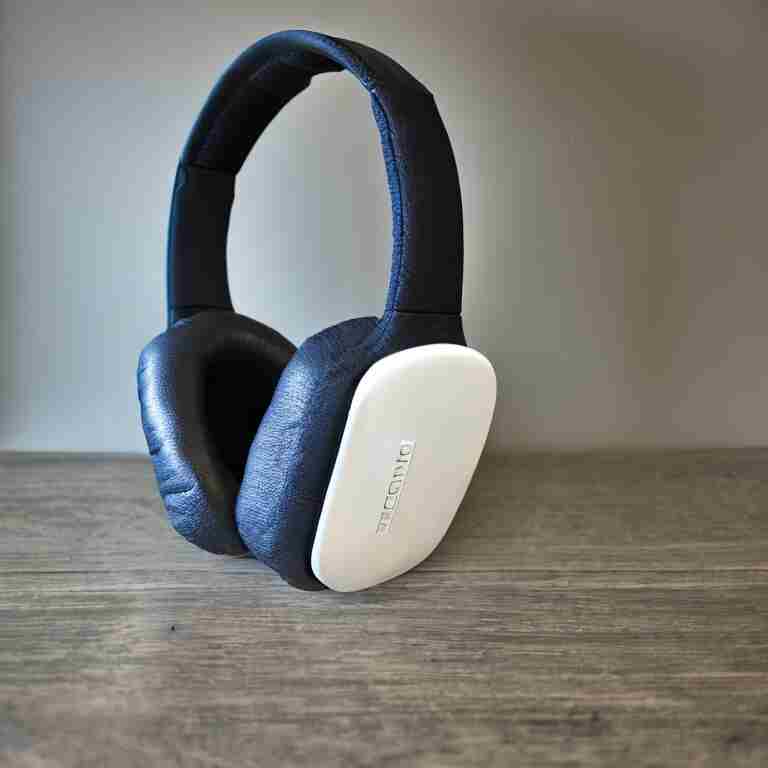} \\

Butterfly & the insect with two large wings adorned in vibrant scales is often seen fluttering through gardens.
&\includegraphics[height=0.12\textwidth, width=0.12\textwidth]{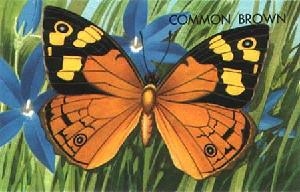}
&\includegraphics[height=0.12\textwidth, width=0.12\textwidth]{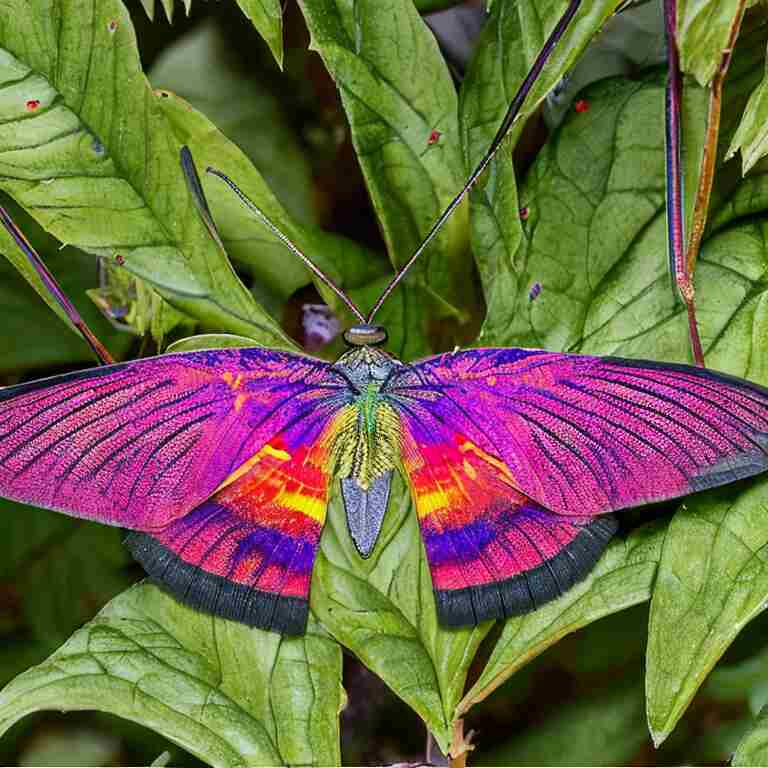}
&\includegraphics[height=0.12\textwidth, width=0.12\textwidth]{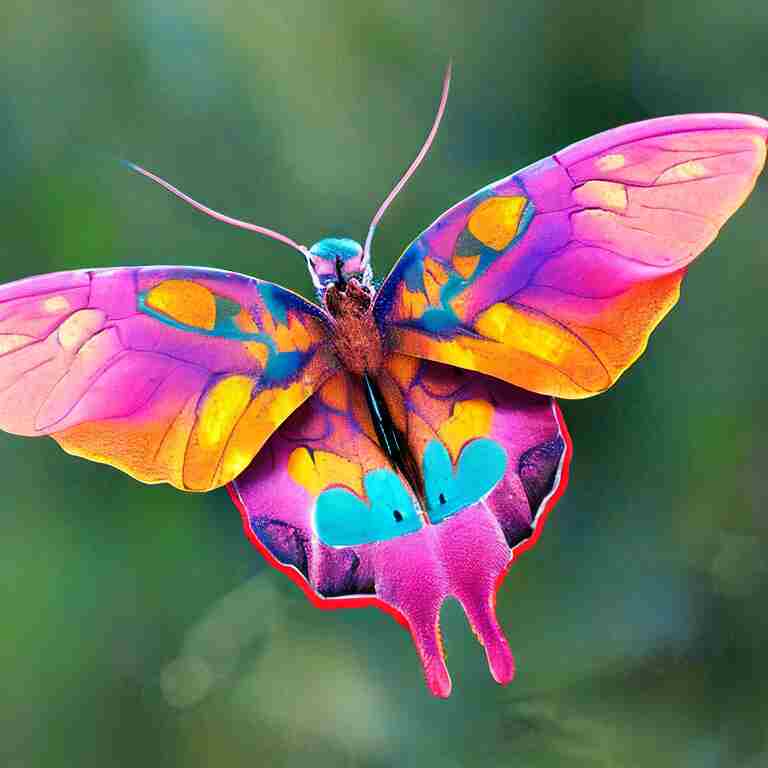}
&\includegraphics[height=0.12\textwidth, width=0.12\textwidth]{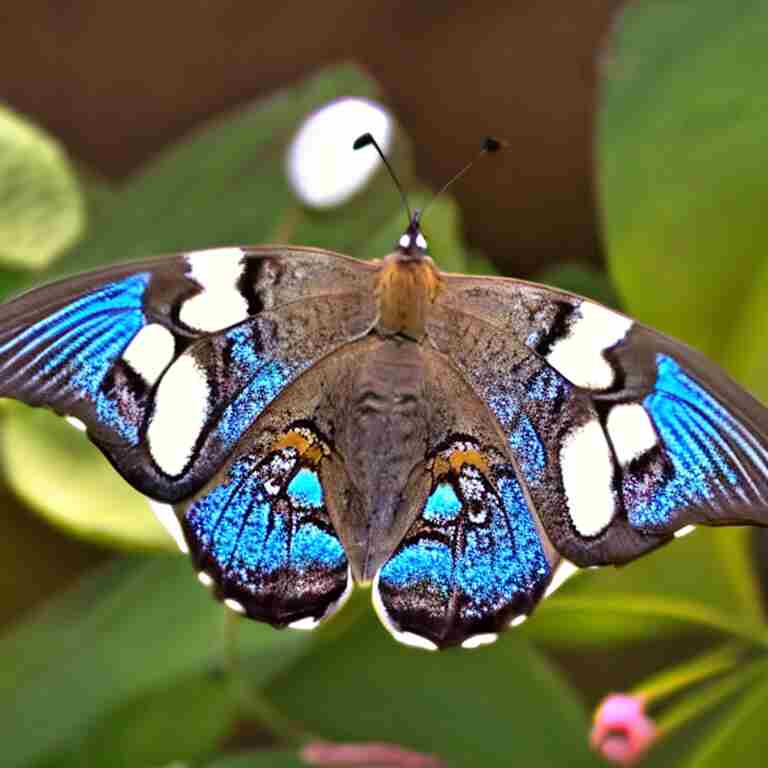} \\

Airplane & a flying vehicle made of metal and equipped with wings is commonly used for air travel.
&\includegraphics[height=0.12\textwidth, width=0.12\textwidth]{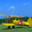}
&\includegraphics[height=0.12\textwidth, width=0.12\textwidth]{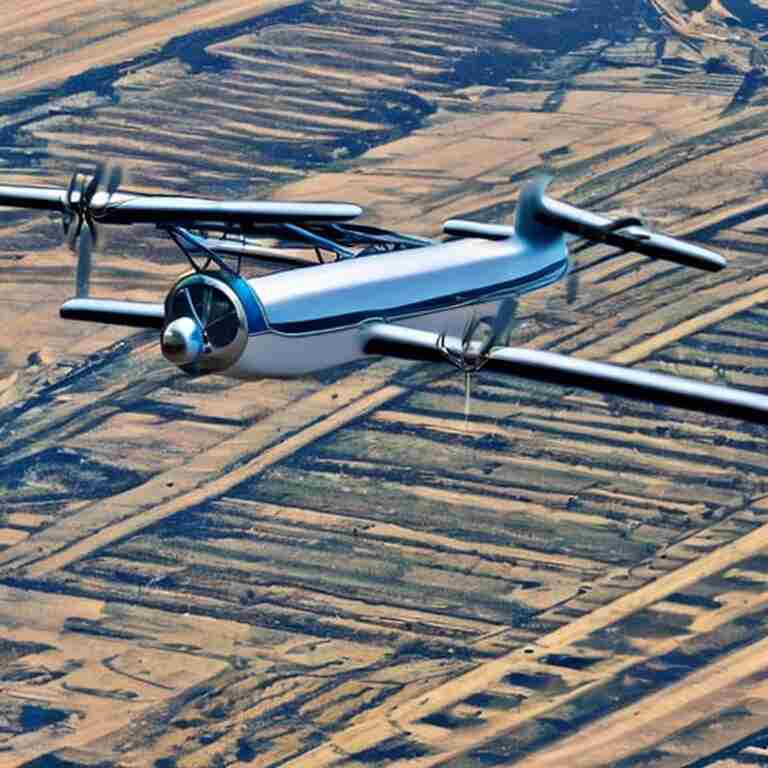}
&\includegraphics[height=0.12\textwidth, width=0.12\textwidth]{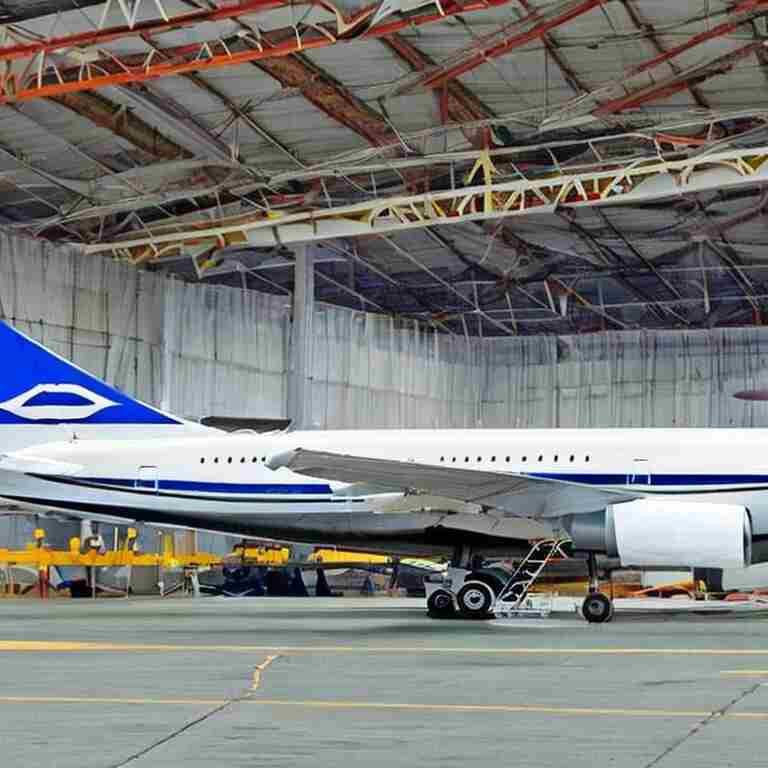}
&\includegraphics[height=0.12\textwidth, width=0.12\textwidth]{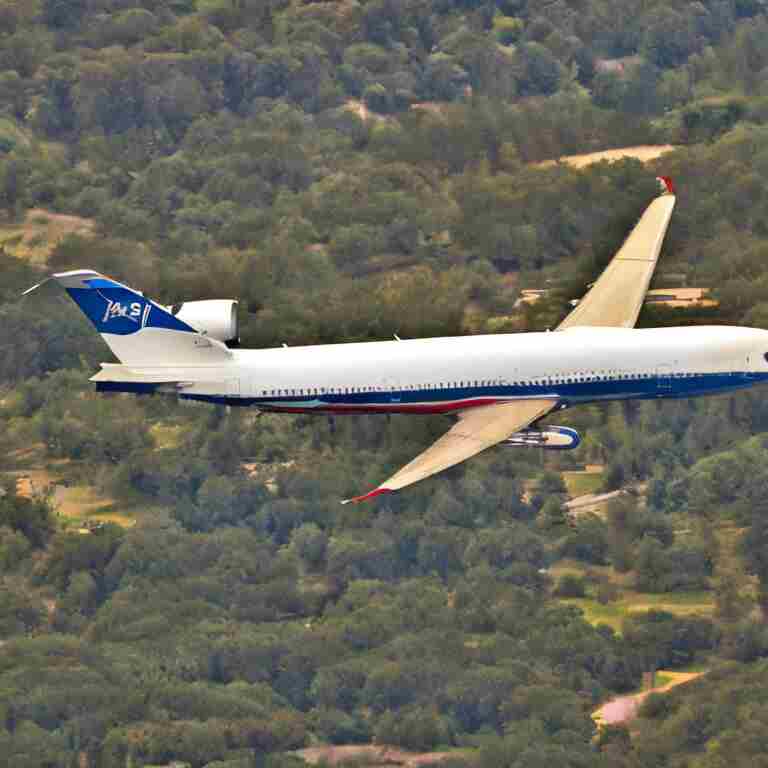} \\

Ship & these vessels are commonly used for transportation or carrying cargo across the ocean.
&\includegraphics[height=0.12\textwidth, width=0.12\textwidth]{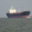}
&\includegraphics[height=0.12\textwidth, width=0.12\textwidth]{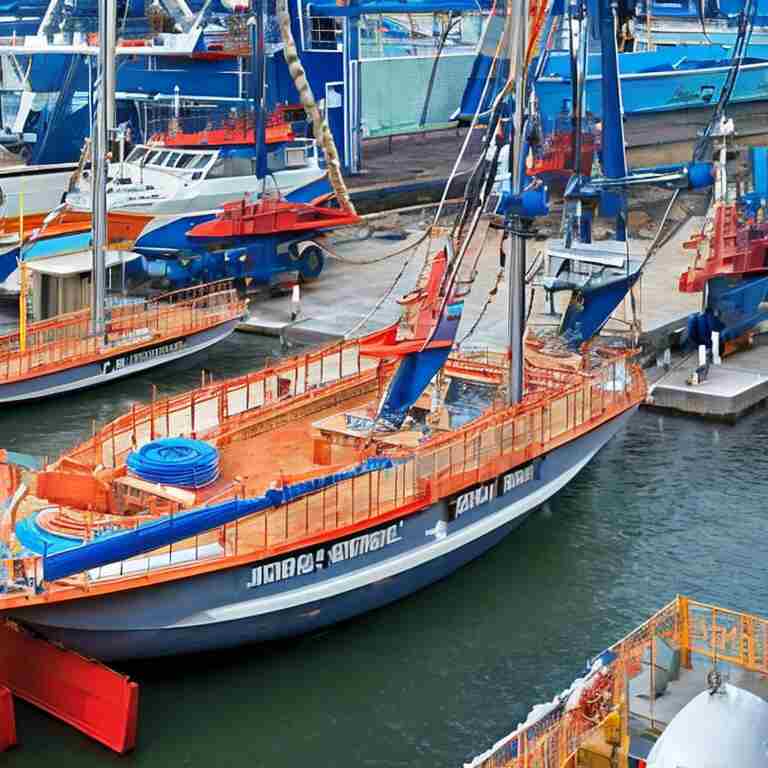}
&\includegraphics[height=0.12\textwidth, width=0.12\textwidth]{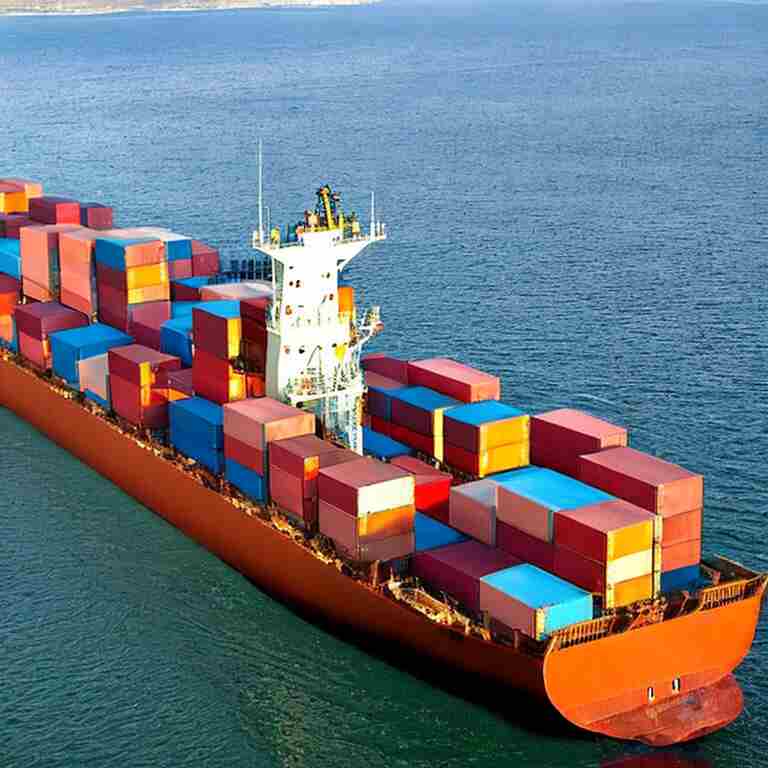}
&\includegraphics[height=0.12\textwidth, width=0.12\textwidth]{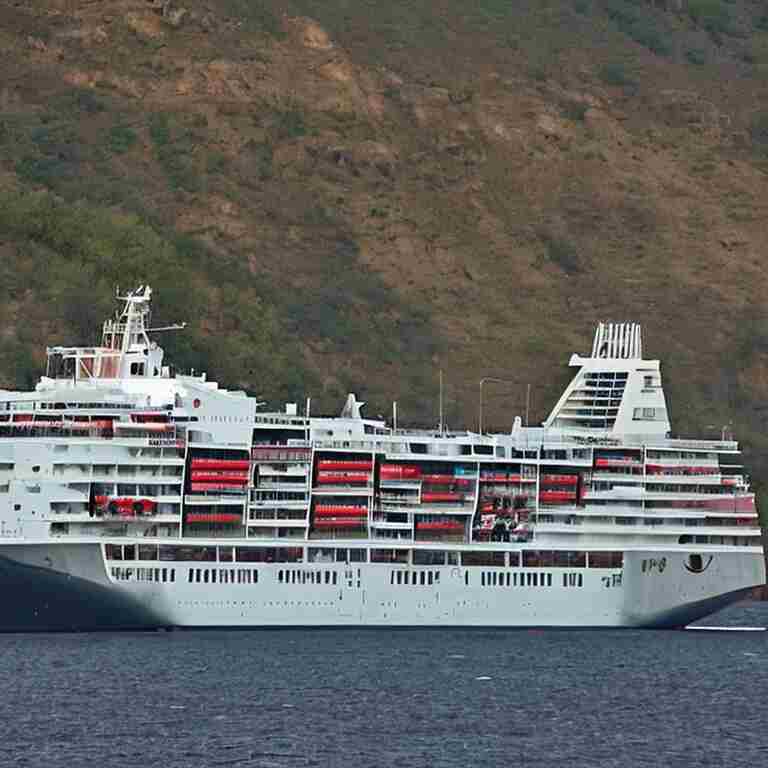} \\

Meshed & the material has a textured surface with tiny gaps scattered
throughout.
&\includegraphics[height=0.12\textwidth, width=0.12\textwidth]{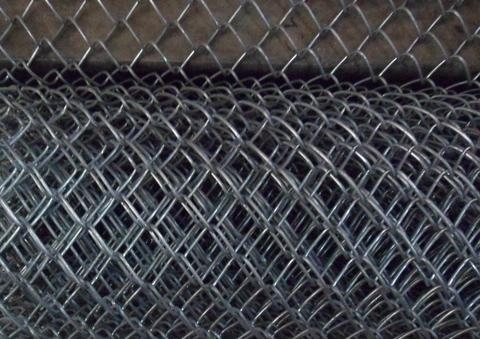}
&\includegraphics[height=0.12\textwidth, width=0.12\textwidth]{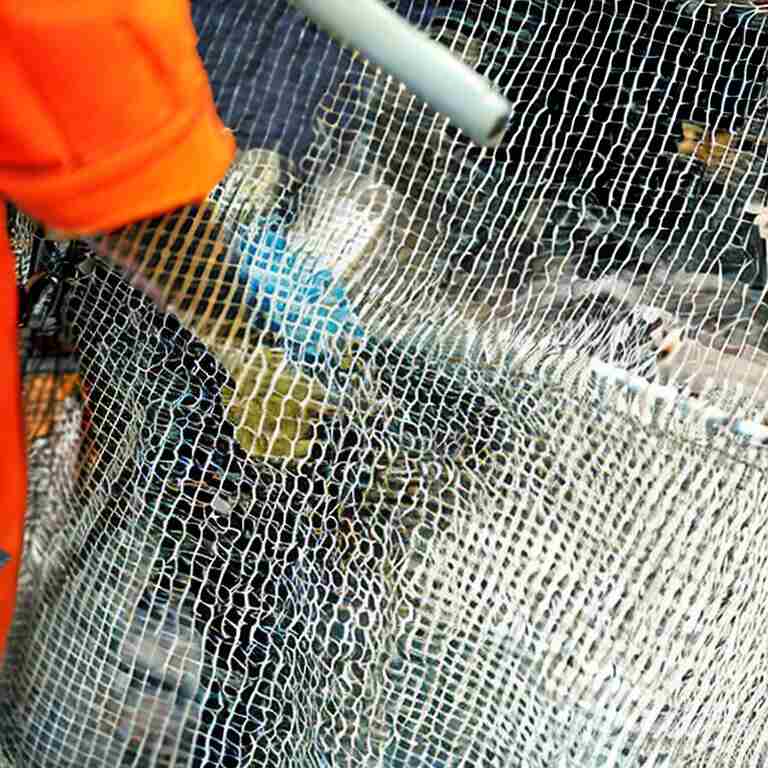}
&\includegraphics[height=0.12\textwidth, width=0.12\textwidth]{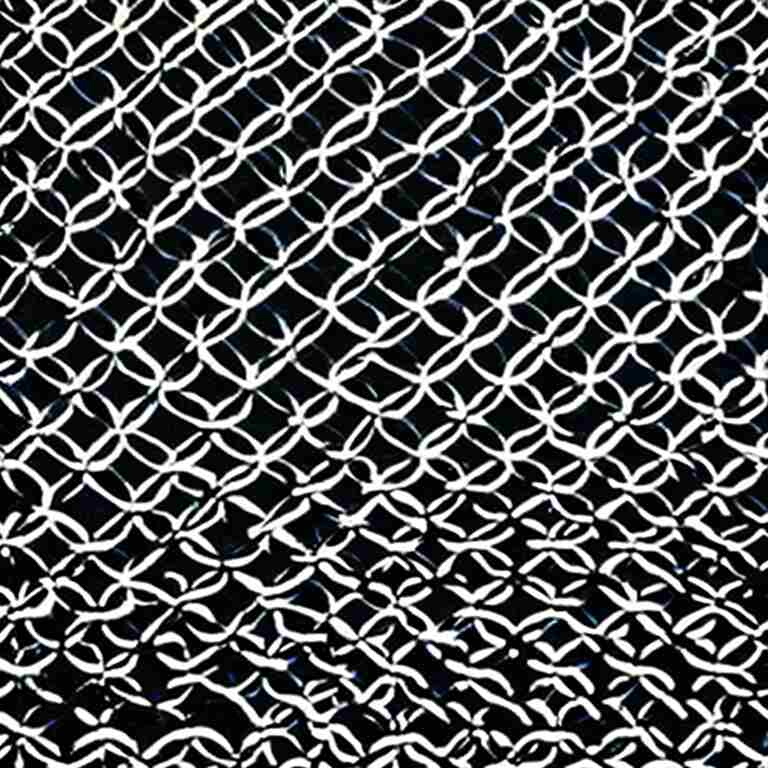}
&\includegraphics[height=0.12\textwidth, width=0.12\textwidth]{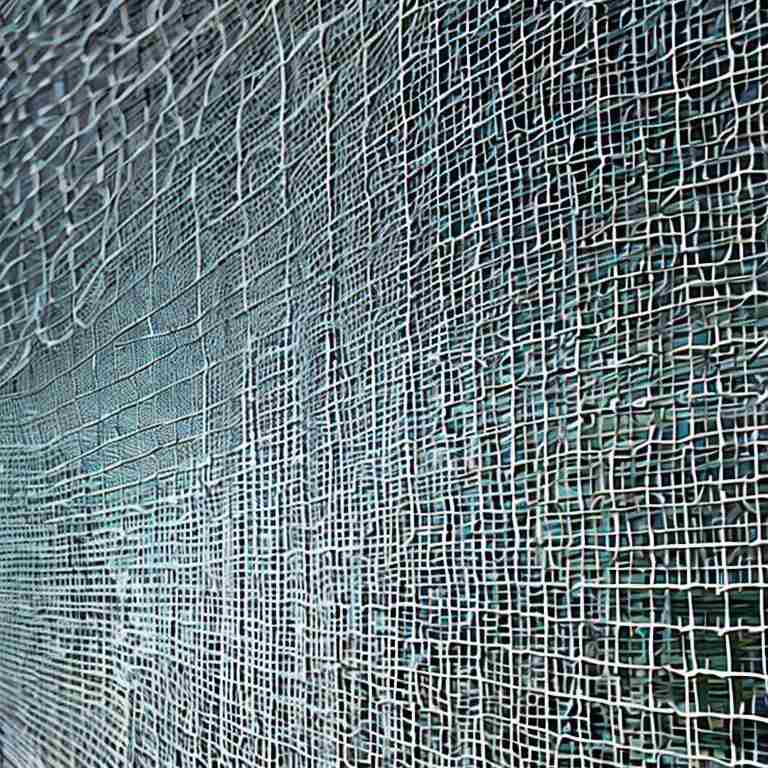} \\

Porous & a material with tiny holes is known for its ability to allow substances to pass through it easily.
&\includegraphics[height=0.12\textwidth, width=0.12\textwidth]{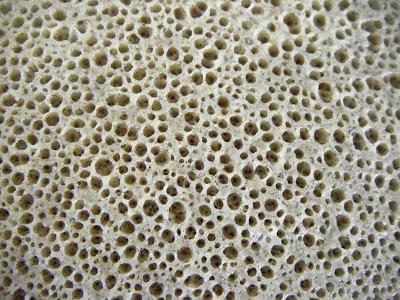}
&\includegraphics[height=0.12\textwidth, width=0.12\textwidth]{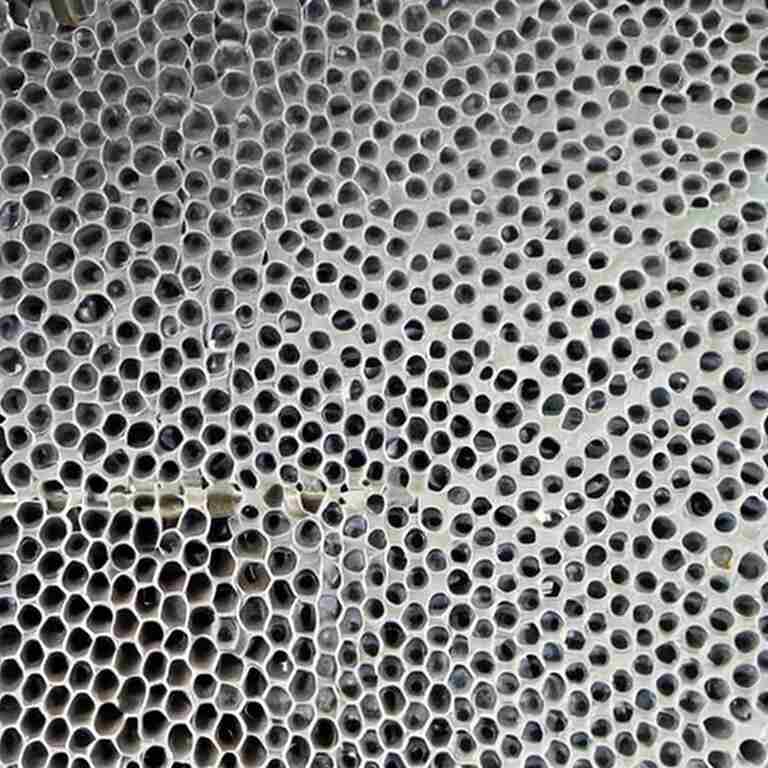}
&\includegraphics[height=0.12\textwidth, width=0.12\textwidth]{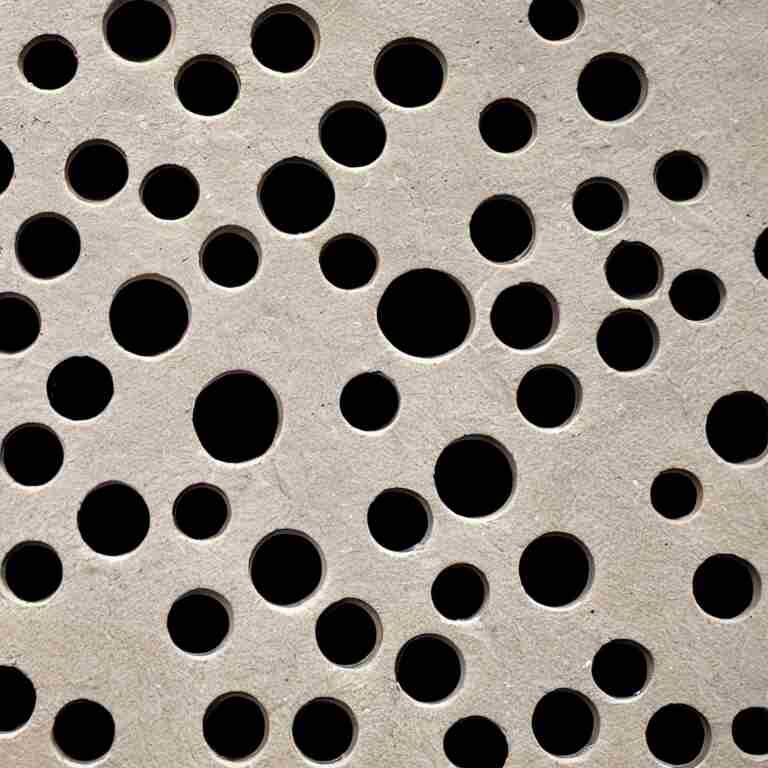}
&\includegraphics[height=0.12\textwidth, width=0.12\textwidth]{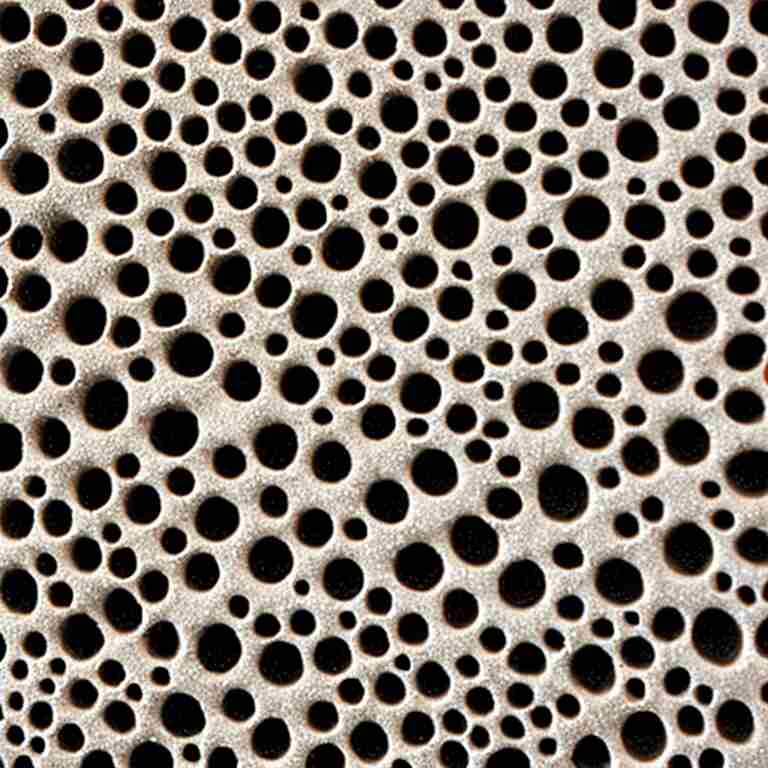} \\

Pink Primrose & the flower’s petals are a deep pink hue, with a bright yellow center.
&\includegraphics[height=0.12\textwidth, width=0.12\textwidth]{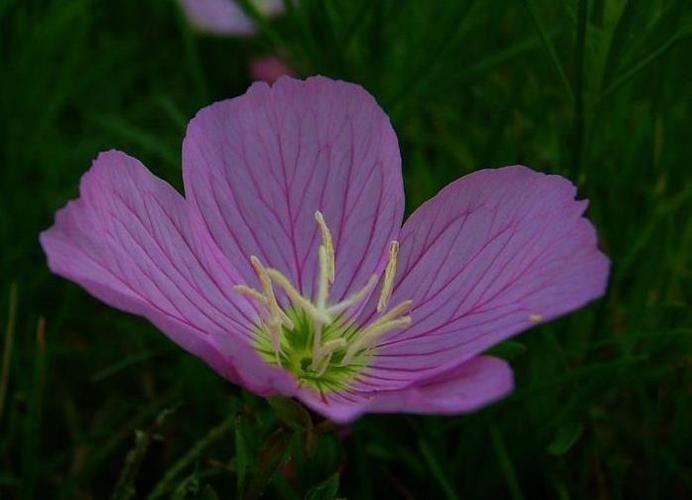}
&\includegraphics[height=0.12\textwidth, width=0.12\textwidth]{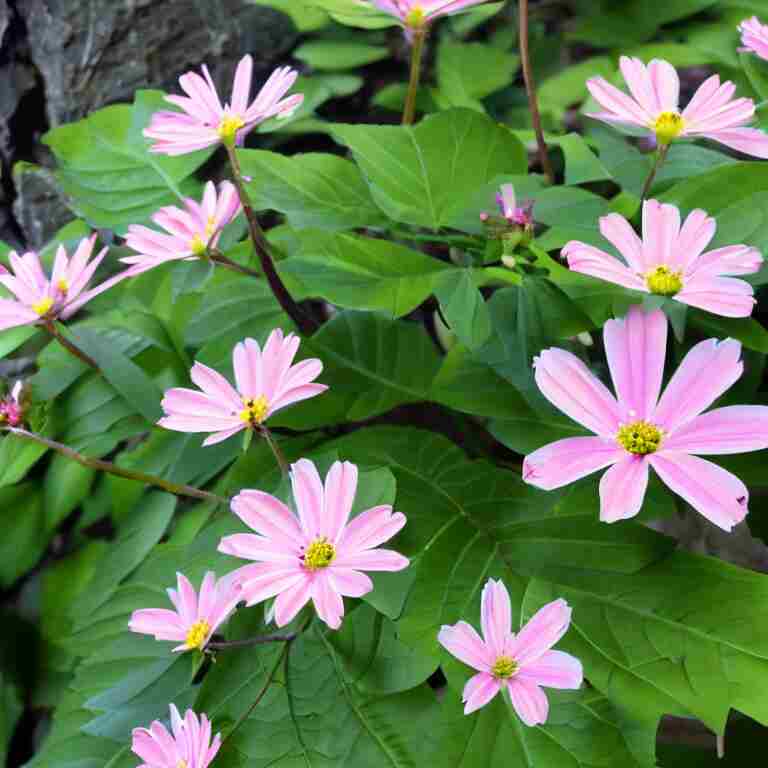}
&\includegraphics[height=0.12\textwidth, width=0.12\textwidth]{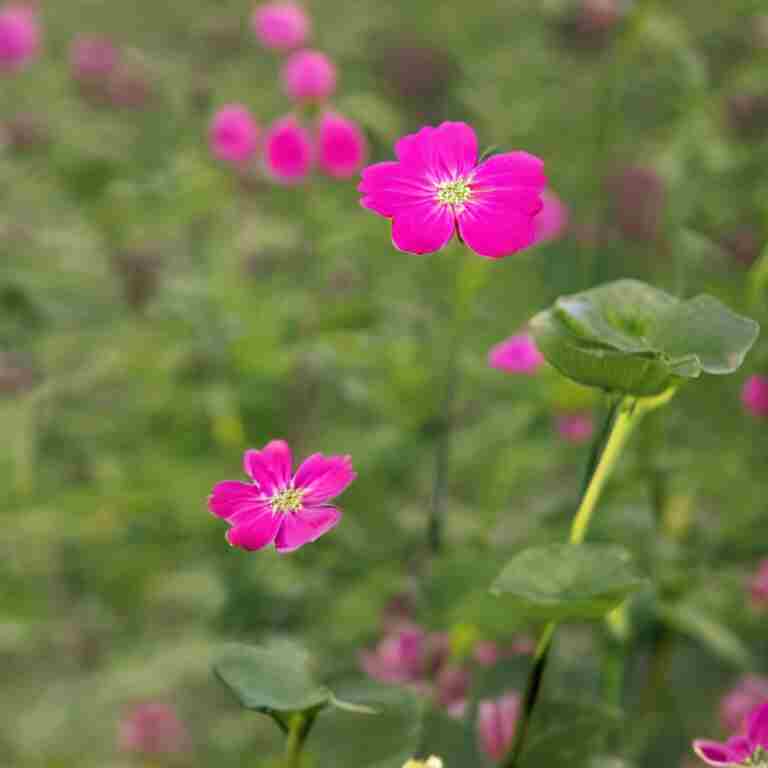}
&\includegraphics[height=0.12\textwidth, width=0.12\textwidth]{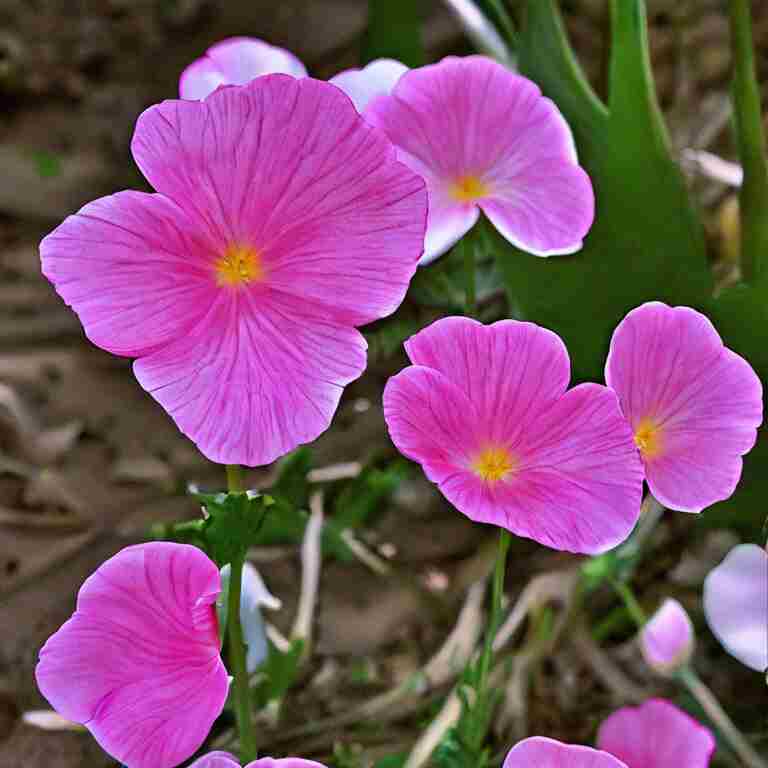} \\

English Marigold & the flower known for its yellow or orange center and red or brown tipped petals is a popular choice among gardeners.
&\includegraphics[height=0.12\textwidth, width=0.12\textwidth]{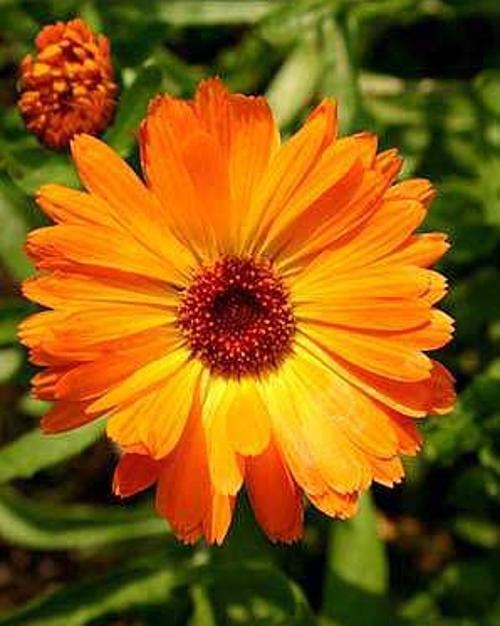}
&\includegraphics[height=0.12\textwidth, width=0.12\textwidth]{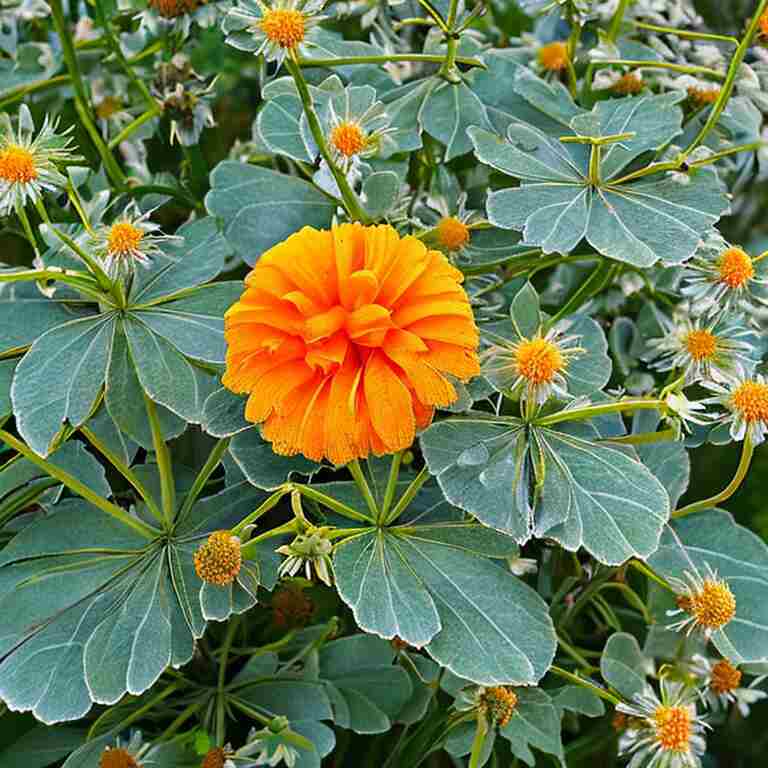}
&\includegraphics[height=0.12\textwidth, width=0.12\textwidth]{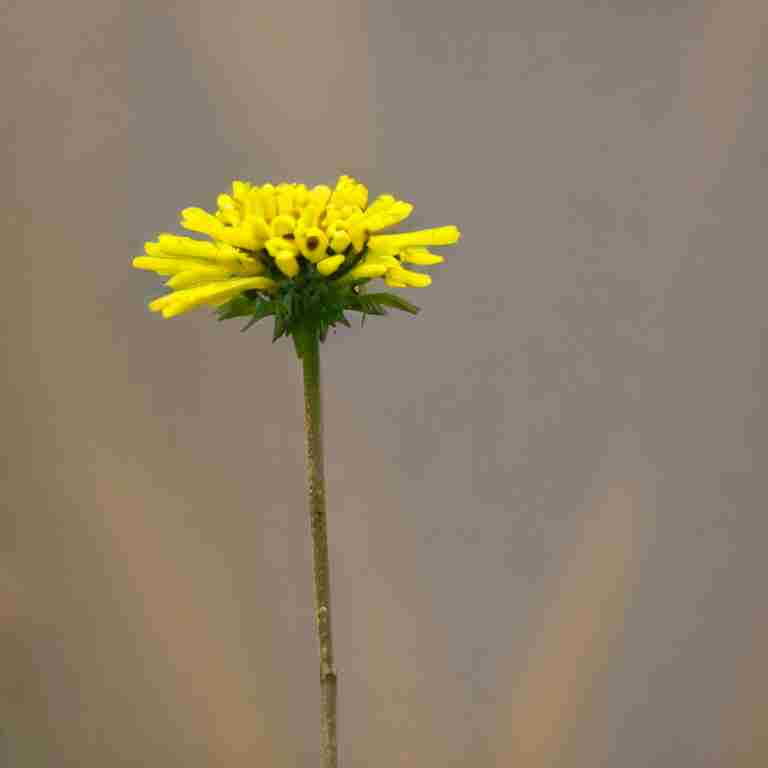}
&\includegraphics[height=0.12\textwidth, width=0.12\textwidth]{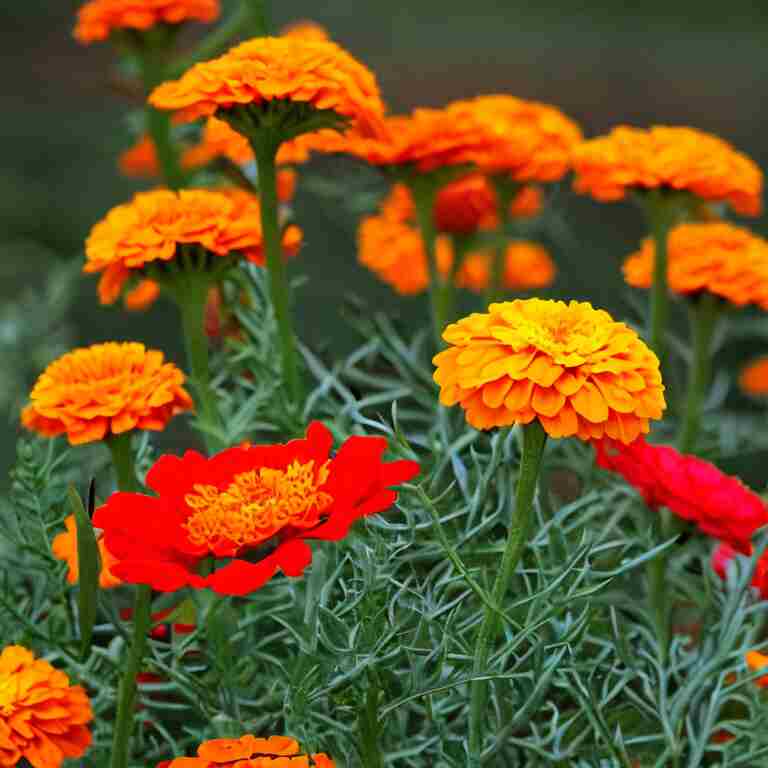} \\

Garlic Bread & the popular dish typically involves a loaf of bread filled with a savory garlic butter mixture.
&\includegraphics[height=0.12\textwidth, width=0.12\textwidth]{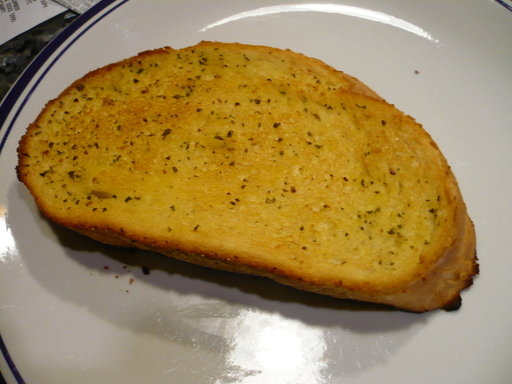}
&\includegraphics[height=0.12\textwidth, width=0.12\textwidth]{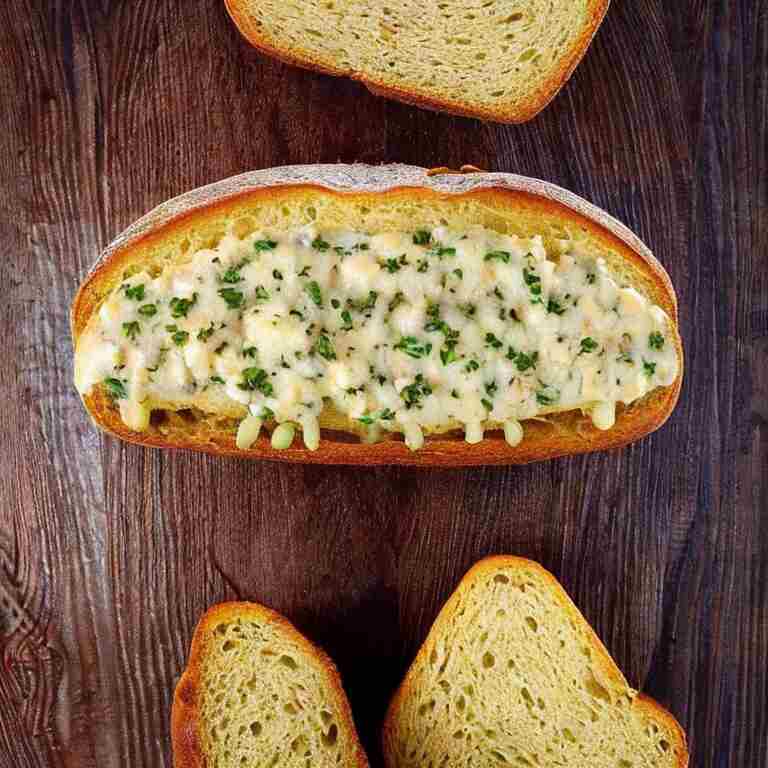}
&\includegraphics[height=0.12\textwidth, width=0.12\textwidth]{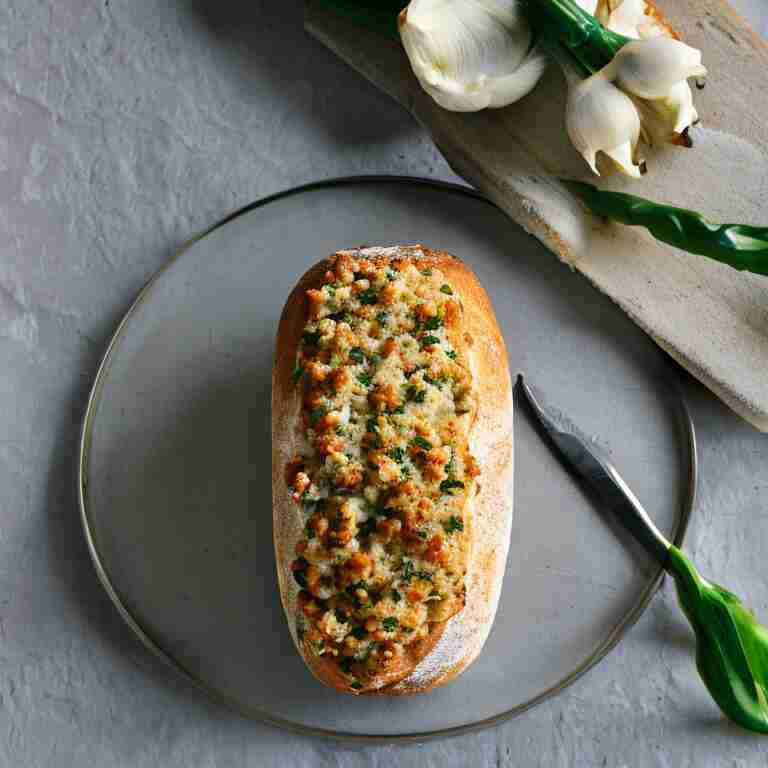}
&\includegraphics[height=0.12\textwidth, width=0.12\textwidth]{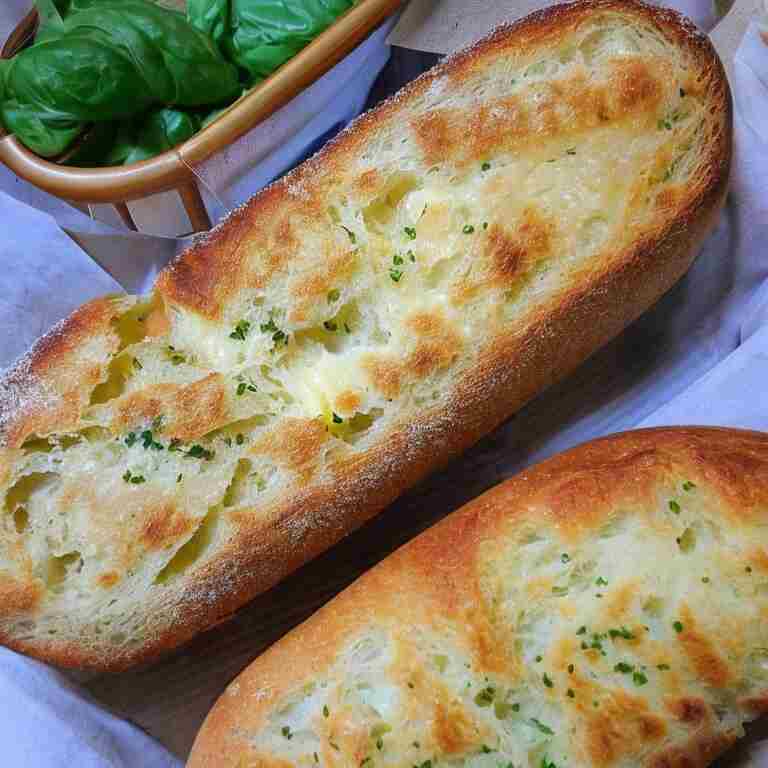} \\

Chicken Wings & they are small, drumstick-shaped pieces of poultry that are typically fried or baked.
&\includegraphics[height=0.12\textwidth, width=0.12\textwidth]{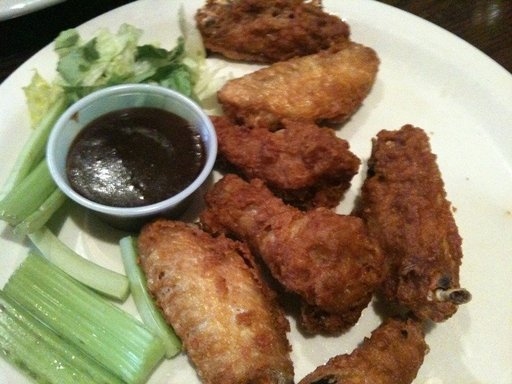}
&\includegraphics[height=0.12\textwidth, width=0.12\textwidth]{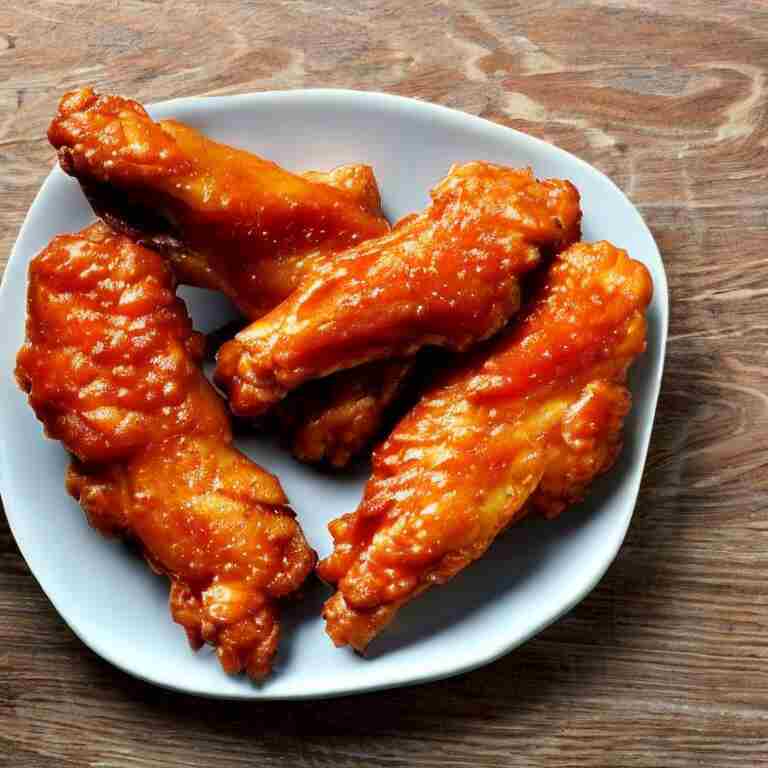}
&\includegraphics[height=0.12\textwidth, width=0.12\textwidth]{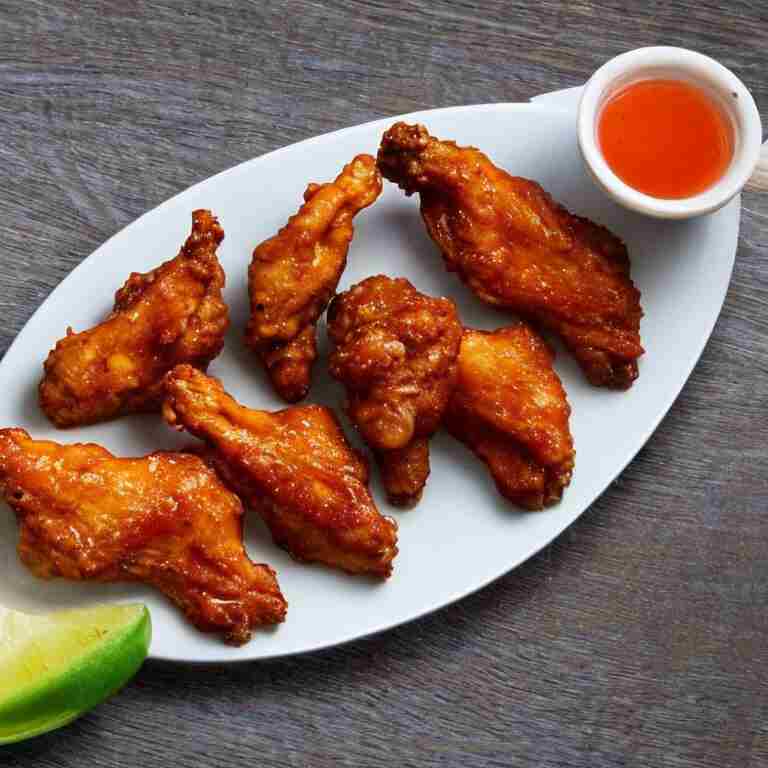}
&\includegraphics[height=0.12\textwidth, width=0.12\textwidth]{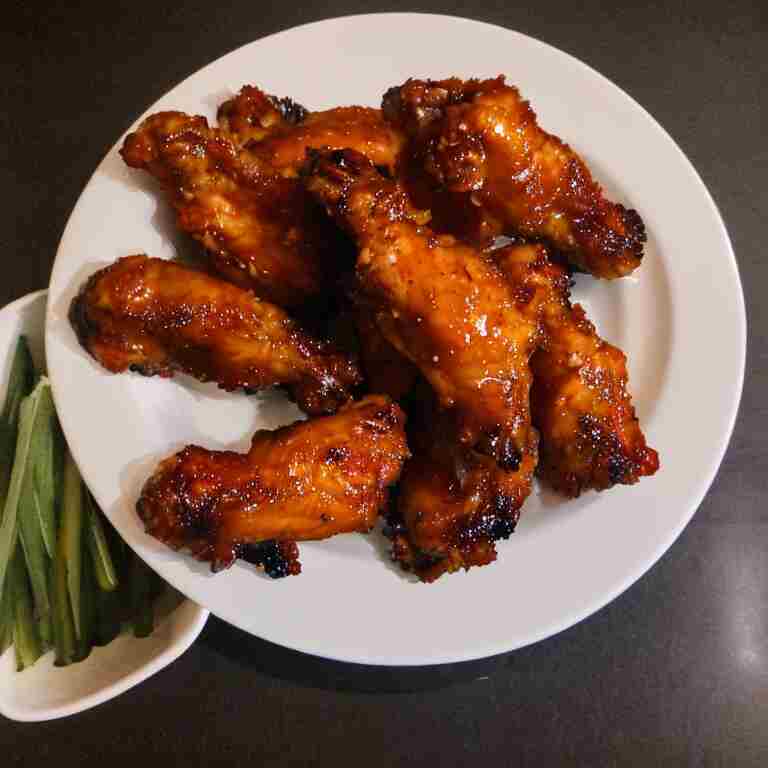} \\

Sphynx & his unique breed of feline has a hairless appearance, resembling a cat without its typical fur coat.
&\includegraphics[height=0.12\textwidth, width=0.12\textwidth]{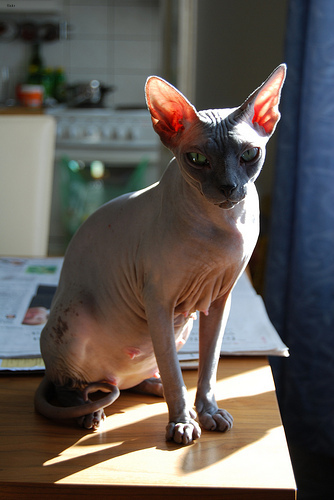}
&\includegraphics[height=0.12\textwidth, width=0.12\textwidth]{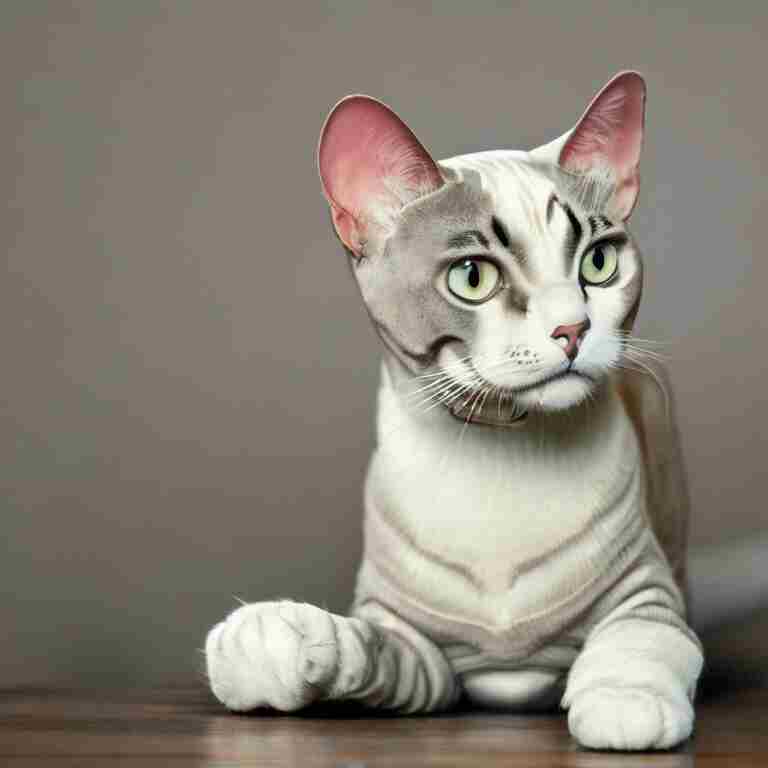}
&\includegraphics[height=0.12\textwidth, width=0.12\textwidth]{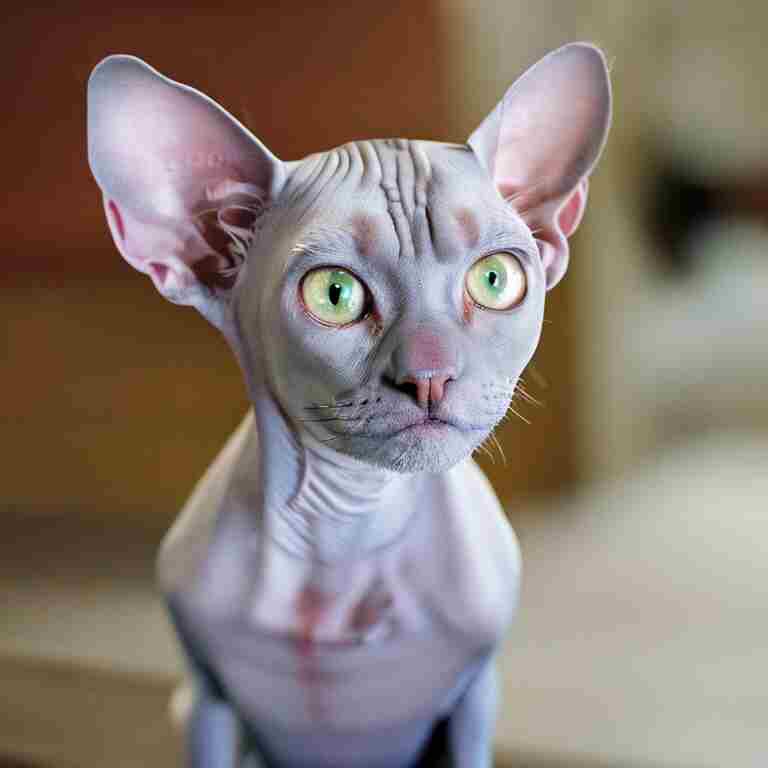}
&\includegraphics[height=0.12\textwidth, width=0.12\textwidth]{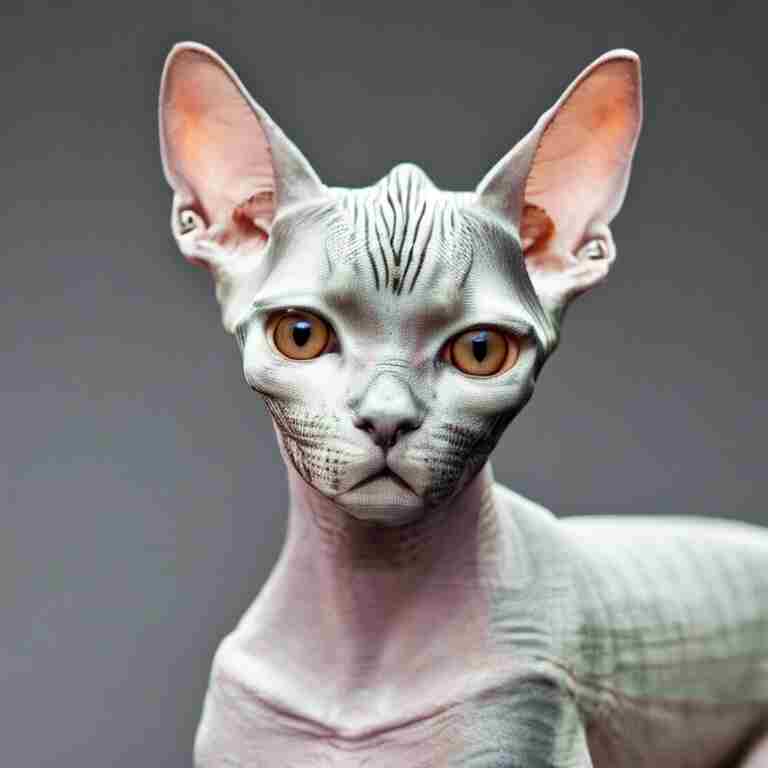} \\

Bengal & the domesticated cat that resembles a small leopard is known for its distinctive markings and sleek appearance.
&\includegraphics[height=0.12\textwidth, width=0.12\textwidth]{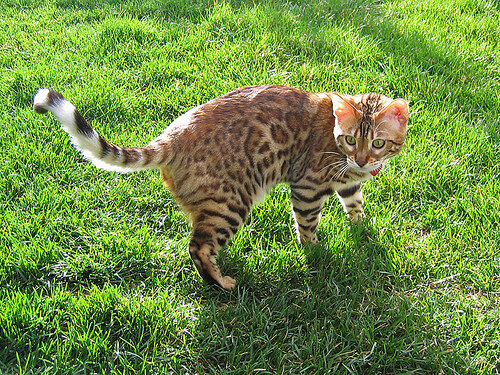}
&\includegraphics[height=0.12\textwidth, width=0.12\textwidth]{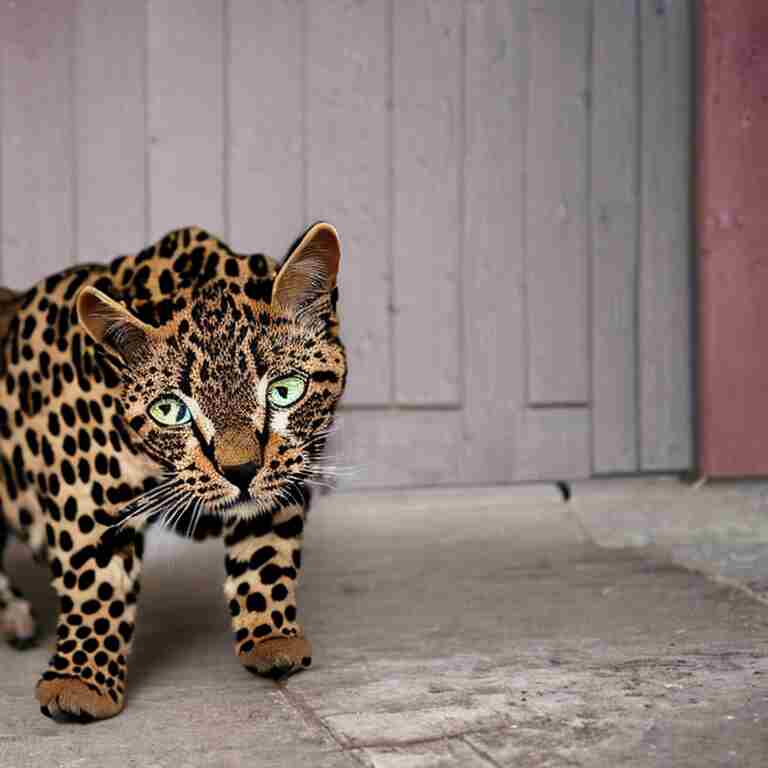}
&\includegraphics[height=0.12\textwidth, width=0.12\textwidth]{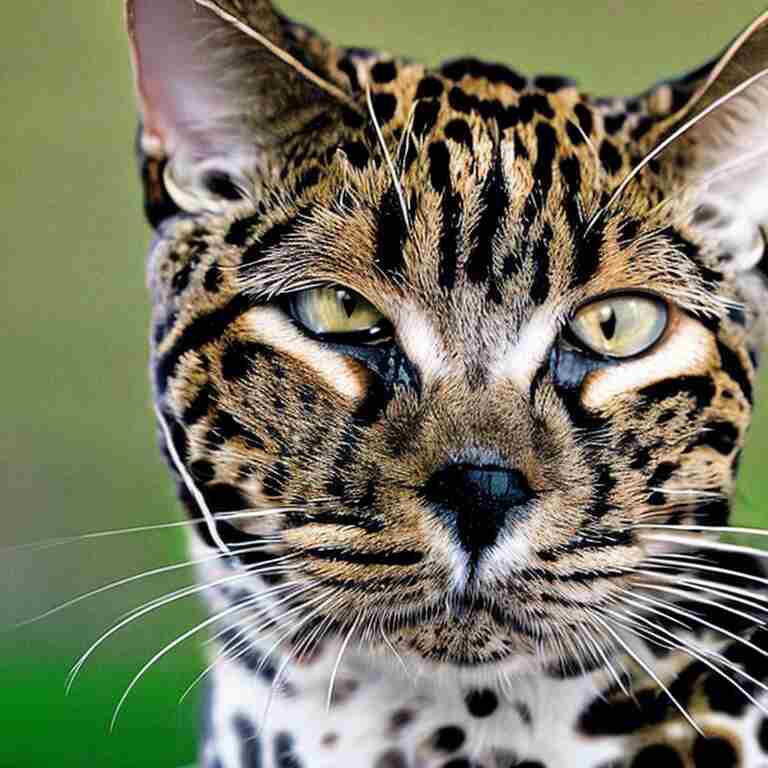}
&\includegraphics[height=0.12\textwidth, width=0.12\textwidth]{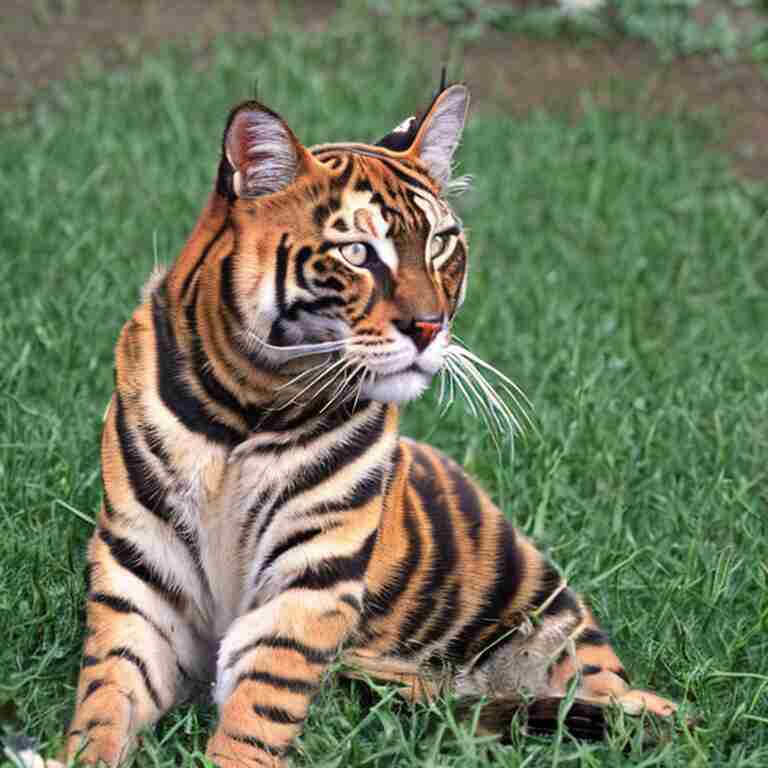} \\

\bottomrule

\end{tabular}
\end{adjustbox}

\vspace{0.5em}
\caption{Additional examples to show qualitative samples generated from the Class description compared to the class name.
\label{tab:description_based}
}
\end{table}

\section{Retrieved Examples for Class and Descriptions Based Retrieval}
\label{sec:qual_top10ret}
\begin{figure}[t]
    \includegraphics[width=\textwidth]{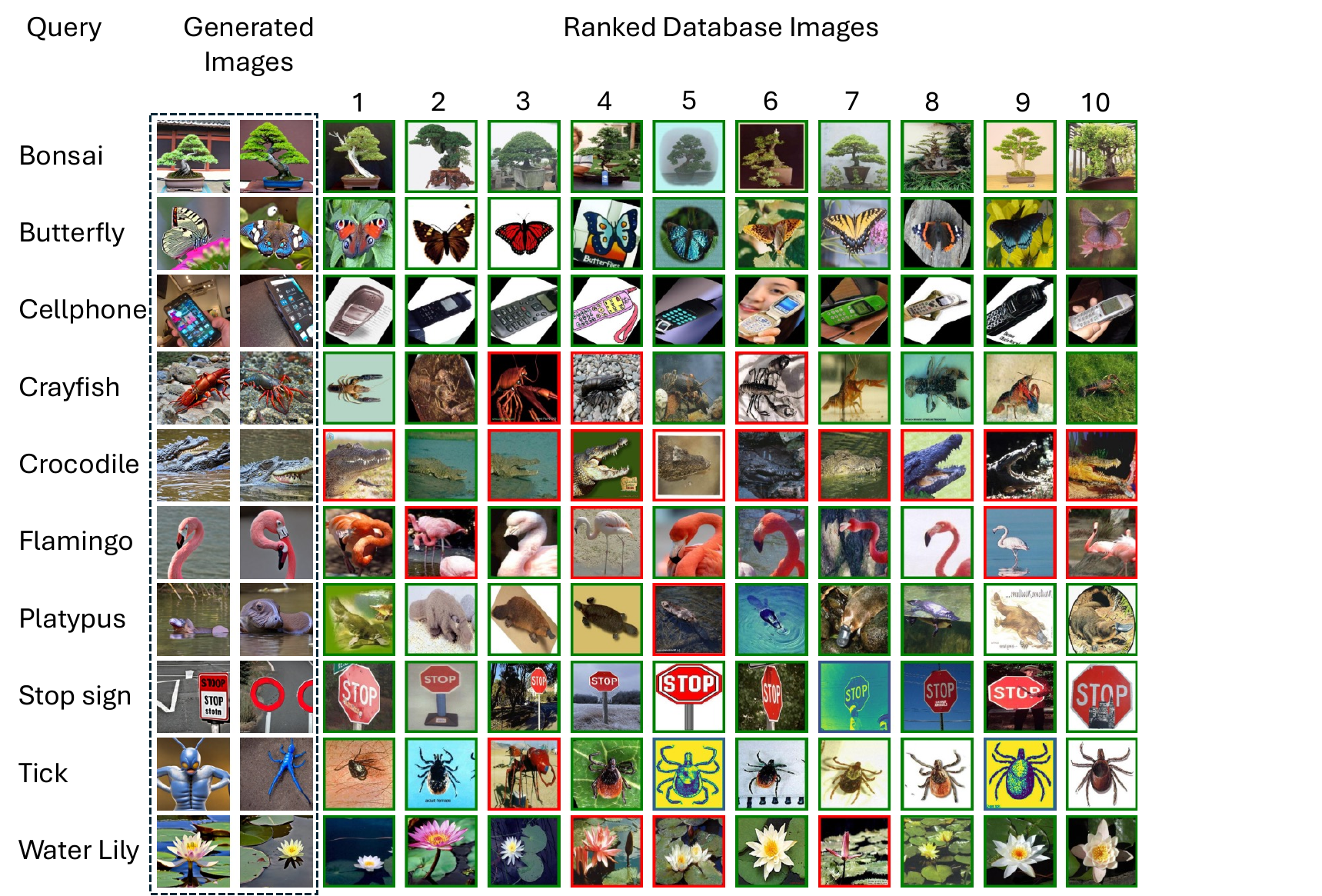}
    
    \caption{
    Class-based retrieval for Caltech101.
    \label{fig:cal_class}
    \vspace{-20pt}
    }
    \end{figure}

\begin{figure}[t]
    
    \includegraphics[width=\textwidth]{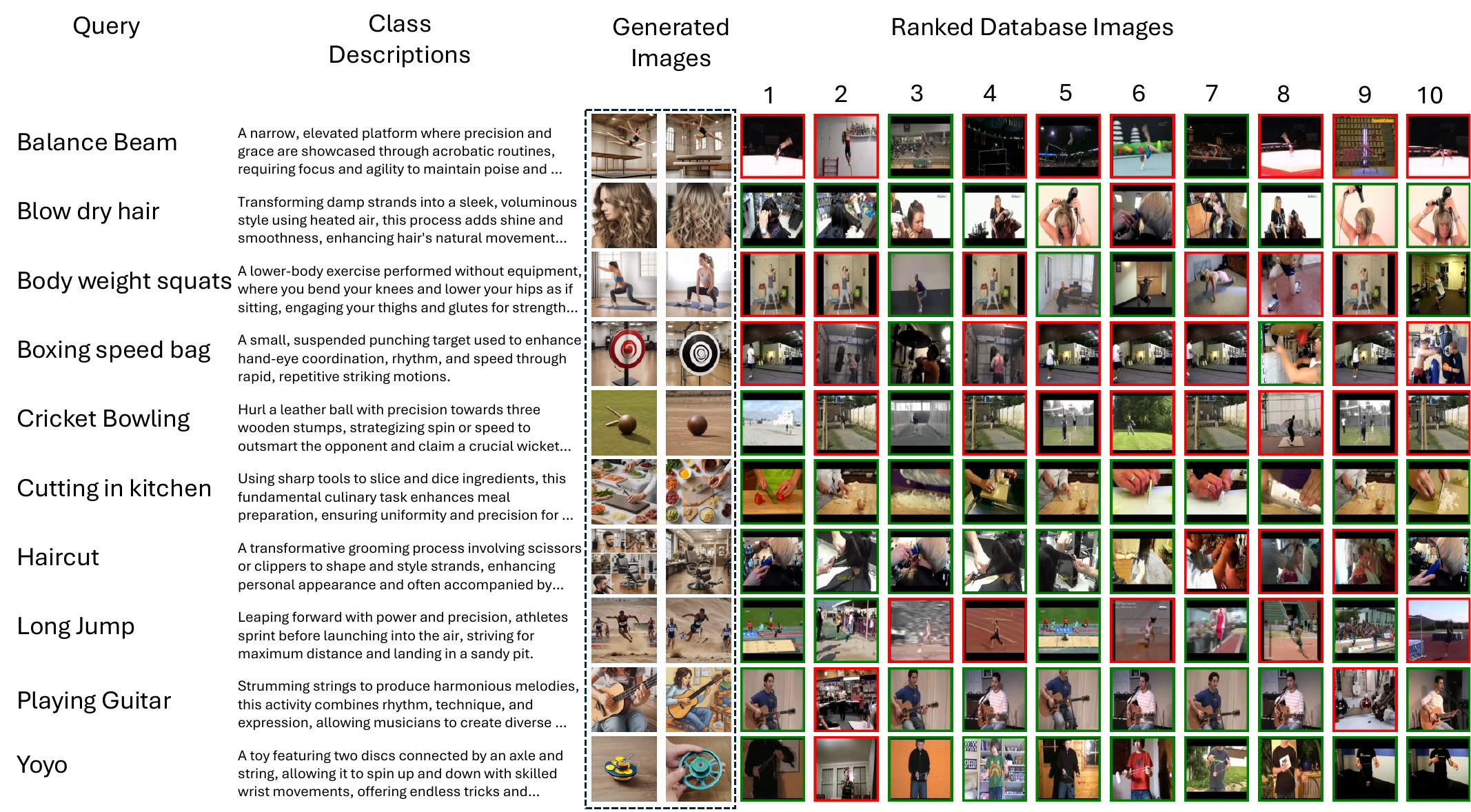}
    
    \caption{
    Description-based retrieval for UCF101.
    \label{fig:ucf_description}
    \vspace{-20pt}
    }
    \end{figure}

From~\Cref{fig:cal_class} till~\Cref{fig:ucf_description}, we present qualitative results for each dataset. For 10 randomly selected classes, we display the query class name, two images generated by SD~\cite{sd} for that class, and the top 10 ranked database images sorted by similarity. Correct matches are highlighted in green, while incorrect ones are shown in red. For description-based retrieval, where the same classes are queried using only their textual descriptions (without the class name), we show the class descriptions, generated images from the description, and top-ranked database images.

    \begin{figure}[t]
    
    \includegraphics[width=\textwidth]{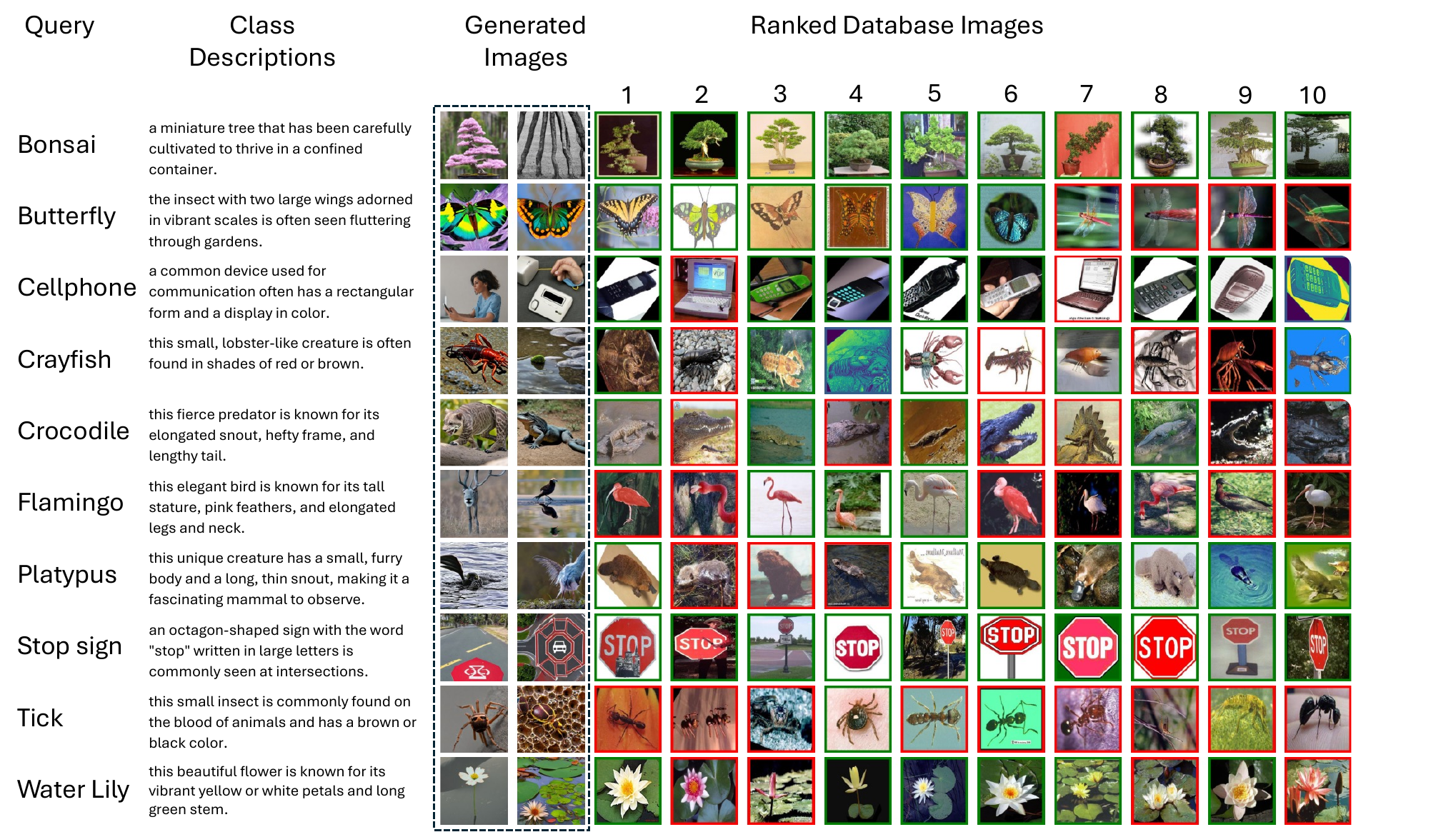}
    
    \caption{
    Description-based retrieval for Caltech101.
    \label{fig:cal_description}
    \vspace{-20pt}
    }
    \end{figure}
    
    \begin{figure}[t]
    \includegraphics[width=\textwidth]{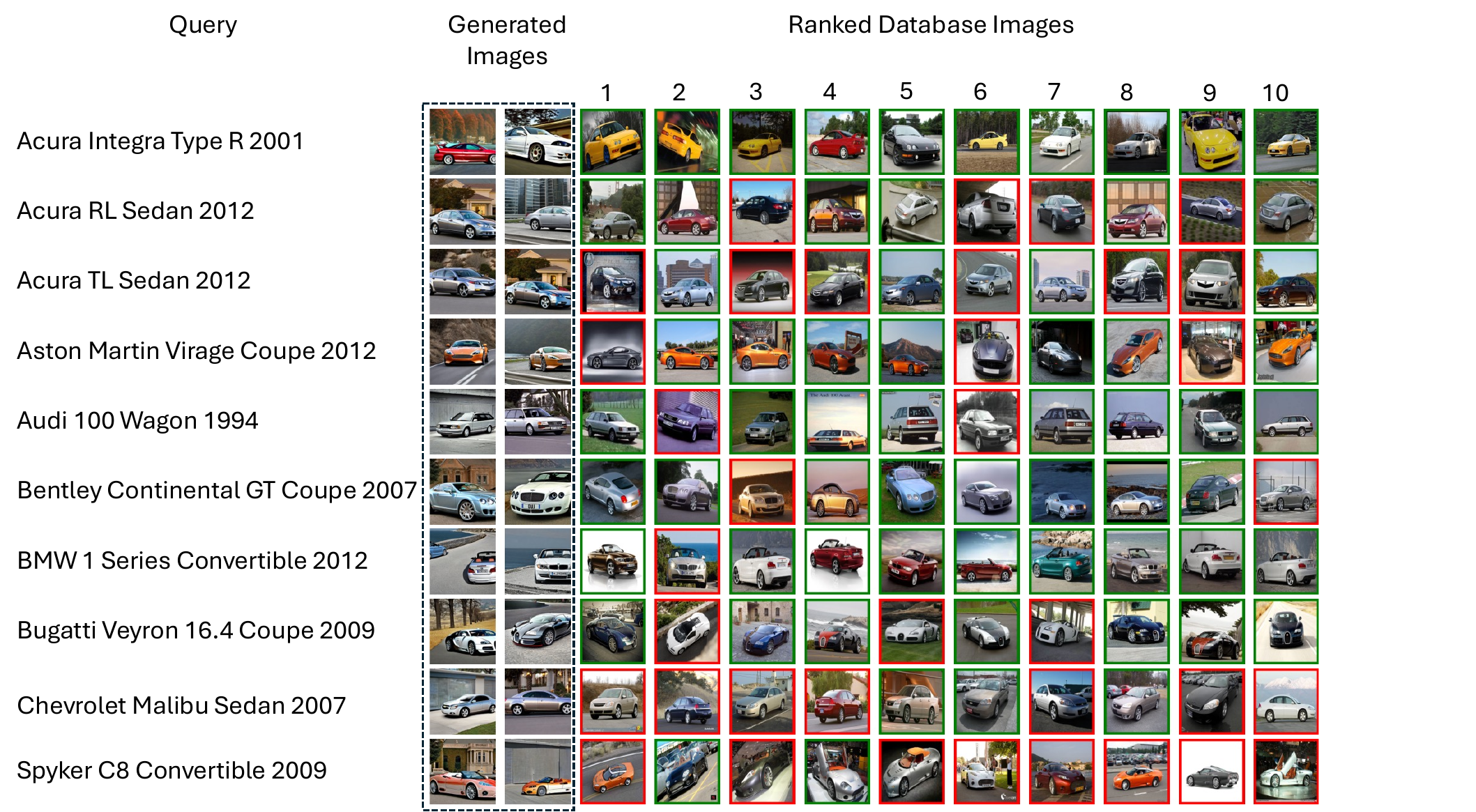}
    
    \caption{
    Class-based retrieval for Stanford Cars.
    \label{fig:cars_class}
    \vspace{-20pt}
    }
    \end{figure}

    \begin{figure}[t]
    
    \includegraphics[width=\textwidth]{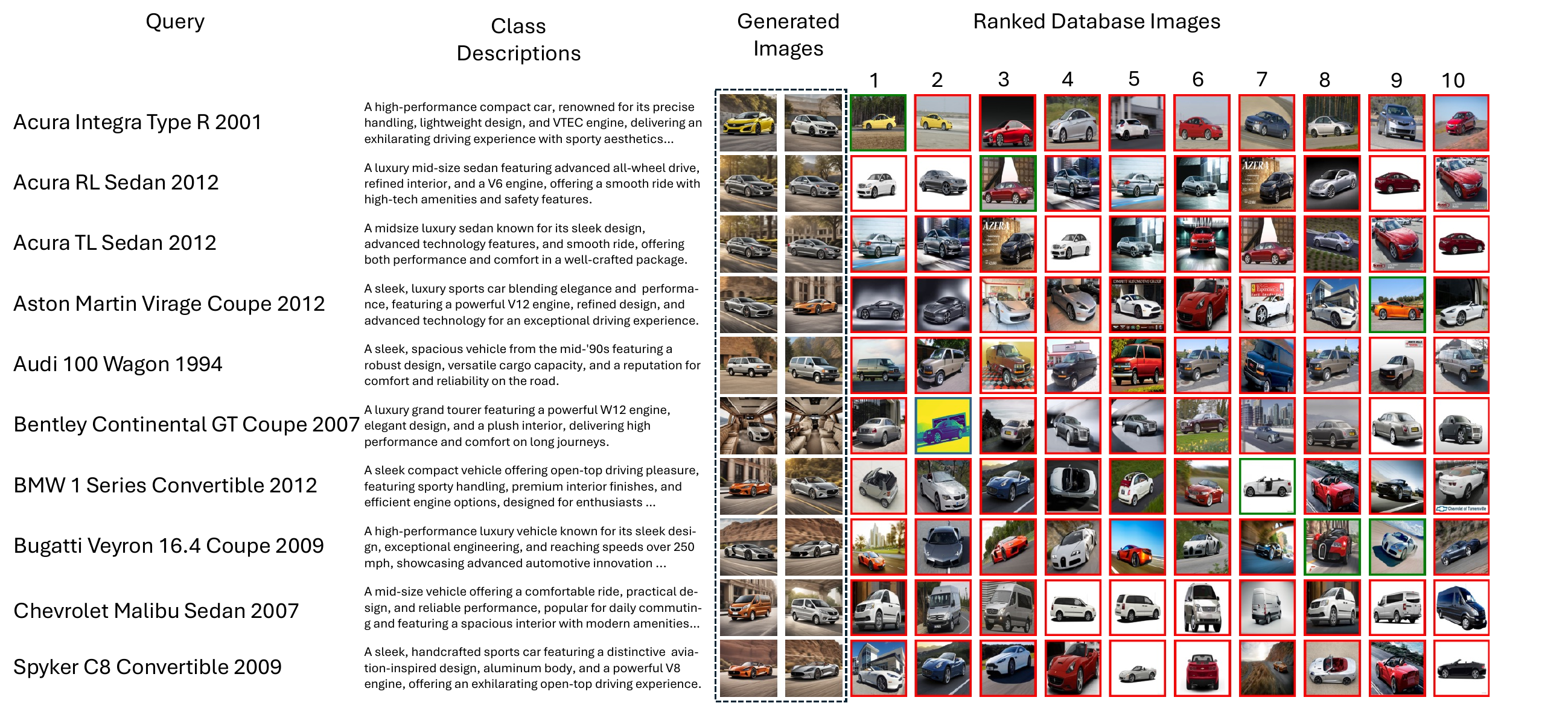}
    
    \caption{
    Description-based retrieval for Stanford Cars.
    \label{fig:cars_description}
    \vspace{-20pt}
    }
    \end{figure}
    
    \begin{figure}[t]
    \includegraphics[width=\textwidth]{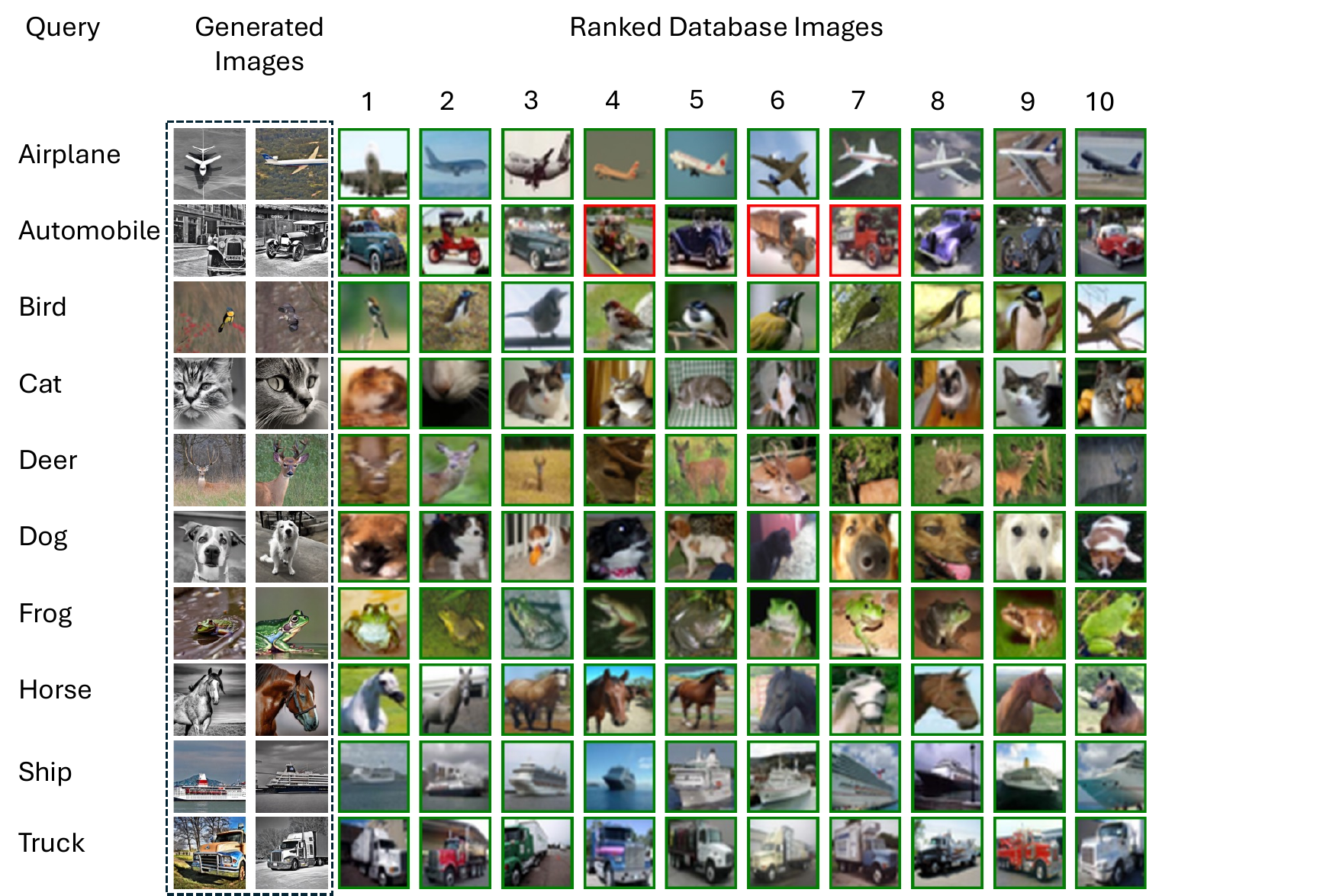}
    
    \caption{
    Class-based retrieval for CIFAR-10.
    \label{fig:cifar10_class}
    \vspace{-20pt}
    }
    \end{figure}

    \begin{figure}[t]
    
    \includegraphics[width=\textwidth]{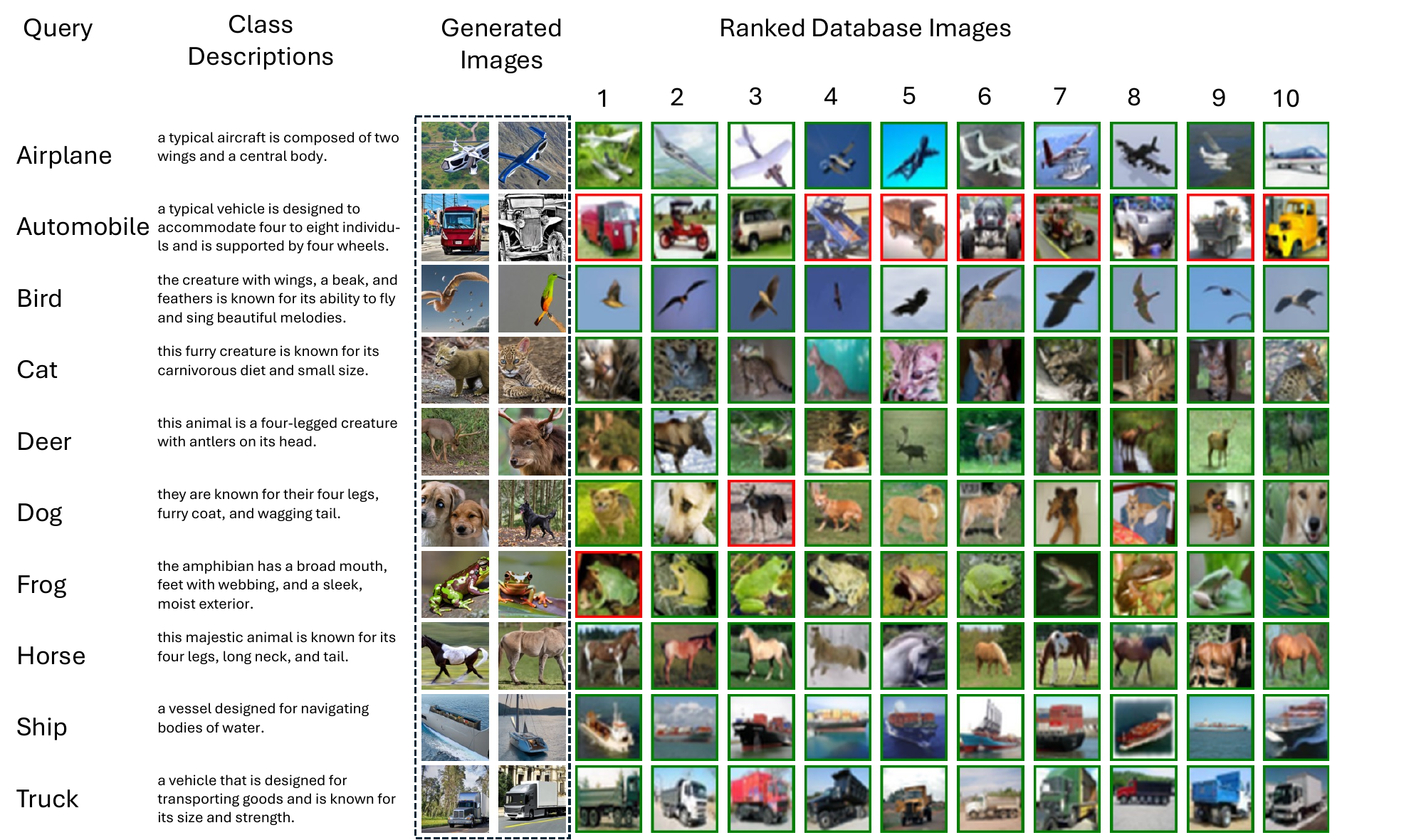}
    
    \caption{
    Description-based retrieval for CIFAR-10.
    \label{fig:cifar10_description}
    \vspace{-20pt}
    }
    \end{figure}
    
    \begin{figure}[t]
    \includegraphics[width=\textwidth]{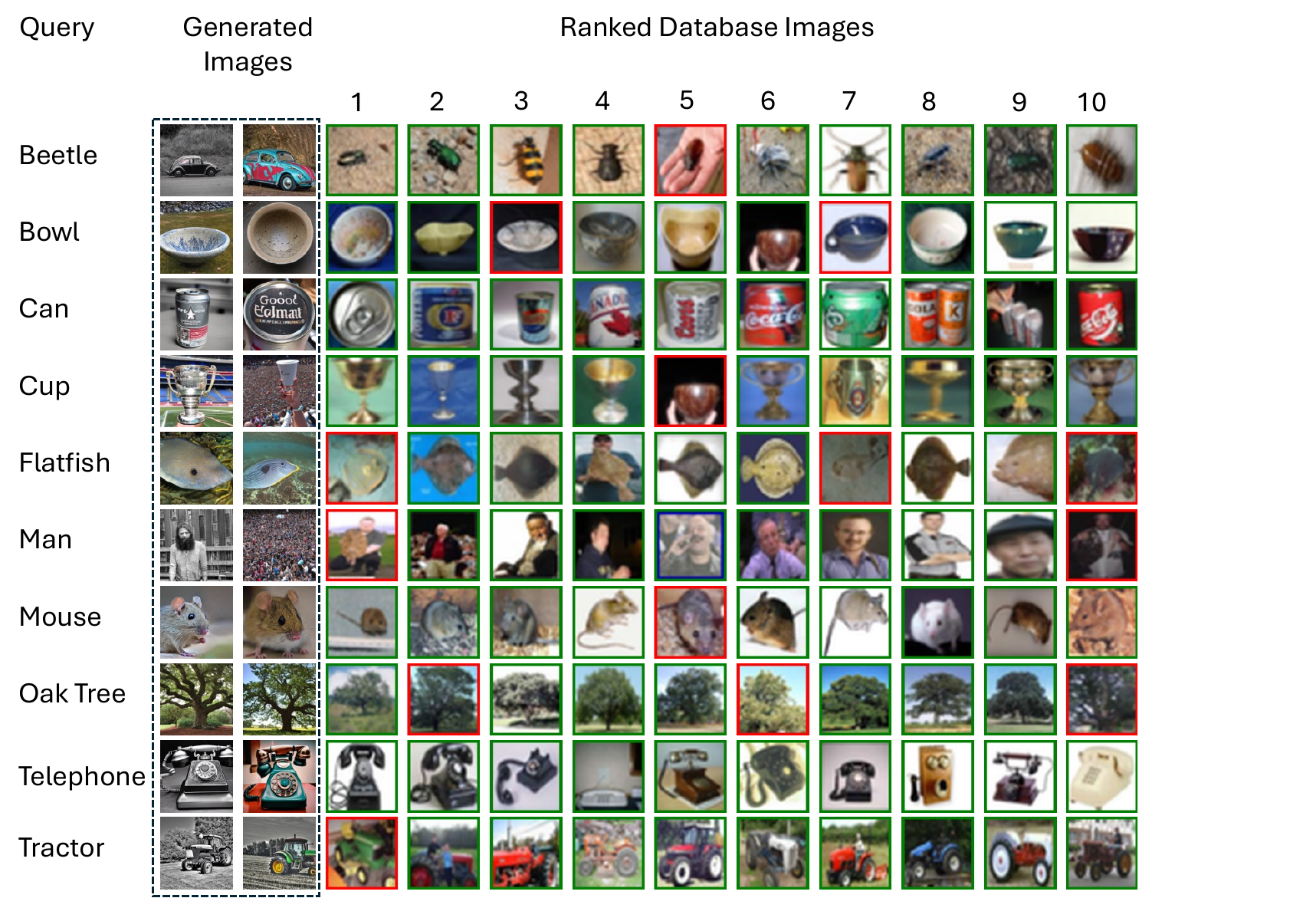}
    
    \caption{
    Class-based retrieval for CIFAR-100.
    \label{fig:cifar100_class}
    \vspace{-20pt}
    }
    \end{figure}

    \begin{figure}[t]
    
    \includegraphics[width=\textwidth]{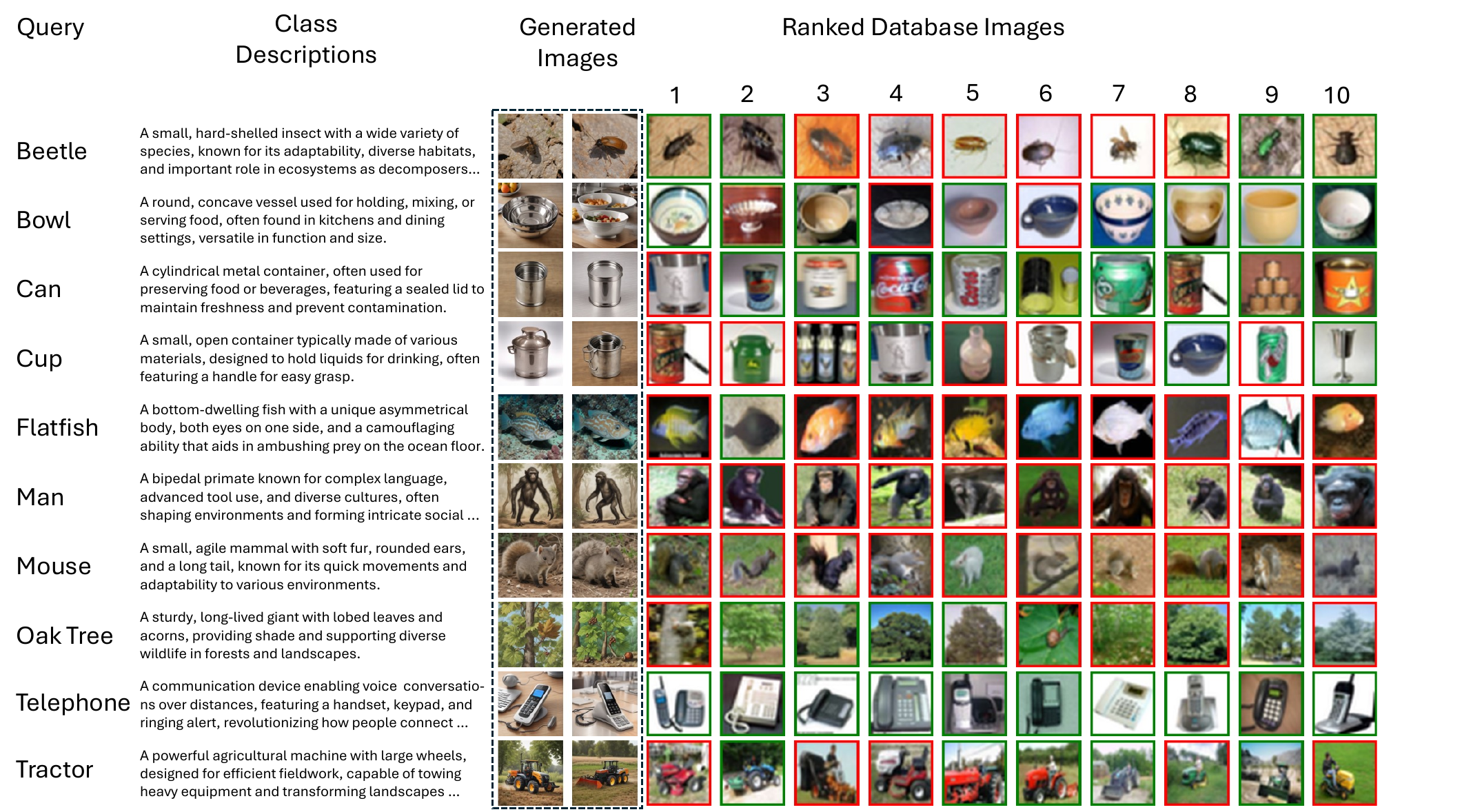}
    
    \caption{
    Description-based retrieval for CIFAR-100.
    \label{fig:cifar100_description}
    \vspace{-20pt}
    }
    \end{figure}
    
    \begin{figure}[t]
    \includegraphics[width=\textwidth]{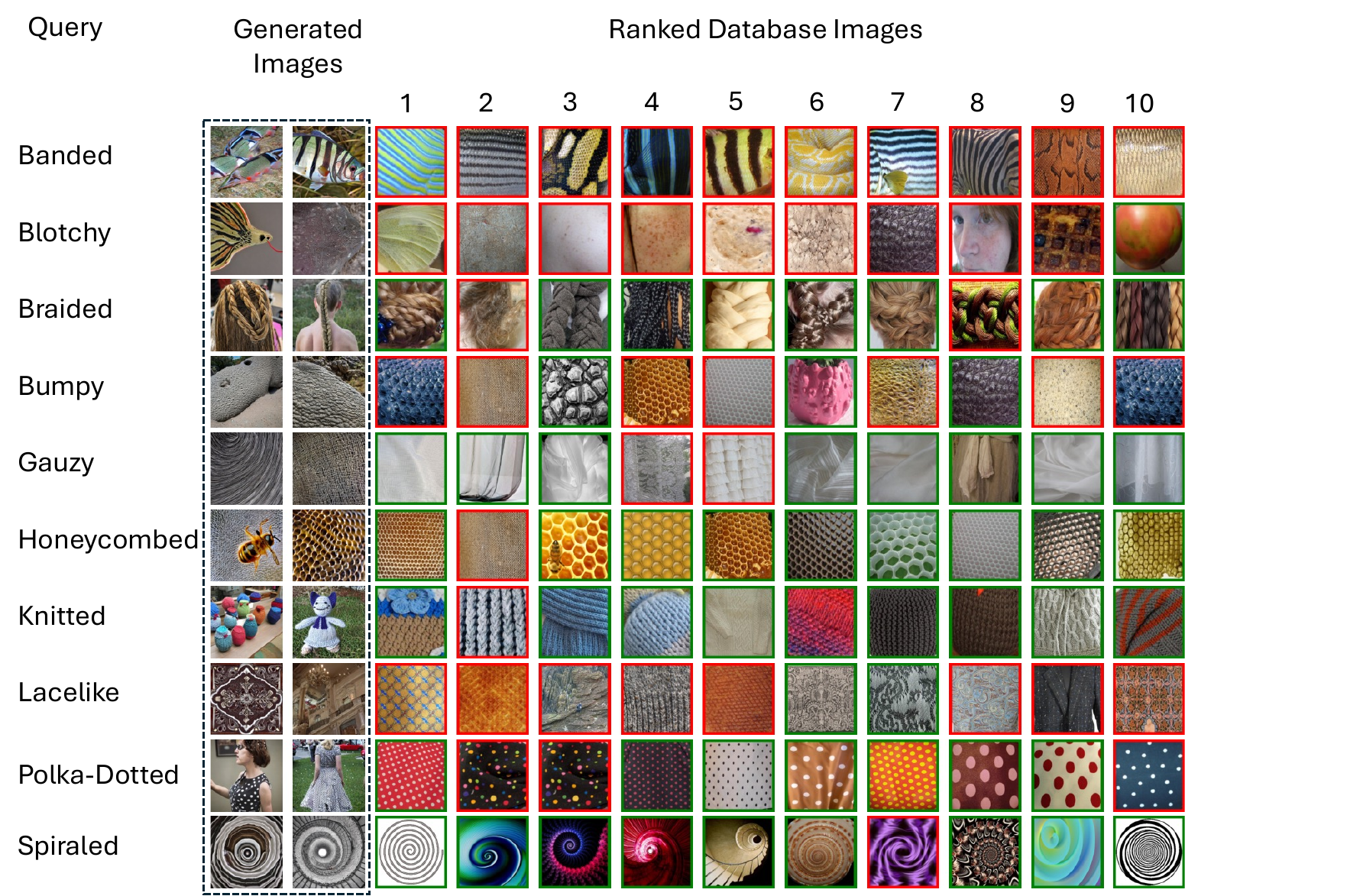}
    
    \caption{
    Class-based retrieval for DTD.
    \label{fig:dtd_class}
    \vspace{-20pt}
    }
    \end{figure}

    \begin{figure}[t]
    
    \includegraphics[width=\textwidth]{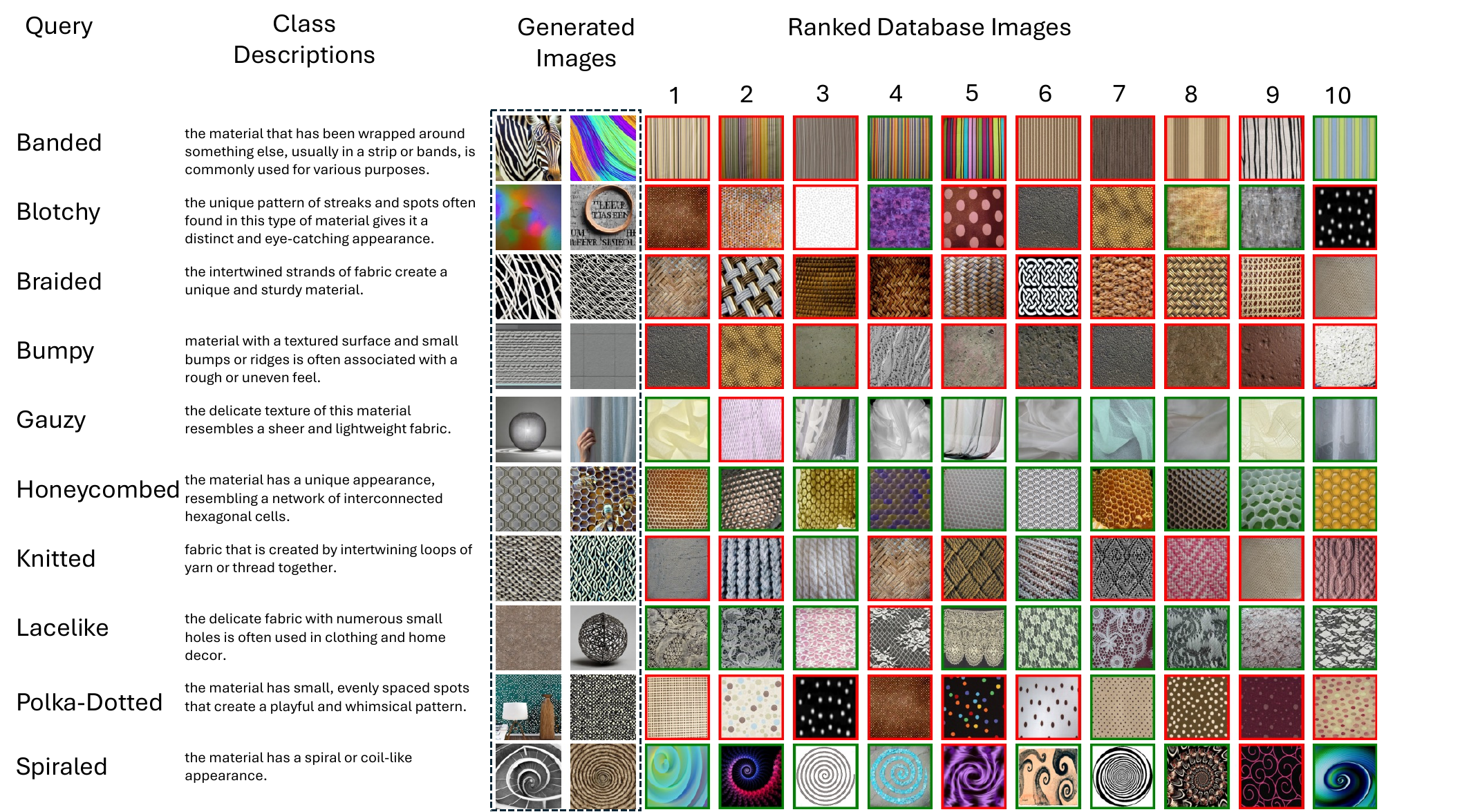}
    
    \caption{
    Description-based retrieval for DTD.
    \label{fig:dtd_description}
    \vspace{-20pt}
    }
    \end{figure}
    
    \begin{figure}[t]
    \includegraphics[width=0.9\textwidth]{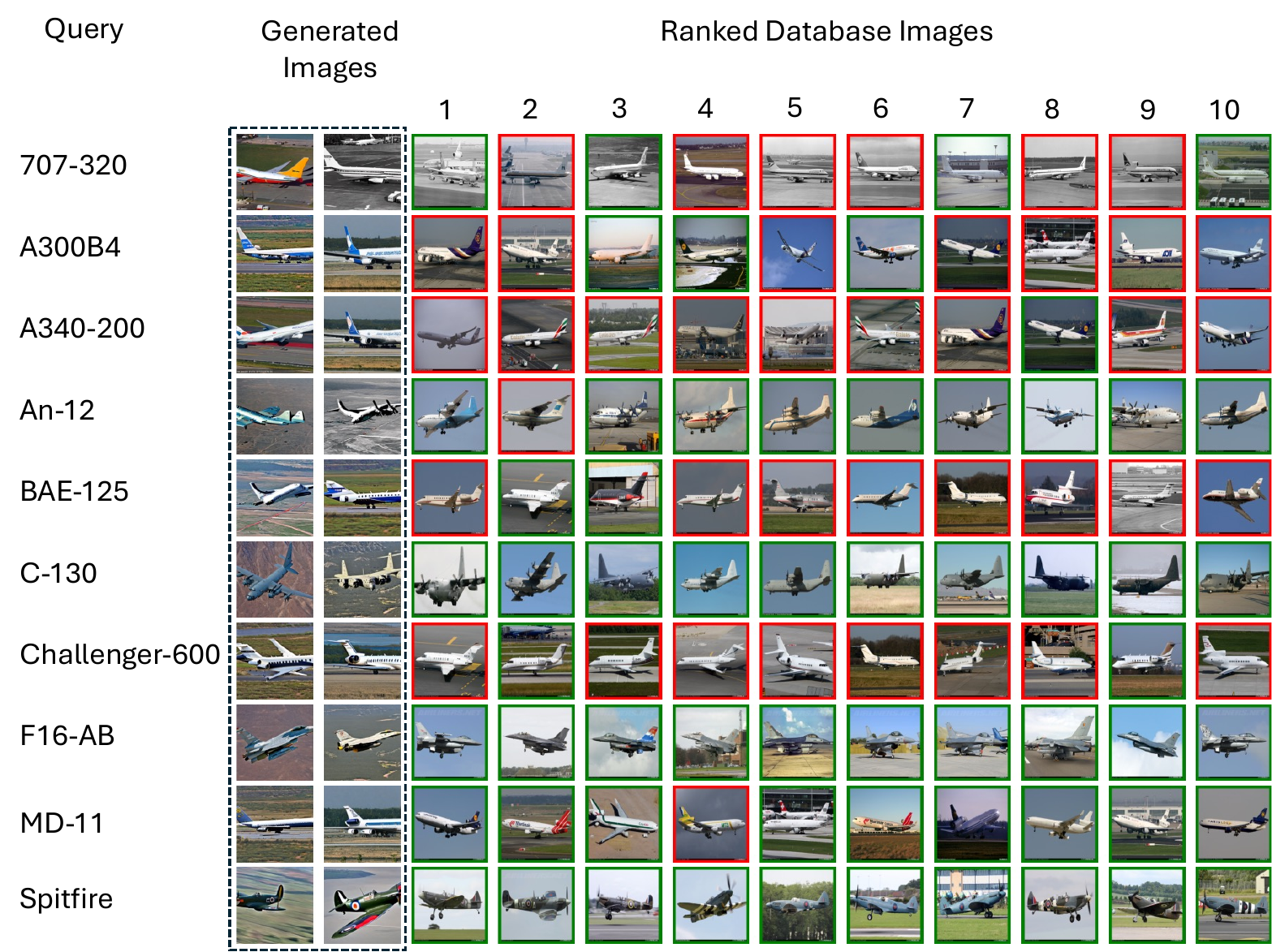}
    
    \caption{
    Class-based retrieval for FGVC Aircraft.
    \label{fig:fgvc_aircraft_class}
    \vspace{-20pt}
    }
    \end{figure}

    \begin{figure}[t]
    
    \includegraphics[width=0.9\textwidth]{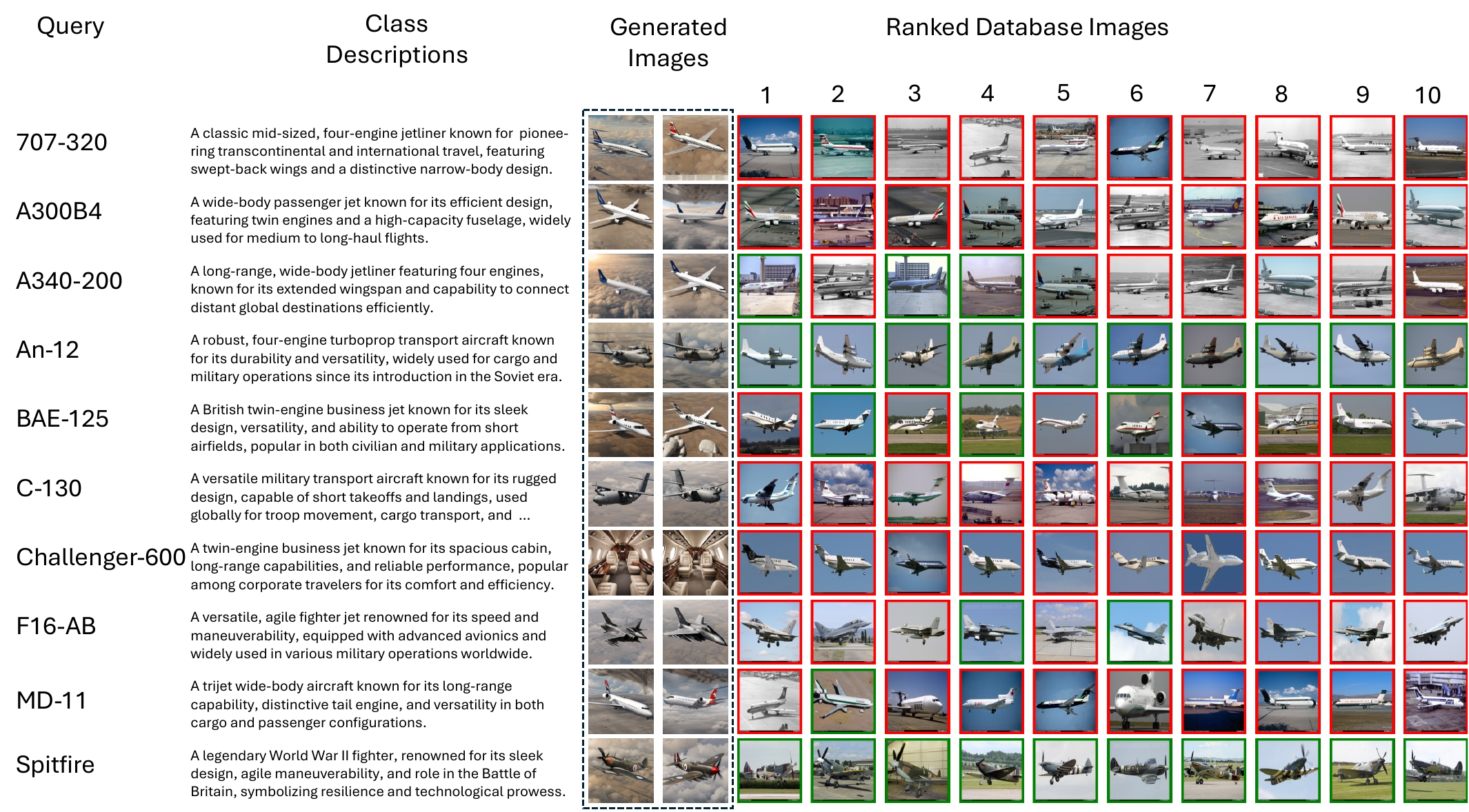}
    
    \caption{
    Description-based retrieval for FGVC Aircraft.
    \label{fig:fgvc_aircraft_description}
    \vspace{-20pt}
    }
    \end{figure}
    
    \begin{figure}[t]
    \includegraphics[width=\textwidth]{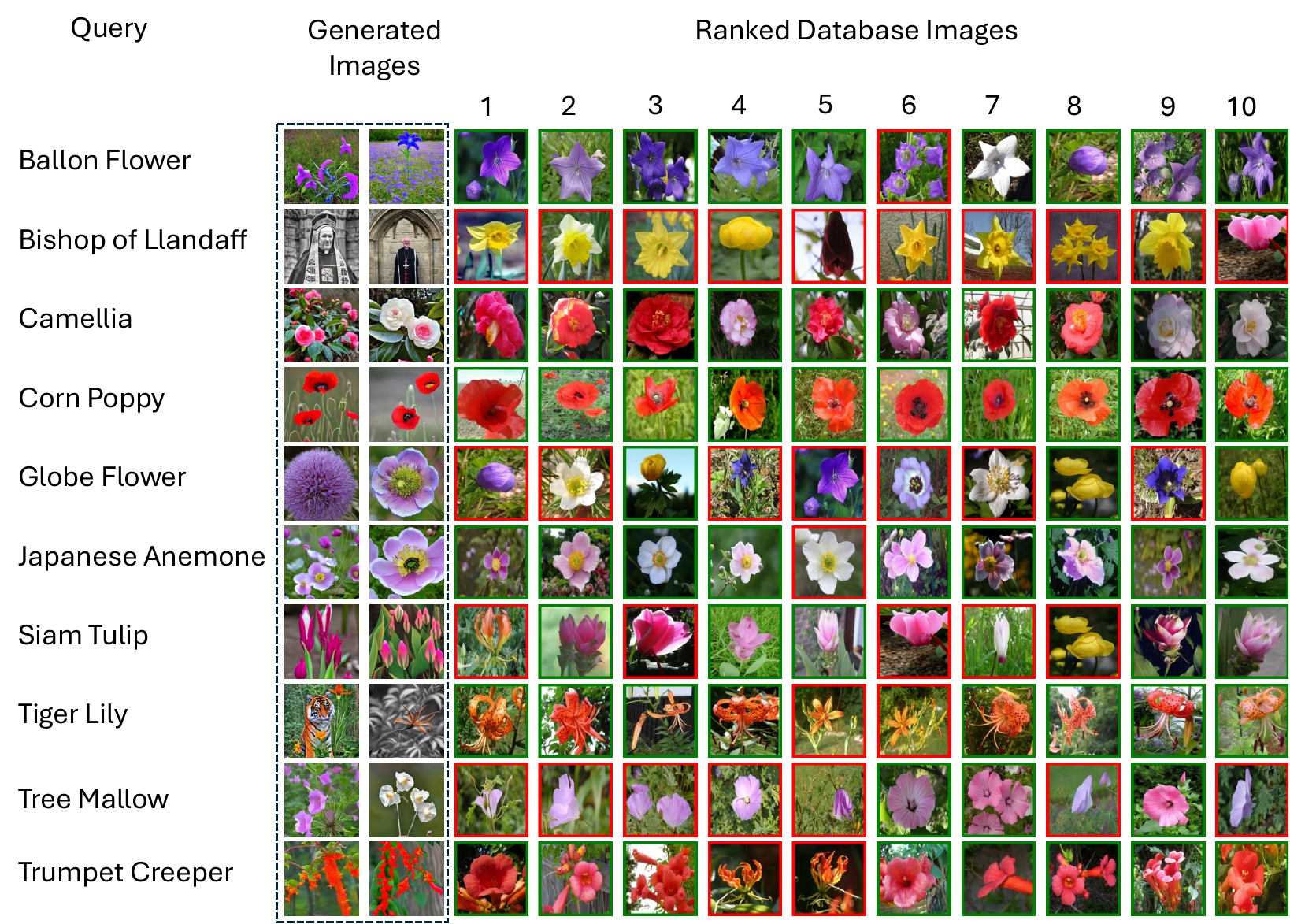}
    
    \caption{
    Class-based retrieval for Flowers 102.
    \label{fig:flowers_class}
    \vspace{-20pt}
    }
    \end{figure}

    \begin{figure}[t]
    
    \includegraphics[width=\textwidth]{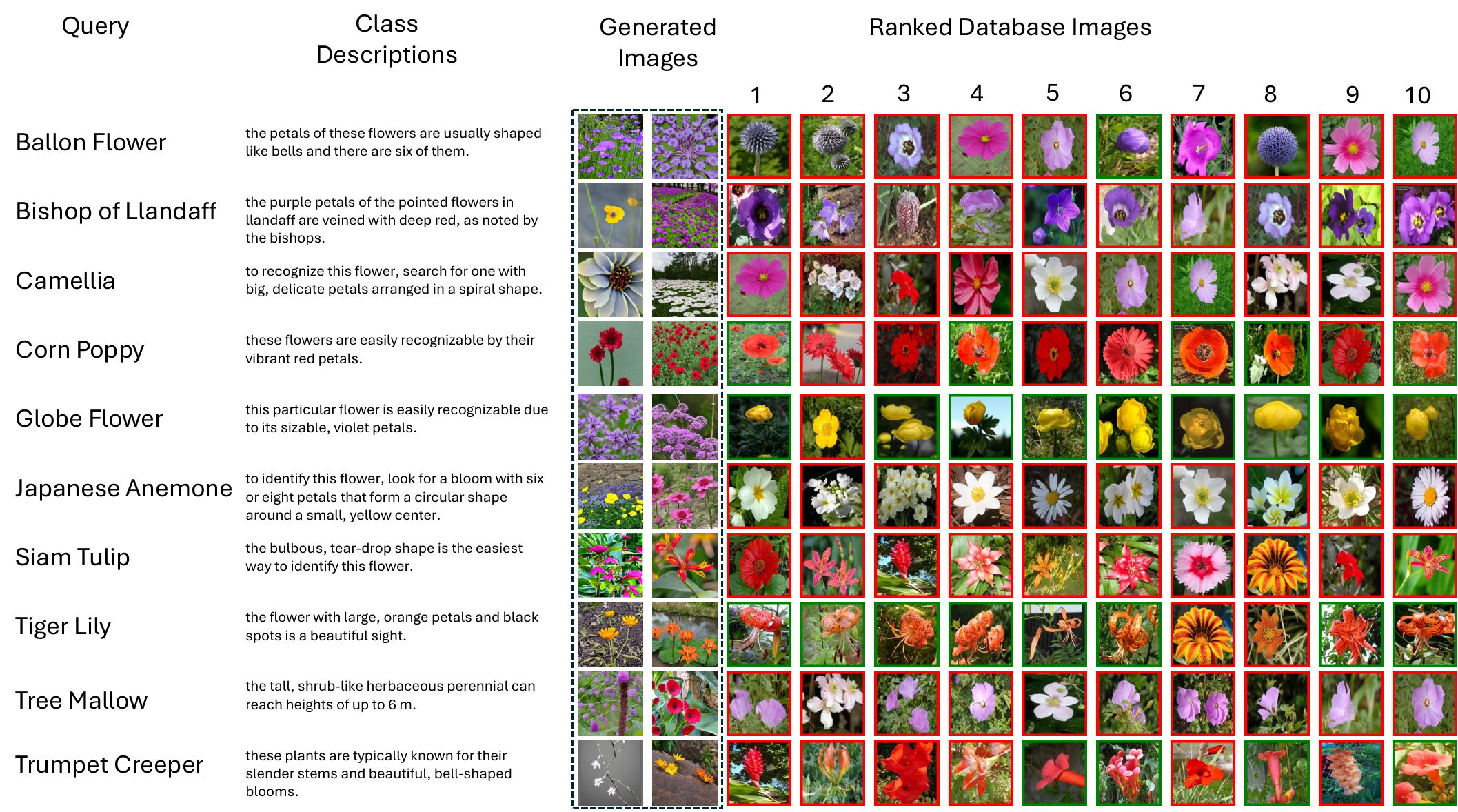}
    
    \caption{
    Description-based retrieval for Flowers 102.
    \label{fig:flowers_description}
    \vspace{-20pt}
    }
    \end{figure}
    
    \begin{figure}[t]
    \includegraphics[width=\textwidth]{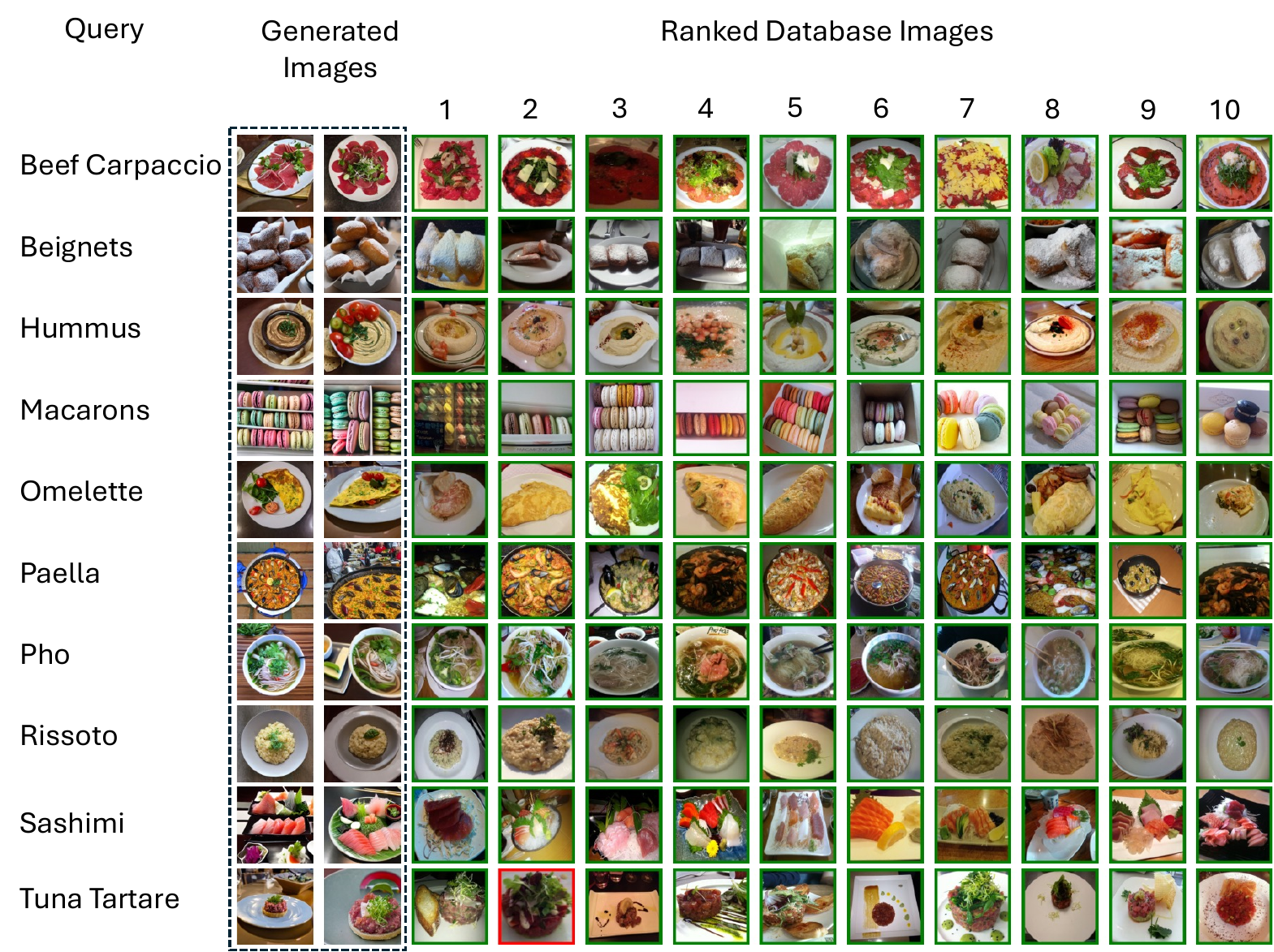}
    
    \caption{
    Class-based retrieval for Food.
    \label{fig:food_class}
    \vspace{-20pt}
    }
    \end{figure}

    \begin{figure}[t]
    
    \includegraphics[width=\textwidth]{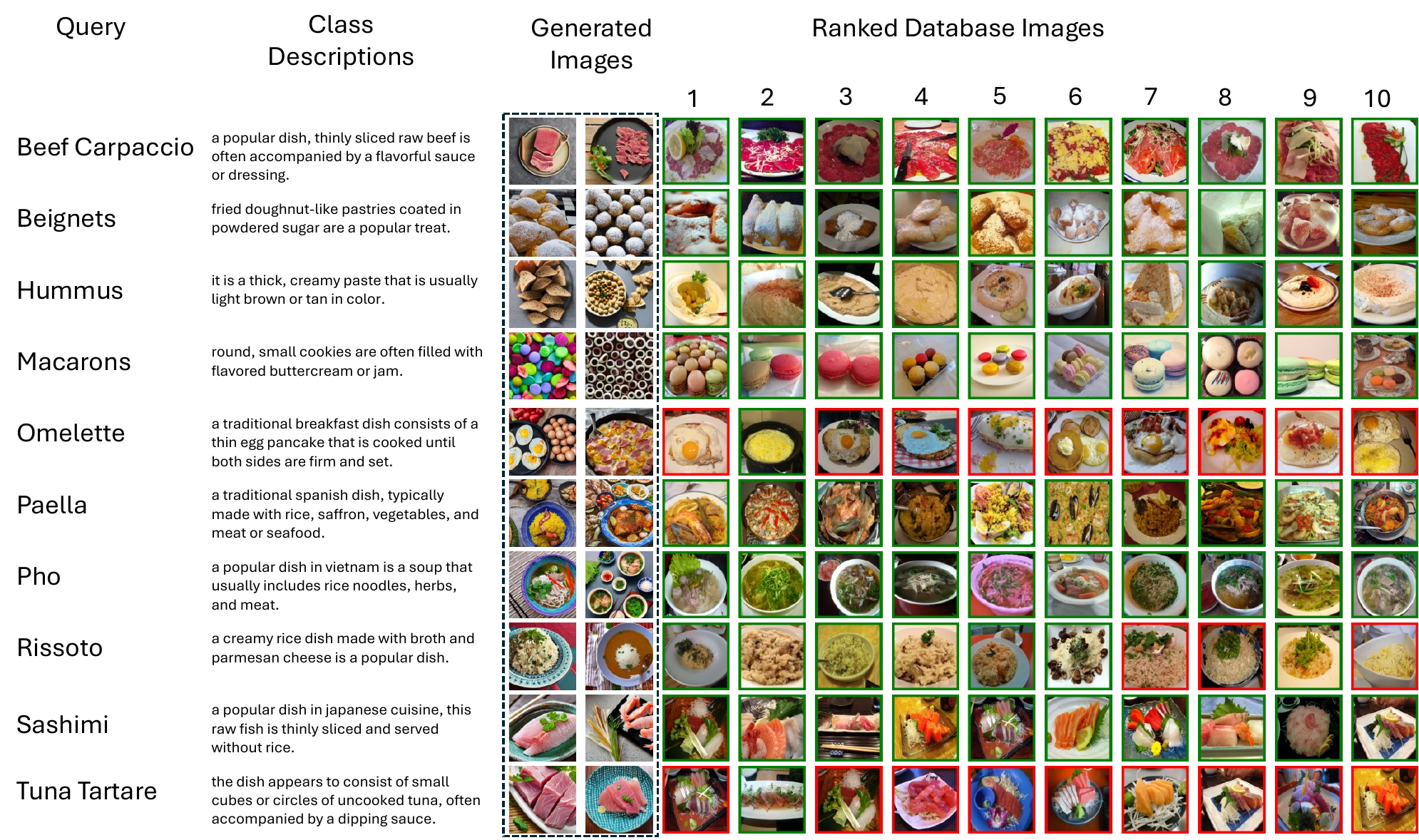}
    
    \caption{
    Description-based retrieval for Food.
    \label{fig:food_description}
    \vspace{-20pt}
    }
    \end{figure}

    \begin{figure}[t]
    \includegraphics[width=\textwidth]{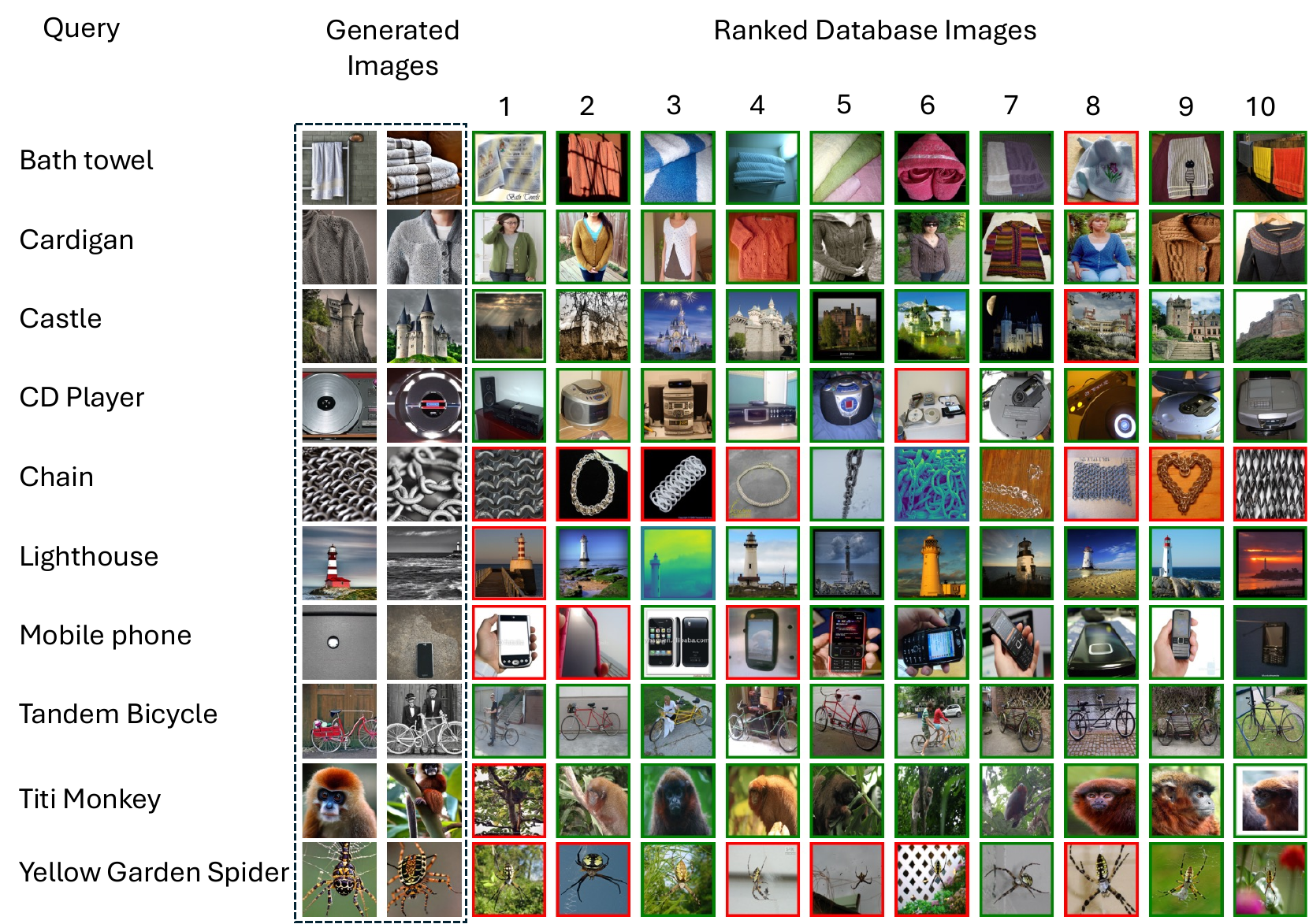}
    
    \caption{
    Class-based retrieval for ImageNet.
    \label{fig:imagenet_class}
    \vspace{-20pt}
    }
    \end{figure}

    \begin{figure}[t]
    
    \includegraphics[width=\textwidth]{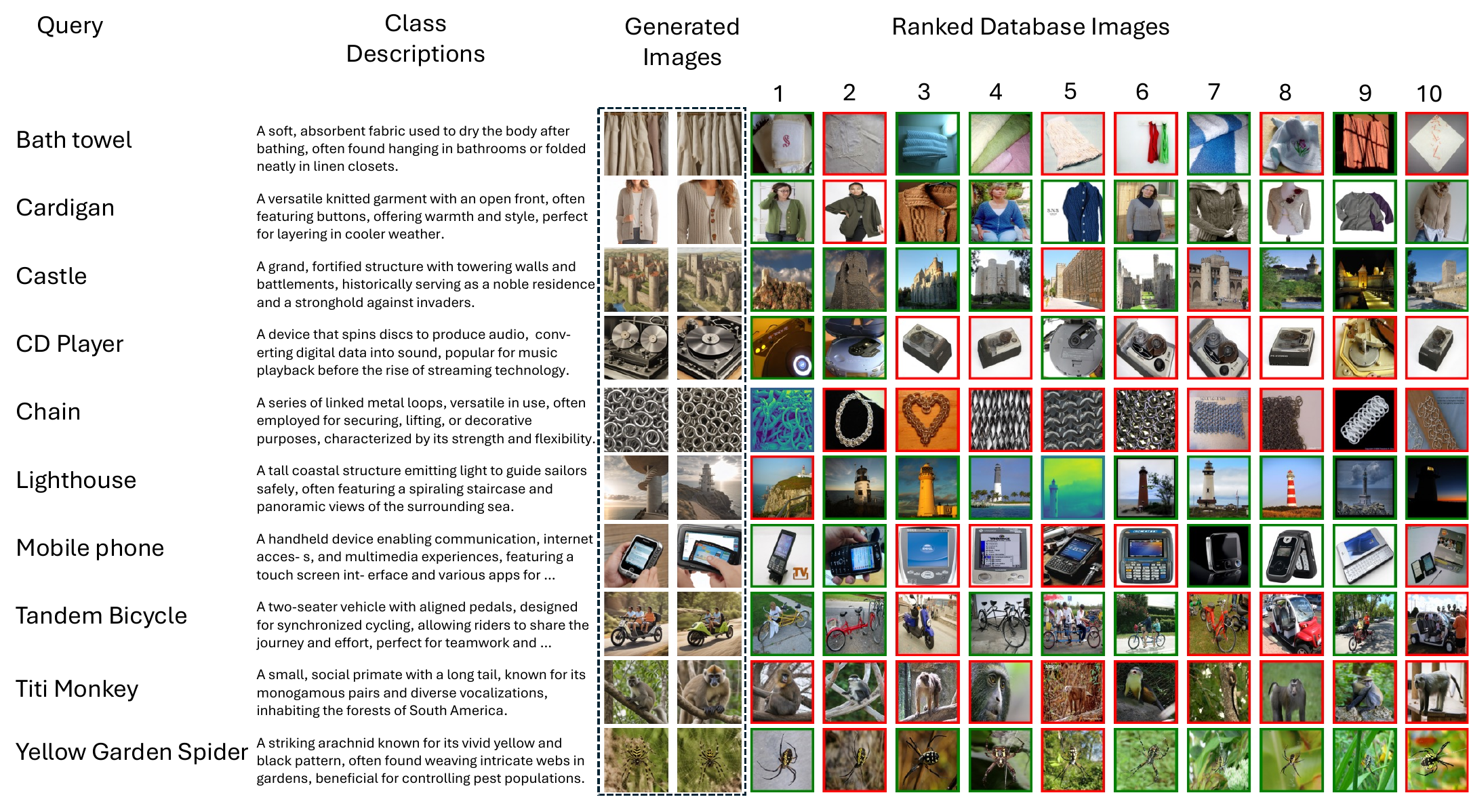}
    
    \caption{
    Description-based retrieval for ImageNet.
    \label{fig:imagenet_description}
    \vspace{-20pt}
    }
    \end{figure}

    \begin{figure}[t]
    \includegraphics[width=\textwidth]{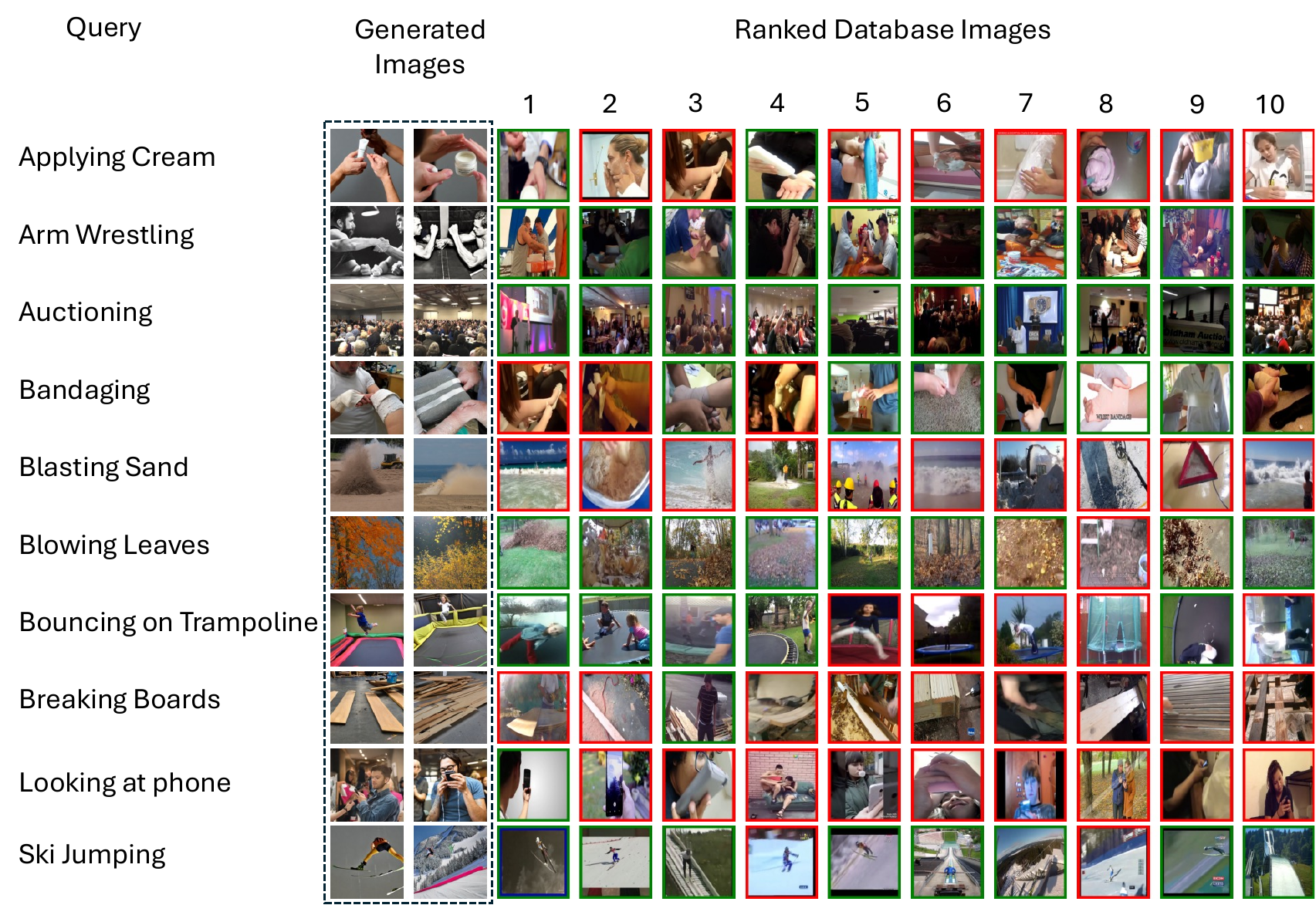}
    
    \caption{
    Class-based retrieval for Kinetics-700.
    \label{fig:k700_class}
    \vspace{-20pt}
    }
    \end{figure}

    \begin{figure}[t]
    
    \includegraphics[width=\textwidth]{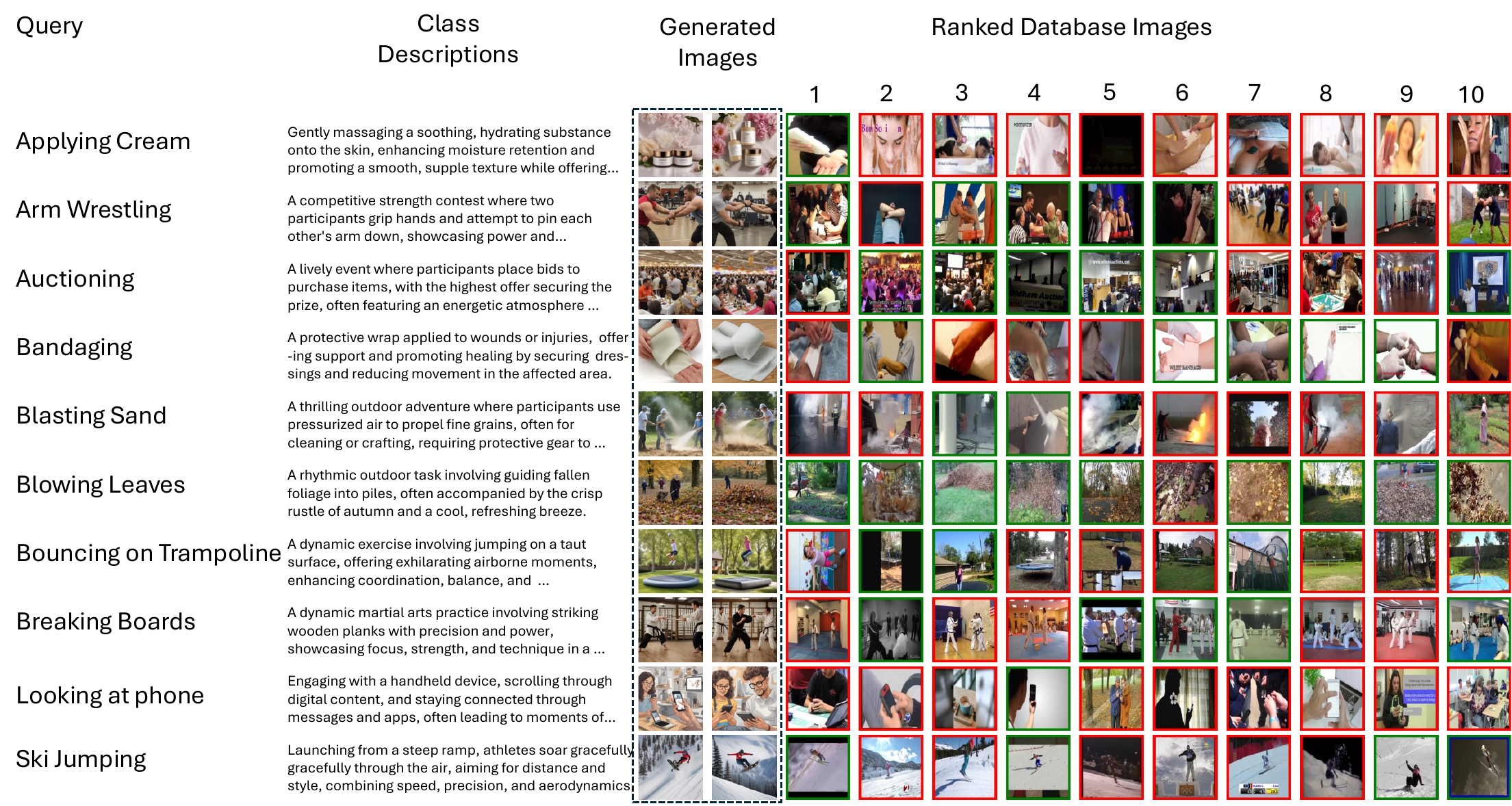}
    
    \caption{
    Description-based retrieval for Kinetics-700.
    \label{fig:k700_description}
    \vspace{-20pt}
    }
    \end{figure}

    \begin{figure}[t]
    \includegraphics[width=\textwidth]{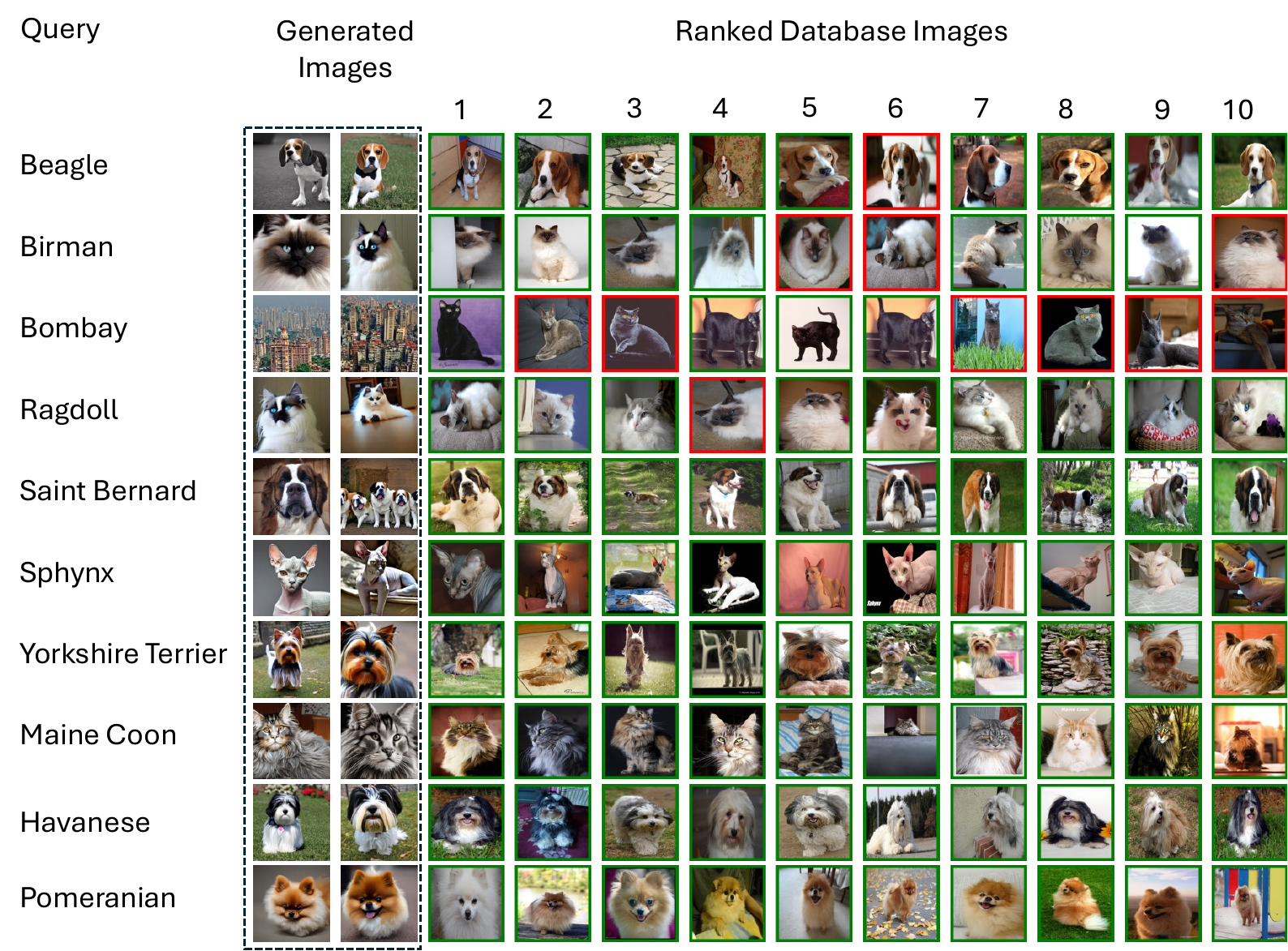}
    
    \caption{
    Class-based retrieval for Oxford Pets.
    \label{fig:pets_class}
    \vspace{-20pt}
    }
    \end{figure}

    \begin{figure}[t]
    
    \includegraphics[width=\textwidth]{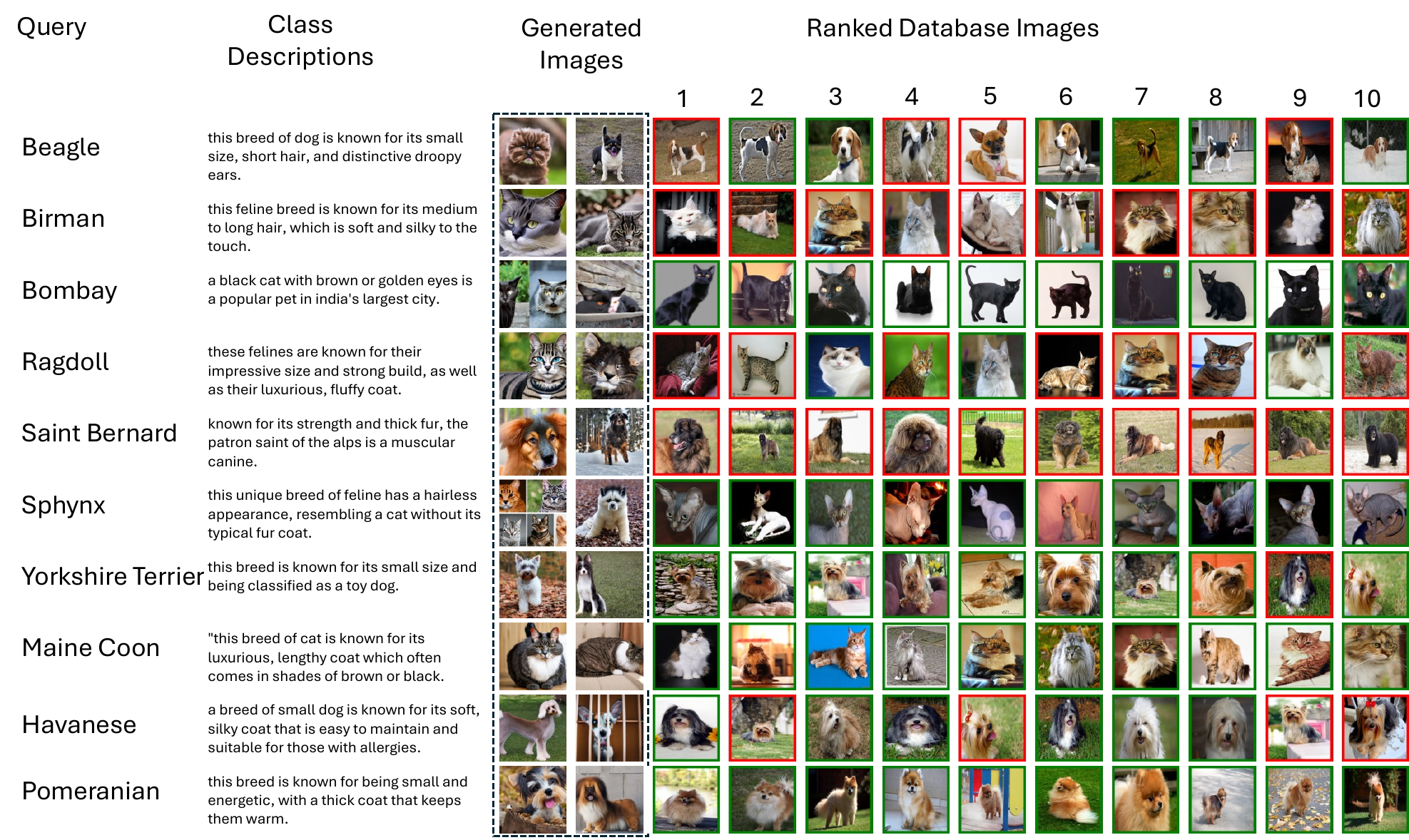}
    
    \caption{
    Description-based retrieval for Oxford Pets.
    \label{fig:pets_description}
    \vspace{-20pt}
    }
    \end{figure}

    \begin{figure}[t]
    \includegraphics[width=\textwidth]{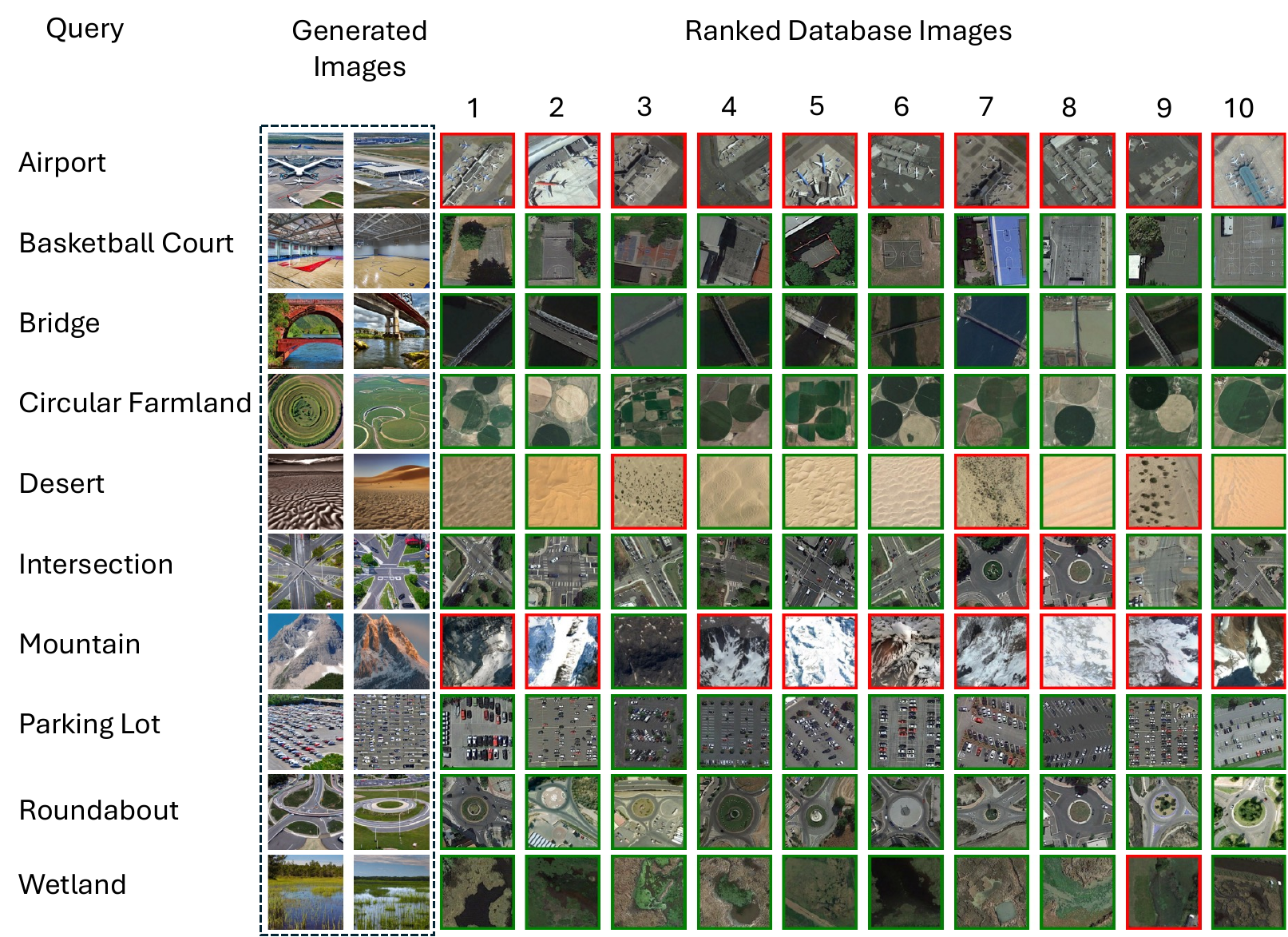}
    
    \caption{
    Class-based retrieval for RESISC-45.
    \label{fig:r45_class}
    \vspace{-20pt}
    }
    \end{figure}

    \begin{figure}[t]
    
    \includegraphics[width=\textwidth]{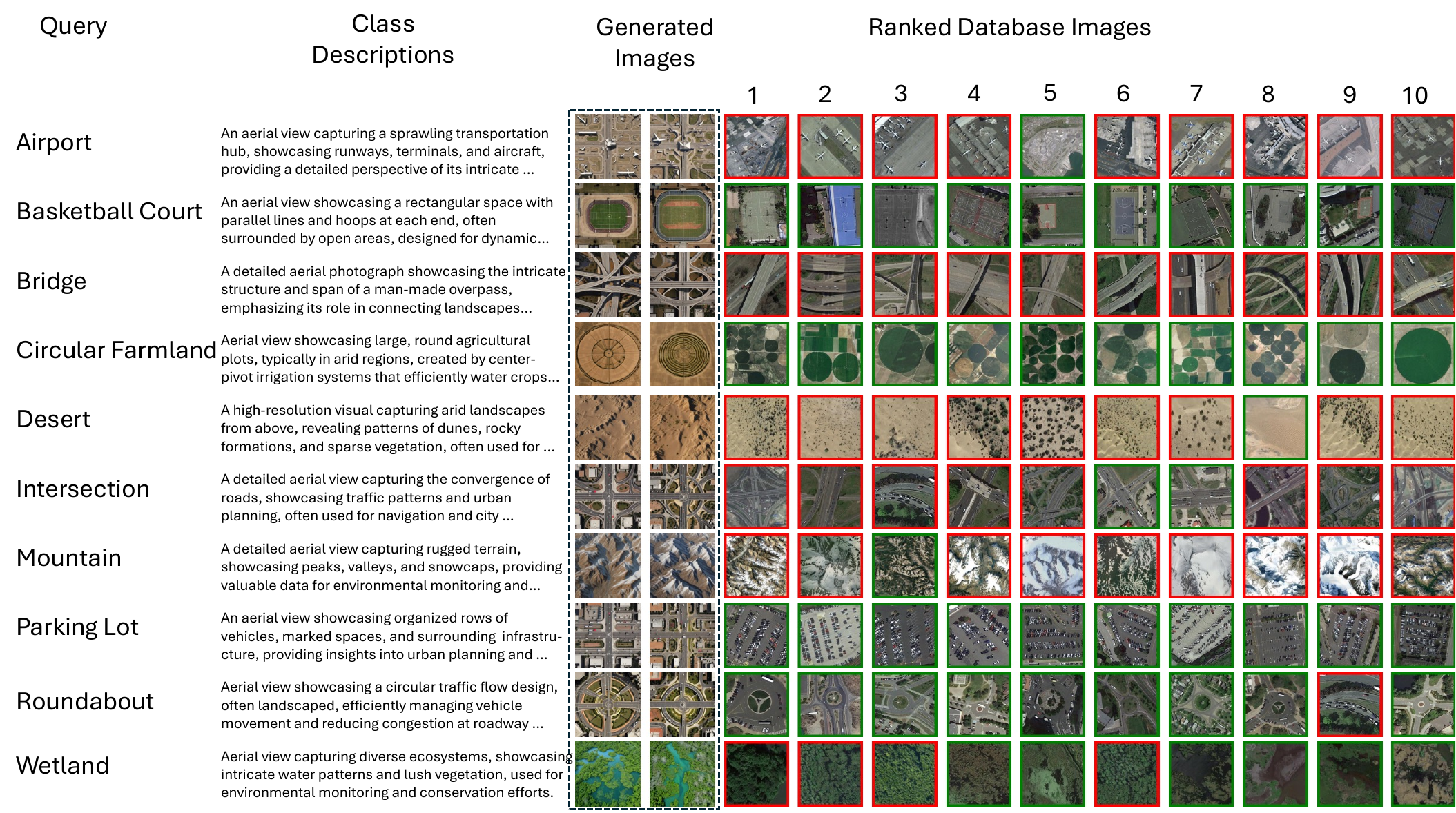}
    
    \caption{
    Description-based retrieval for RESISC-45.
    \label{fig:r45_description}
    \vspace{-20pt}
    }
    \end{figure}

    \begin{figure}[t]
    \includegraphics[width=\textwidth]{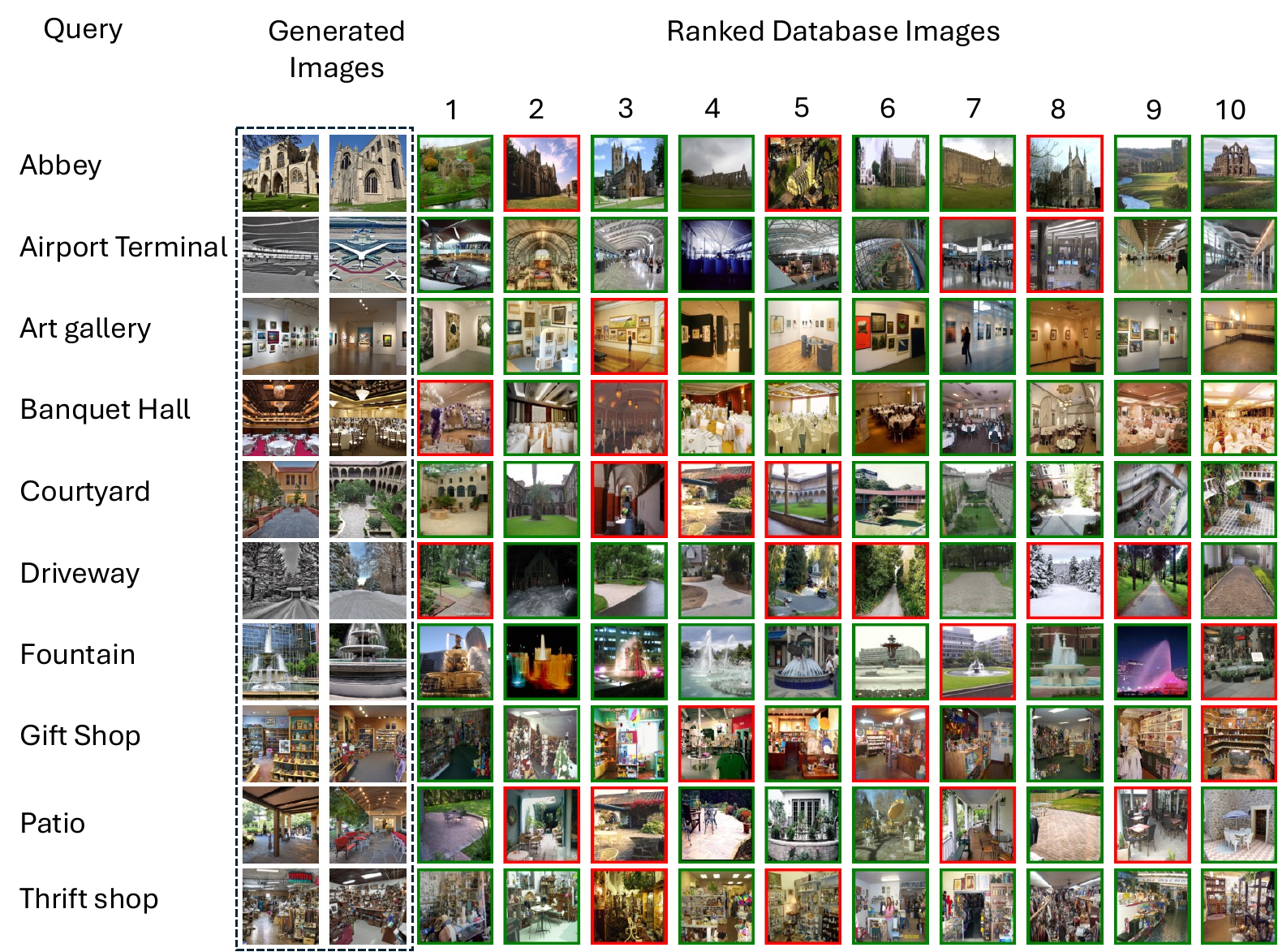}
    
    \caption{
    Class-based retrieval for SUN397.
    \label{fig:sun397_class}
    \vspace{-20pt}
    }
    \end{figure}

    \begin{figure}[t]
    
    \includegraphics[width=\textwidth]{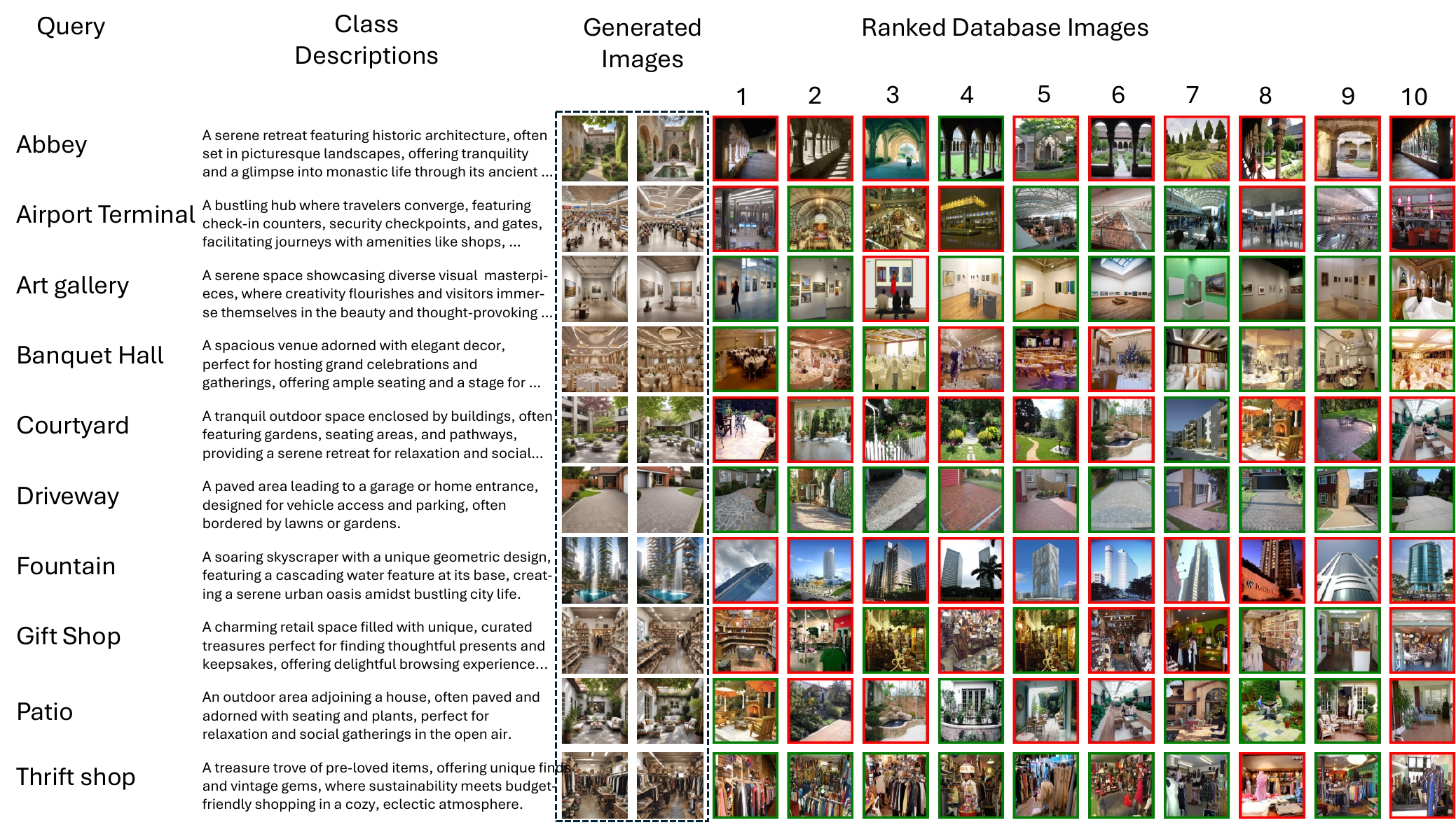}
    
    \caption{
    Description-based retrieval for SUN397.
    \label{fig:sun397_description}
    \vspace{-20pt}
    }
    \end{figure}

    \begin{figure}[t]
    \includegraphics[width=\textwidth]{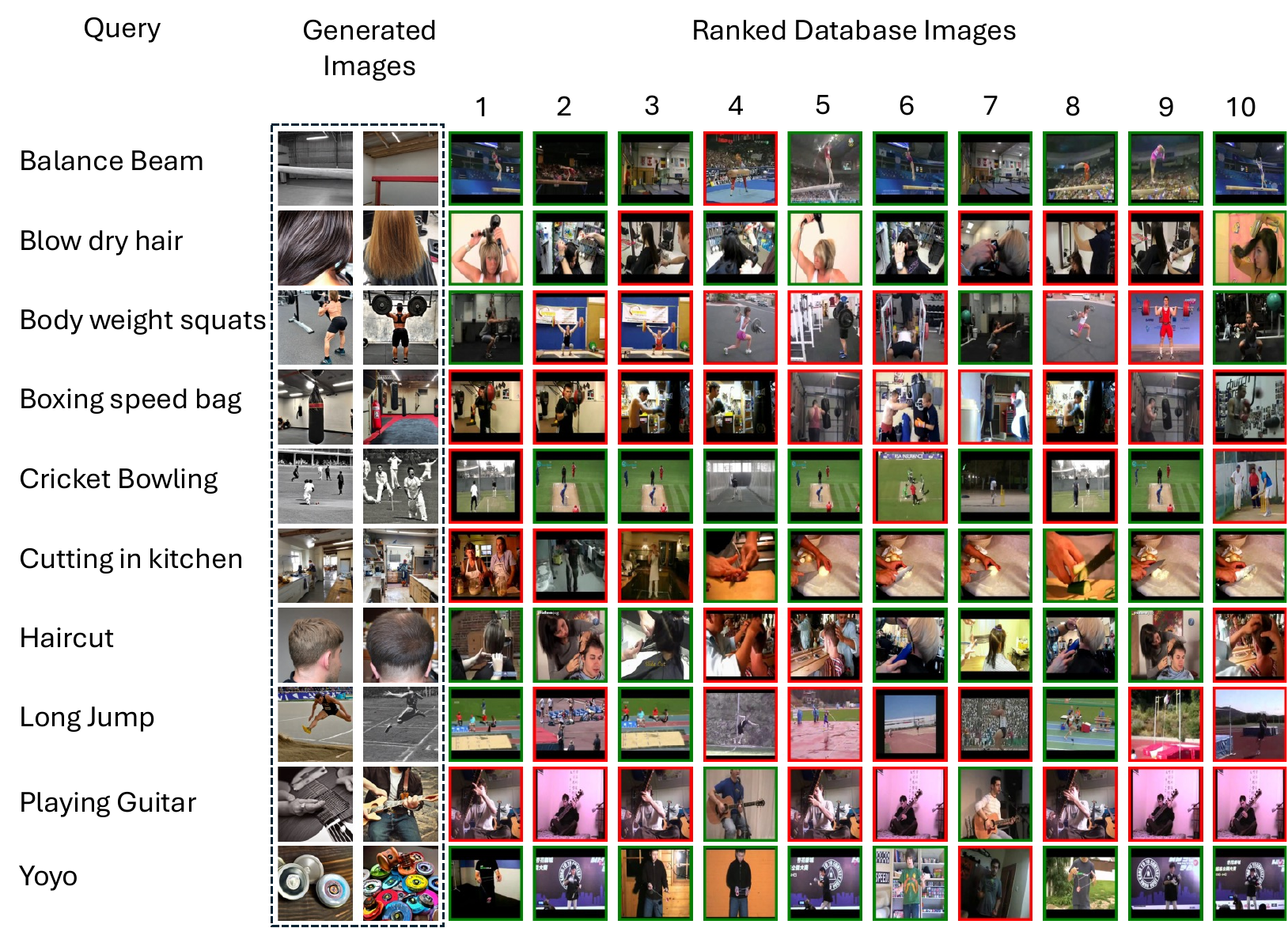}
    
    \caption{
    Class-based retrieval for UCF101.
    \label{fig:ucf_class}
    \vspace{-20pt}
    }
    \end{figure}

\end{document}